\documentclass[11pt, a4paper, logo, copyright, gdm]{google}

\usepackage[authoryear, sort&compress, round]{natbib}

\usepackage{xspace}
\usepackage{tcolorbox}
\tcbuselibrary{breakable, skins}
\usepackage{algorithm}
\usepackage{mathtools}
\usepackage{algpseudocode}
\usepackage{amsmath}
\usepackage{amssymb}
\usepackage{float}
\usepackage{multirow}
\usepackage{enumitem}
\usepackage{subcaption}
\usepackage{listings}
\usepackage{xcolor}
\usepackage{multicol}
\usepackage[
  smartEllipses,
  hashEnumerators,    
  fencedCode,         
  pipeTables,         
]{markdown}
\usepackage{pdflscape} 
\usepackage{rotating}  
\usepackage{booktabs}  

\definecolor{codebg}{HTML}{F8F9FA}
\definecolor{codeborder}{HTML}{DEE2E6}

\newcommand{\inlinecode}[1]{\fcolorbox{codeborder}{codebg}{\texttt{#1}}}

\uselogo{}

\title{DiffusionGemma Technical Report}

\newtoks\authfootnote

\author{DiffusionGemma Team, Google DeepMind\textsuperscript{1}}

\renewcommand{\copyrightext}{%
  \footerfont \textsuperscript{1}Contributions, Acknowledgements and full author list: \hyperref[sec:contributions]{Page \pageref*{sec:contributions}}. Correspondence: \texttt{diffusiongemma-external@google.com}.\\[1pt]
  \textsuperscript{2} The Gemma 4 E4B and E2B models are designed primarily for mobile applications and consequently are slower than expected on GPU.\\[1pt]
  \textsuperscript{3} FP8 performance numbers are not available for Nemotron and LLaDA. FP8 quantization provides an upper-bound speedup of $2\times$, with typical gains closer to $1.5\times$ depending on model architecture.\\[1pt]
  \textcopyright\, \the\year{} Google DeepMind. All rights reserved%
}

\newcommand{\sdrl}{\texttt{{\small SD$\cdot$RL}}\xspace}
\newcommand{\sdrlfootnotesize}{\texttt{{\footnotesize SD$\cdot$RL}}\xspace}

\newcommand{\enc}{\operatorname{Encoder}_{\theta}}
\newcommand{\dec}{\operatorname{Decoder}_{\theta}}
\newcommand{\softmax}{\operatorname{Softmax}}
\newcommand{\unif}{\operatorname{Unif}}
\newcommand{\Vcal}{\mathcal{V}}

\newcommand{\step}{\operatorname{Step}}

\begin{abstract}
We introduce DiffusionGemma, an experimental open-weight language model that uses discrete diffusion to generate text at exceptionally high speed. Rather than decoding one token at a time, DiffusionGemma iteratively refines blocks of 256 tokens in parallel, avoiding the sequential decoding bottleneck of conventional autoregressive (AR) large language models. Instead of training from scratch, we obtain DiffusionGemma by fine-tuning the mixture-of-experts Gemma 4 model with 3.8B activated and 25.2B total parameters. Our compute-efficient two-stage training pipeline uses fewer than 10\% of the starting AR model's total training token budget. The first stage uses supervised fine-tuning to teach bidirectional denoising, while the second stage combines reinforcement learning with sampler distillation to jointly improve generation quality and inference efficiency. DiffusionGemma establishes a new Pareto frontier for the trade-off between generation speed and model capability. Averaged across our full evaluation suite, it generates around 20 tokens per forward pass and achieves roughly 1,500 output tokens per second on a single NVIDIA H100 GPU, which is substantially faster than AR models even with state-of-the-art speculative decoding. DiffusionGemma also retains the starting model's support for thinking mode, multimodal inputs, and long contexts. Despite diffusion fine-tuning, it remains capable of AR generation with only minor performance degradation, suggesting a path toward hybrid diffusion-AR decoding.
\end{abstract}

\begin{document}

\maketitle

\begin{figure}[!b]
    \vspace{-4pt}
    \centering
    \includegraphics[width=0.85\linewidth, trim=0cm 0cm 0cm 0cm, clip]{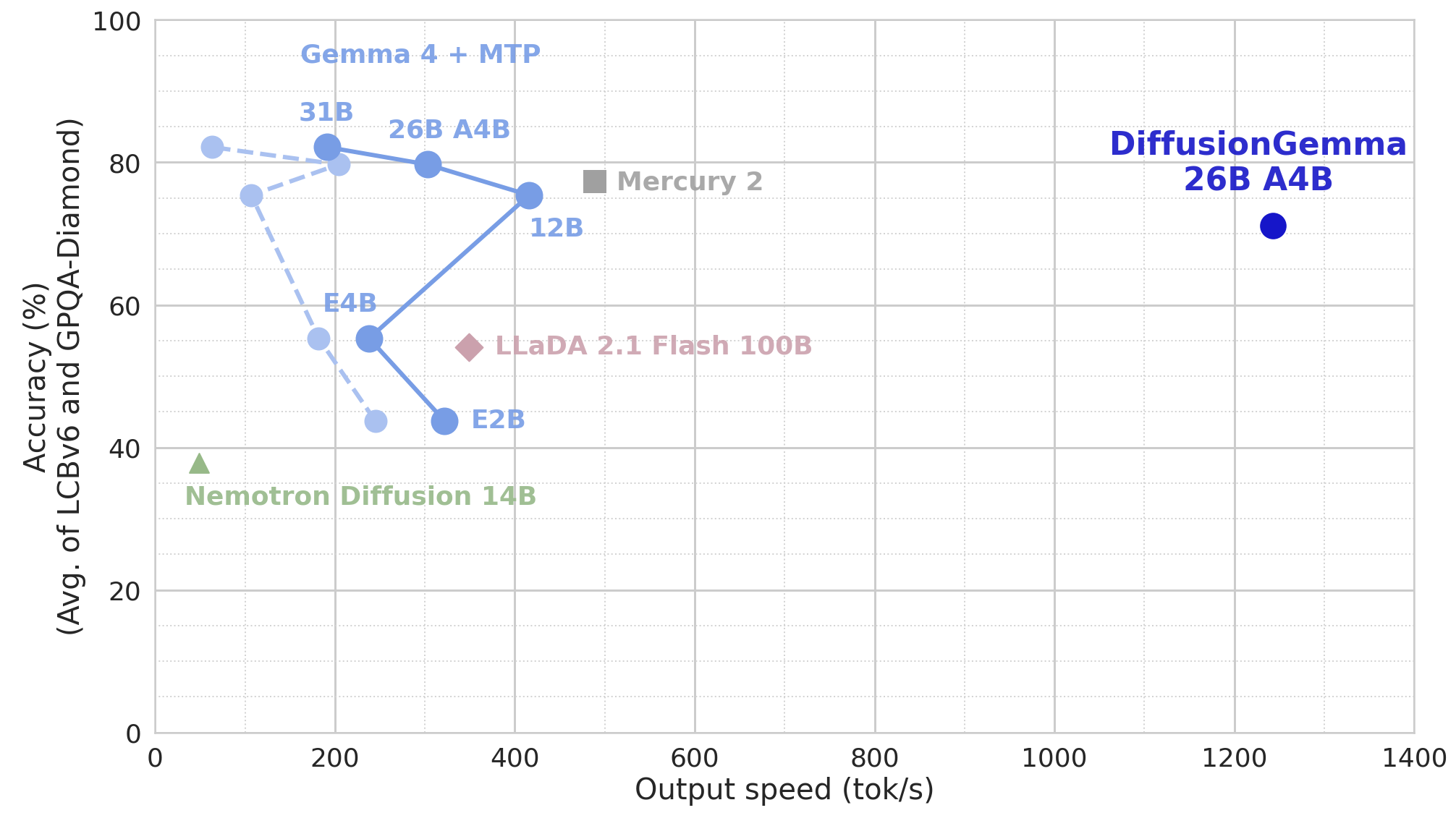}
    \caption{\textbf{Pareto plot of quality versus output decoding speed comparing DiffusionGemma to the Gemma 4 model family and other diffusion models.} Quality and output speed are calculated as the average across GPQA-Diamond and LiveCodeBench-v6 for all models. Gemma 4 and DiffusionGemma output speeds are measured on a single H100 (FP8, batch size~1).\textsuperscript{2} Nemotron~14B is measured on a single H100 (bfloat16, batch size~1). LLaDA~2.1 Flash~100B is measured on 8$\times$ NVIDIA~B200 GPUs (bfloat16, batch size~1).\textsuperscript{3} Mercury 2 was measured via the public OpenRouter API with \texttt{high} reasoning effort (see Appendix~\ref{sec:competitor-eval-methodology} for speed estimation). The light blue dots and dashed line represent the Gemma 4 family of models \emph{without} MTP.
    }
    \label{fig:final-pareto}
\end{figure}

\newpage

\section{Introduction}

Autoregressive (AR) models dominate the current Large Language Model (LLM) landscape, but their strict left-to-right, token-by-token generation creates a problematic memory bottleneck. When serving many requests simultaneously, one can achieve acceptable throughput by batching, but serving single or low-concurrency requests is fundamentally memory-bound: time spent transferring model weights and context KV cache from memory to the accelerator far exceeds time spent on actual computation. This leaves the accelerator's compute units under-utilized and limits per-user generation speed. Speculative decoding can improve utilization by drafting a candidate sequence, typically 8 tokens, from a small ``drafter'' model and feeding the draft to the LLM for verification~\citep{leviathan2023fast,chen2023accelerating,xia-etal-2024-unlocking}. With a draft length of 8 this can produce 3-6 tokens per forward pass (TPF)~\citep{li2026eagle}. However, the draft-then-verify paradigm is still limited: on one hand, AR drafters are bottlenecked by sequential generation; on the other hand, parallel drafters---building on early blockwise decoding~\citep{stern2018blockwise} and lookahead strategies~\citep{zhao2024lookahead}---exhibit declining acceptance rates at later draft positions~\citep{cheng2026dspark}.

Text diffusion circumvents this bottleneck by predicting entire blocks of tokens simultaneously (Figure~\ref{fig:text_diffusion_sample}), effectively shifting execution from a memory-bound regime toward a compute-bound one. Recently, there has been a surge of interest in text diffusion \citep{deschenaux2024promises}, with models like Gemini Diffusion \citep{deepmind2025geminidiffusion}, Mercury \citep{labs2025mercury}, LLaDA \citep{nie2026large,bie2025llada2}, Seed Diffusion \citep{song2025seed}, and Nemotron-Labs-Diffusion \citep{fu2026nemotronlabsdiffusion}. However, the current landscape forces a stark compromise between speed, intelligence, and accessibility. Some models, e.g., Gemini Diffusion and Mercury, are locked behind proprietary APIs. On the other hand, existing open-weights alternatives either exhibit limited reasoning capabilities and multimodal understanding, or fail to deliver on the extreme latency benefits promised by the diffusion technology, or both. Until now, there has been no text diffusion model that is highly intelligent, exceptionally fast, and openly accessible.

\begin{figure}[t!]
    \centering
    \includegraphics[width=1.0\linewidth,trim=0cm 6cm 0cm 4cm,clip]{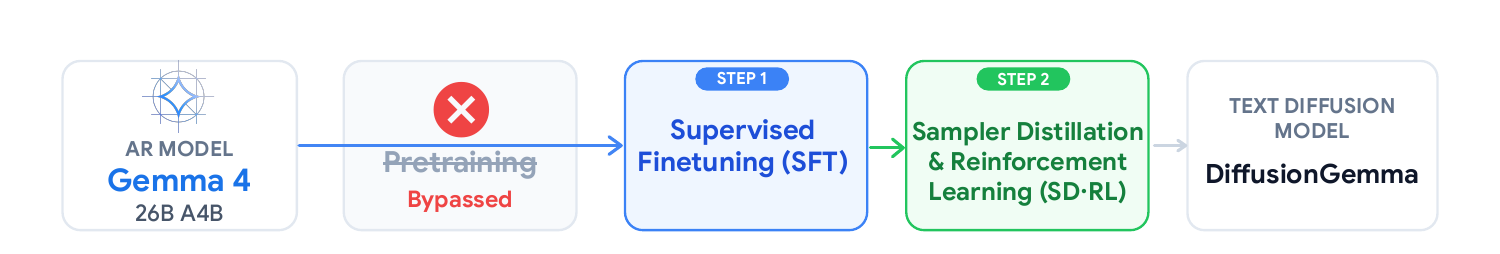}
    \caption{\textbf{Overview of our two-stage training pipeline that converts an autoregressive model (Gemma 4 26B A4B) into a text diffusion model (DiffusionGemma).} Initialized from the AR model weights, the model first undergoes SFT that adapts it to discrete text diffusion and bidirectional attention across 256-token canvases. This is followed by an online sampler distillation and reinforcement learning phase, which jointly maximizes reward-driven generation quality and compresses the denoising steps to unlock ultra-low latency.
    }
    \label{fig:recipe}
\end{figure}

We introduce DiffusionGemma to bridge this gap. A finetuned text diffusion variant of the Gemma~4 26B A4B mixture-of-experts (MoE)
model \citep{gemmateam2026gemma4technicalreport}, the model establishes a new Pareto frontier in the intelligence-to-speed trade-off for generative text modeling (Figure~\ref{fig:final-pareto}). Notably, it extends the speed frontier beyond both prior text diffusion models and the Gemma 4 AR family across all parameter scales (from E2B up to 31B), even when the AR models are equipped with multi-token prediction~\citep[MTP,][]{google2026mtpgemma4, deepseekai2024deepseekv3}, a state-of-the-art speculative decoding technique. To unlock this new speed-to-intelligence frontier, DiffusionGemma maximizes algorithmic efficiency by outputting blocks of 256 tokens simultaneously in around 12 forward passes. Put differently, DiffusionGemma generates on average 20 TPF---a step-change improvement over the $\sim$3-6 TPF that can be achieved by state-of-the-art speculative decoding methods. This massive reduction in total forward passes directly offsets the computational cost of the diffusion model's heavier individual forward passes and as a result, DiffusionGemma yields generation speeds of around 1,500 tokens per second (TPS) on a single NVIDIA H100 GPU. Furthermore, early benchmarking by third-party inference providers such as \citet{unsloth} demonstrates these speeds can scale up to 2,000 TPS on the NVIDIA RTX 6000.

To avoid the prohibitive computational cost of pretraining it is common practice to initialize text diffusion models from pretrained AR models \citep{han2024transfer,Gong2024ar2diff}. We warm-start DiffusionGemma from the final post-trained, publicly released weights of the Gemma 4 26B A4B MoE model~\citep{gemmateam2026gemma4technicalreport}. By adopting the same transformer backbone, we efficiently repurpose the AR weights to support both causal attention for context encoding and bidirectional attention for the diffusion process while inheriting their full capabilities. We employ a two-stage training pipeline (see Figure~\ref{fig:recipe}) that utilizes less than 10\% of the AR model’s total training tokens:

\begin{enumerate}[topsep=0pt]
    \item \textbf{Supervised fine-tuning (SFT)}: We first run an SFT phase to adapt the model to attend to a context of clean tokens as well as denoising a block of 256 noisy tokens (as dictated by the diffusion process) with bidirectional attention across the block.
    \item \textbf{Sampler distillation and reinforcement learning (\sdrl)}: We then apply an online learning phase that simultaneously improves generation quality (by maximizing rewards) and unlocks ultra-low latency (by substantially reducing the number of forward passes).
\end{enumerate}

By preserving features of its AR starting point, DiffusionGemma exhibits strong multimodal understanding, long-context capabilities, and thinking mode, enabling it to generate reasoning traces prior to responding. These are hallmarks of current frontier systems \citep{geminiteam2025gemini}. Although adapting the model to the diffusion regime introduces a performance penalty compared to the original AR baseline, the massive gains in inference speed offer an entirely new operating point for latency-critical applications. DiffusionGemma retains the ability to generate text autoregressively. This \textit{dual mode} opens up the possibility to route requests dynamically based on latency constraints and task complexity, as well as to employ hybrid decoding approaches.

\begin{table}[t!]
\centering
\begin{minipage}[t]{0.43\textwidth}
\centering
\begin{tabular}[t]{ll}
\toprule
Total & 25.2B \\
Activated & 3.85B \\
Vision Encoder & 550M \\
Embedder & 740M \\
Self-Conditioning & 7.8M \\
Active / Total Experts & 8 / 128 \\
 &  + 1 shared \\
\bottomrule
\end{tabular}
\caption{\textbf{Parameter counts.} DiffusionGemma's architecture is a mixture-of-experts transformer with a vocabulary of 262k tokens. The total number of activated parameters does not include the vision encoder. It includes an additional MLP block for the purpose of self-conditioning.}
\label{tab:parameter-counts}
\end{minipage}\hfill
\begin{minipage}[t]{0.53\textwidth}
\centering
\begin{tabular}[t]{ll}
\toprule
Canvas Length & 256 \\
Sampler Maximum Denoising Steps & 48 \\
Adaptive Stopping Entropy Threshold & 0.005 \\
Token Selection Entropy Threshold & 0.1 \\
Temperature Schedule (Linear) & 0.8 $\rightarrow$ 0.4 \\
\bottomrule
\end{tabular}
\caption{\textbf{Diffusion denoising and recommended sampler hyperparameters.} In practice, the average sampler denoising steps is 12 with adaptive stopping, depending on the task.}
\label{tab:diffusion-hyperparams}
\end{minipage}
\end{table}

\paragraph{Open-weights release.} We release DiffusionGemma with open weights under a permissive Apache 2.0 license, aiming to democratize access to state-of-the-art text diffusion technology. By providing full, unrestricted access to the model parameters, we hope to empower a diverse ecosystem of researchers, developers, and practitioners. For the research community, this transparent release provides a robust, white-box baseline to deeply probe the underlying mechanics of discrete diffusion, accelerating foundational research and pushing the theoretical frontier of text generation. For developers, the liberal licensing removes friction, ensuring seamless integration into both experimental prototypes and commercial applications without restrictive usage barriers.

Crucially, DiffusionGemma's highly efficient architecture makes it an ideal foundation for rapid, domain-specific adaptation. By dramatically lowering the computational barrier to entry, the community can easily finetune ultra-fast, task-specific models tailored to their own unique use cases, even in environments with constrained compute budgets. This capacity for lightweight, rapid iteration has catalyzed immediate community adoption. Despite the model being available for only a few weeks at the time of writing, it is already serving as the engine for specialized downstream applications. These rapidly emerging use cases showcase the model's versatility across highly diverse and demanding domains, spanning from multilingual automatic speech recognition~\citep{interfaze2026diffusion} to interactive radiology report drafting in healthcare~\citep{van2026discrete}.

\paragraph{Outline.}
The remainder of this technical report is organized as follows. Section~\ref{sec:background} formalizes the discrete diffusion framework and the theoretical foundations of our approach. Section~\ref{sec:architecture} details the DiffusionGemma architecture, including the bidirectional decoding mechanism and sampling algorithm. We then describe the two stages of our training pipeline: the SFT phase in Section~\ref{sec:sft}, followed by our online learning phase combining sampler distillation and reinforcement learning in Section~\ref{sec:sdrl}. Section~\ref{sec:inference_optimizations} breaks down our low-level inference optimizations. Section~\ref{sec:experiments} presents experimental results. Section~\ref{sec:downstream_sft} provides instructions and a practical example of finetuning DiffusionGemma for downstream applications. In Section~\ref{sec:advantages}, we showcase some practical advantages of text diffusion. Finally, we discuss limitations and known issues in Section~\ref{sec:limitations}, before concluding in Section~\ref{sec:conclusion}.

\section{Generative Text Modeling with Discrete Diffusion} \label{sec:background}

Historically, LLMs have treated text generation as a strictly sequential process~\citep{bengio2003neural, mikolov2010recurrent, graves2013generating, sutskever2014sequence, Vaswani2017, gu2018nonautoregressive}. Classical AR language modeling factorizes the joint probability of a sequence $x$ of length $L$ into an exact product of conditional probabilities: $p(x) = \prod_{i=1}^L p(x^i \mid x^{<i})$, where $i$ indicates the token position in the sequence. While theoretically rigorous, this single-token factorization fundamentally limits generation speed on modern hardware accelerators due to severe memory-bandwidth constraints as well as strictly preventing the model from bidirectionally revising tokens based on future tokens~\citep{leviathan2023fast, savinov2022sundae}.

Diffusion models approximate the joint distribution $p(x)$ by factorizing the generative process over a sequence of noise levels (a Markov chain) rather than strictly left-to-right spatial positions~\citep{sohl2015deep, ho2020denoising, song2020score, holderrieth2025}, building upon foundational work in score-matching and continuous-time flow models~\citep{song2019generative, song2021denoising, song2021maximum, meng2022sdedit, lipman2023flow, debortoli2021diffusion}. During training, a forward process gradually corrupts clean data into random noise; a neural network is trained to learn the reverse denoising process. During inference, the model generates full sequences of data by iteratively refining the output sequence in parallel, starting from random noise.

Adapting diffusion to text requires handling the discrete nature of language~\citep{yi2024diffusion}. Early approaches adapt standard Gaussian diffusion to text by mapping discrete tokens into a continuous embedding space. These models differ primarily in how they map back to text: architectures like Diffusion-LM \citep{li2022diffusionlm} and SED \citep{strudel2022self} produce continuous vectors that have to be forcibly projected or ``rounded'' back to discrete tokens, whereas methods like CDCD \citep{dieleman2022continuous} maintain continuous processes but predict categorical logits directly to sample tokens. Regardless of the decoding mechanism, these formulations face theoretical and geometric challenges that ultimately limit their effectiveness compared to native discrete diffusion models. In particular, hard rounding operations fundamentally break exactness of diffusion likelihood bounds established by continuous-time and variational formulations \citep{kingma2021variational} and the valid vocabulary tokens occupy an infinitesimally small fraction of a high-dimensional latent space. As a result, the reverse generative process often drifts into empty, ``meaningless'' regions of the space; when rounded to nearest token, these degenerate embeddings map to arbitrary, unrelated tokens and produce incoherent text \citep{shabalin2025tencdm, meshchaninov2025cosmos}. To combat this spatial drift, methods have been developed to rely on per-step discretization mechanisms, but these generally underutilize the continuous space and restrict generative flexibility \citep{hu2026elf}.

Discrete diffusion has emerged as a highly effective alternative \citep{austin2021structured, sahoo2024simple, lou2024discrete, zhao2024unified}. This paradigm generalizes the single-step corruption heuristics of early bidirectional models into formal, multi-step Markov processes. The probabilistic transitions defined between categorical states in modern diffusion, such as absorbing mask states and multinomial distributions across the vocabulary, are direct descendants of masking and random token swapping introduced by BERT \citep{devlin2019bert, liu2019roberta, sanh2019distilbert}, as well as the plausible token replacements of ELECTRA \citep{clark2020electra}. By operating directly on discrete states rather than continuous vectors, discrete diffusion avoids embedding projection mismatches entirely and offers better theoretical grounding for categorical transitions \citep{gat2024discrete}. DiffusionGemma builds upon this lineage of discrete diffusion to ensure high-fidelity token generation.

\subsection{Probability Paths and Denoising}

To formalize our discrete diffusion framework, we adopt the continuous-time Markov chain (CTMC) approach established in recent literature \citep{campbell2022continuous, campbell2024generative, gat2024discrete}. Let $\Vcal$ define a categorical token vocabulary of size $V$. We refer to the sequence of tokens undergoing iterative refinement as the \emph{canvas}, with a length of $C$. A realization of this canvas at time $t \in [0, 1]$ is denoted by the vector $x_t \in \Vcal^C$. We construct a marginal probability path to smoothly interpolate between a clean data distribution $p$ at time $t=0$ and a fully corrupted source canvas at time $t=1$. At the start of this path, the clean tokens are defined as $x_0 \sim p(\cdot)$, which ultimately transition into the fully corrupted state $x_1 \sim \unif(\Vcal^C)$. Here, $\unif(\Vcal^C)$ denotes the uniform prior over the joint state space, where each of the $C$ tokens is independently uniformly distributed over $\Vcal$. This corruption process parallels the explicit transition matrices—such as uniform noise or absorbing masking—commonly utilized in order-agnostic discrete diffusion models \citep{hoogeboom2022autoregressive}. By operating in continuous time, this framework bypasses the sequential, left-to-right decoding bottleneck of traditional AR models, allowing tokens across the entire canvas to be processed and refined in parallel.

\begin{figure*}[!t]
    \centering
    \includegraphics[width=1.0\linewidth]{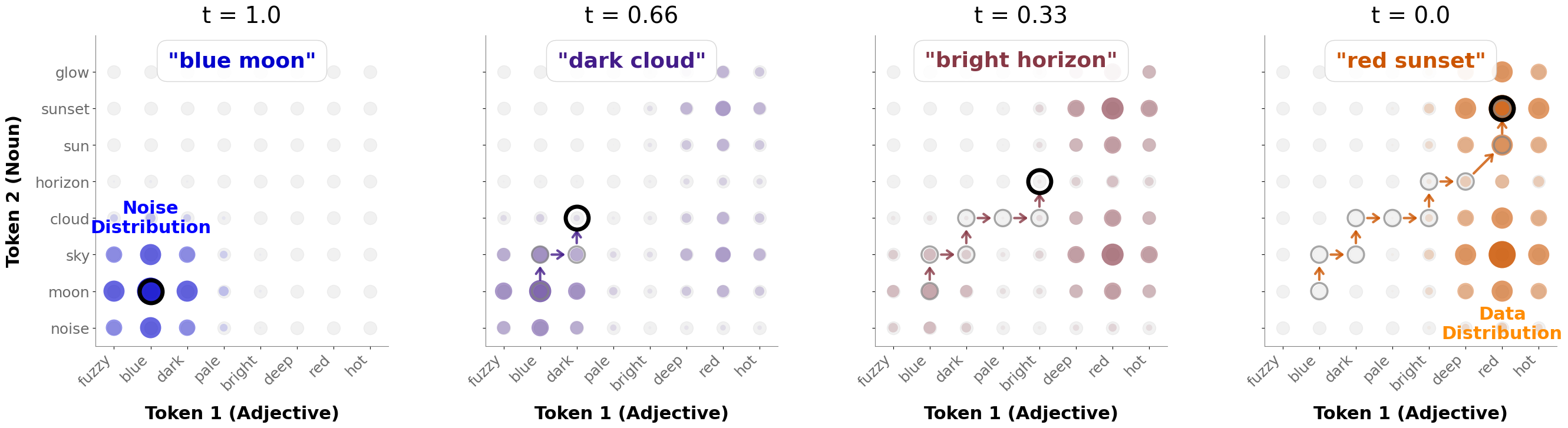}
    \caption{\textbf{Stylized example of discrete diffusion probability paths and parallel sampling trajectory.} For illustrative purposes, the state space uses distinct, token-specific vocabularies: adjectives on the horizontal axis and nouns on the vertical axis. As time moves backward from $t = 1.0$ (noise) to $t = 0.0$ (data), the marginal distribution smoothly interpolates, concentrating mass away from the uniform noise distribution and onto valid data modes. Black circles track the discrete jump transitions of an individual sequence realization (from ``blue moon'' at $t=1.0$ to ``red sunset'' at $t=0.0$), demonstrating how parallel canvas dimensions coordinate non-autoregressively over time.}
    \label{fig:discrete_diff}
\end{figure*}

\subsubsection{The Forward Process}
The forward process dictates the transition from clean text tokens to uniformly distributed tokens. Conditioned on a fixed clean starting canvas $x_0$, the forward transition probability path factorizes independently over each token coordinate $i$:
\begin{equation}\label{eq:factorized_path_pmf}
    \mathbb{P}(X_t=x_t \mid X_0 = x_0) =
    \prod_{i=1}^C \left[ \kappa_t \delta(x_t^i, x_0^i) + (1 - \kappa_t) \frac{1}{V} \right],
\end{equation}
where $\kappa_t \in [0,1]$ is a smoothly varying, monotonically decreasing noise schedule from $\kappa_0 = 1$ to $\kappa_1 = 0$, and $\delta$ represents the Kronecker delta. In practical terms, as time $t$ progresses toward $1$, each token is increasingly likely to be replaced by a token sampled uniformly at random from the vocabulary \citep{hoogeboom2021argmax, austin2021structured}. We use this closed-form forward process to generate training data from clean text.

\subsubsection{The Backward Denoising Process}
To reconstruct data from noise, we learn a reverse process to undo this categorical corruption. Discrete flow matching theory shows that to perfectly reverse the forward trajectory, we need to infer the conditional distribution of the original uncorrupted tokens given a corrupted state \citep{campbell2024generative, gat2024discrete, ou2025absorbing}. For a given realization $X_t = x_t$, stepping backward in time by a small increment $\Delta t$ is governed by some transition mapping, which we denote $\step$, which outputs the probability distribution for the next intermediate step\footnote{As a typical example, the function $\step$ can be instantiated as a discrete Euler update step introduced in \citet{gat2024discrete}, in which $
\mathbb{P}(X_{t-\Delta t}^i = v \mid X_t = x_t) \approx \delta(v, x_t^i) - \Delta t \frac{\dot{\kappa}_t}{1 - \kappa_t} \left[ \mathbb{P}(X_0^i = v \mid X_t = x_t) - \delta(v, x_t^i) \right]$ with $\kappa_t \in [0, 1]$.}:
\begin{equation}\label{eq:transition-mapping}
\mathbb{P}(X_{t-\Delta t} = \cdot \mid X_t = x_t) \approx \operatorname{Step}\Big(x_t, \mathbb{P}(X_{0} = \cdot \mid X_t = x_t)\Big).
\end{equation}
To compute this update, we need the true posterior distribution of clean tokens given a noisy state, $\mathbb{P}(X_{0}^i = v \mid X_t = x_t)$. We approximate this posterior with a neural network $p_\theta(v \mid x_t)$. During generation, we use this approximation to sample the next token state: $x_{t-\Delta t} \sim \operatorname{Step}(x_t, p_\theta(\cdot \mid x_t))$.

Figure \ref{fig:discrete_diff} presents a highly stylized example to illustrate this reverse sampling trajectory within a simplified two-token canvas ($C=2$). As time runs backward from $t=1.0$ to $t=0.0$, the model smoothly shifts probability mass away from a uniform noise distribution (e.g., around ``blue moon'') toward valid data modes. Concurrently, individual sequence coordinates undergo continuous-time jump transitions---visualized by the step-by-step path from ``blue moon'' at $t=1.0$, to ``dark cloud'' at $t=0.66$, and finally landing on a high-probability mode like ``red sunset'' at $t=0.0$. This highlights how parallel dimensions coordinate over time without requiring sequential left-to-right generation.

Because this denoising process is factorized, it assumes conditional independence during individual reverse steps. A newly updated token at position $i$ is conditioned globally on the current noisy canvas~$x_t$, but it cannot see the simultaneous sampling choices made at other positions ($j \neq i$). This structural tradeoff can occasionally introduce local inconsistencies or conflicting grammatical predictions. In the next section, we counteract these uncoordinated mechanics directly through our \emph{self-conditioning architecture} and \emph{entropy-bounded sampling} strategies.

\section{The DiffusionGemma Architecture} \label{sec:architecture}

The DiffusionGemma architecture functions as an encoder-decoder transformer \citep{Vaswani2017} with \emph{shared weights}~$\theta$. Rather than pretraining a diffusion model from scratch, we initialize our model with the publicly released Gemma 4 26B A4B MoE checkpoint~\citep{gemmateam2026gemma4technicalreport}. Tying our architecture to an existing AR backbone is a pragmatic choice that trades off absolute generation quality for a substantial reduction in the computational cost of training. This initialization also allows us to inherit the base model's advanced features---such as its extended context window and native multimodal understanding. Moreover, our experiments demonstrate that our final model weights retain the ability to be sampled from autoregressively (see Table~\ref{tab:model_comparison}).

The remainder of this section details how DiffusionGemma uses this architecture to generate text at inference time. We break this process down into three components: the block-autoregressive decoding strategy for handling long sequences (Section~\ref{sec:block-ar}), the general framework for denoising individual canvases (Section~\ref{sec:denoising}), and a specific entropy-bounded sampler (Section~\ref{sec:sampling}).

\begin{figure}[t!]
    \centering
    \includegraphics[width=0.9\textwidth]{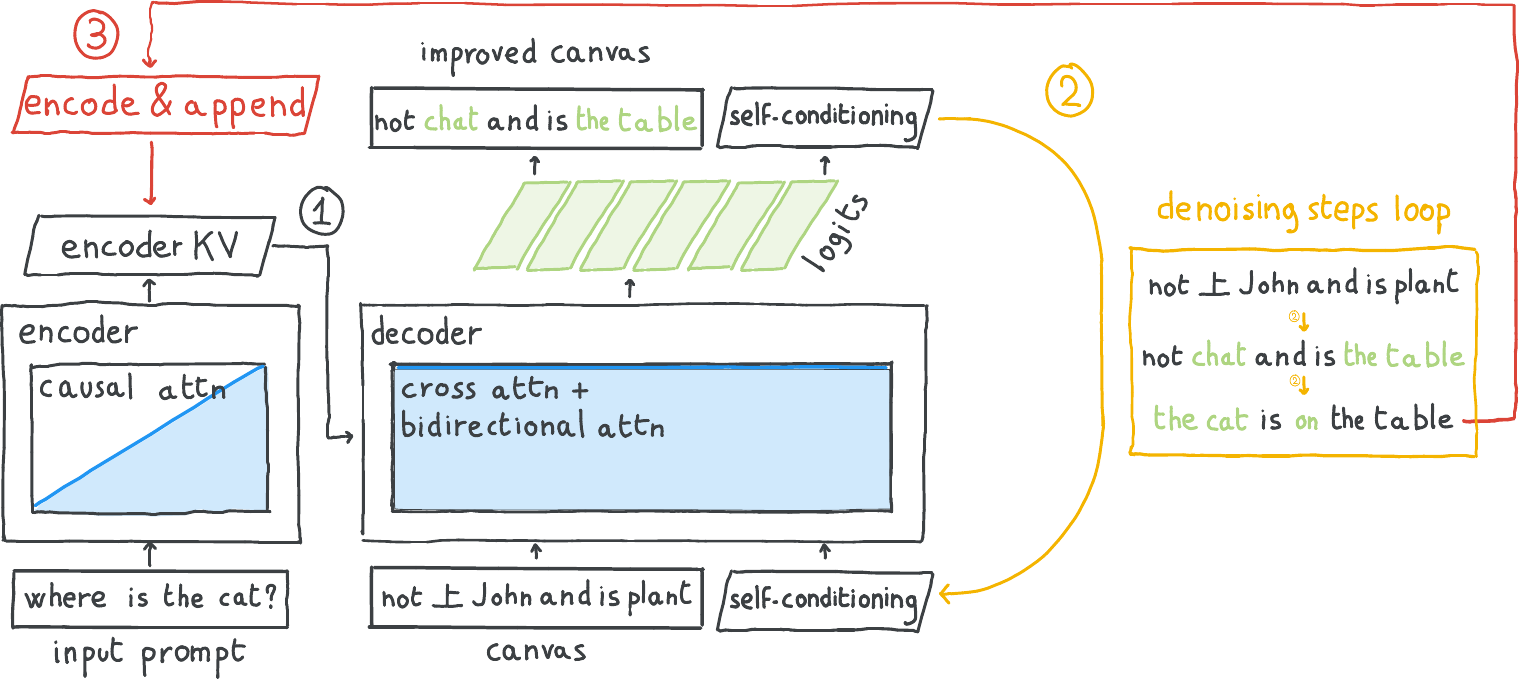}
    \caption{\textbf{The DiffusionGemma generation pipeline.} The process consists of three main stages: 1) \textbf{Context encoding:} The input prompt is processed by the causal encoder to initialize the Key-Value (KV) cache. 2) \textbf{Denoising loop:} A noisy canvas is iteratively refined by the decoder, using bidirectional attention across the canvas and cross-attention to the KV cache, until the text is fully denoised. 3) \textbf{Encode \& append:} The finalized clean canvas is passed back through the causal encoder and appended to the KV cache, setting the context for the next block of tokens.}
    \label{fig:encoder-decoder}
\end{figure}

\subsection{Block-Autoregressive Generation}\label{sec:block-ar}

The discrete flow matching formulation introduced in Section~\ref{sec:background} operates on fixed-length sequences. To generate open-ended text, we use a block-AR generation strategy. The model denoises a canvas of 256 tokens at a time, and once a canvas is fully denoised, it is committed to the sequence history, and the model begins denoising the next canvas.

\paragraph{KV cache initialization.} As illustrated in Figure~\ref{fig:encoder-decoder}, the first step in block-AR generation is to encode the context, $h \in \Vcal^L$, into a KV cache $H$. The context includes system instructions and user prompts of maximum token count $L$.\footnote{In practice, $h$ also includes interleaved multimedia embeddings which are not part of the token vocabulary.} We write the encoding as
\begin{equation}
    \label{eq:encoder}
    H = \enc(h),
\end{equation}
where $\enc$ is the transformer forward pass with weights $\theta$ and causal attention masking.

\paragraph{Canvas denoising conditioned on KV cache.} By cross-attending to the KV cache $H$, canvas generation can be conditioned on system instruction, user inputs and past responses. Canvas generation starts from a canvas of uniformly random tokens and is iteratively denoised using the process described in Sections~\ref{sec:denoising} and~\ref{sec:sampling}. Once a canvas is fully denoised,\footnote{A canvas is fully denoised either by reaching $t=0$ or by the adaptive stopping condition described in Section~\ref{sec:sampling}.} denoted by $\hat{x}_0$, its keys and values are appended to the KV cache (depicted in red in Figure~\ref{fig:encoder-decoder}):
\begin{equation}
    \label{eq:updated_kv_cache}
    H \gets H \oplus \enc(\hat{x}_0).
\end{equation}
We repeat this denoising process for subsequent canvases until the model generates a special token marking the end of the model's turn, after which all canvases $\hat{x}_0$ are concatenated into a final response. Since we use causal attention in the encoder, it lets the encoder update the KV cache by appending only the newest canvas. This constitutes an architectural inversion of standard encoder-decoder models such as BART \citep{lewis2020bart} or T5 \citep{raffel2020exploring}. Whereas those models utilize a bidirectional encoder for context and a causal decoder for generation, our approach utilizes a causal encoder for the sequence history and a bidirectional decoder for the diffusion-based canvas generation. This causal encoding prevents early context tokens from attending to the latest canvas, but it eliminates the need to re-encode the growing context from scratch, making our approach scalable to long reasoning generations and yielding a structural blend of diffusion and AR generation. This block-AR strategy was developed by our group in an unpublished June 2023 manuscript; analogous approaches to blockwise sequence modeling and KV caching were independently developed by~\citet{wu2025fastdllmv2}, \citet{arriola2025block}, and~\citet{deschenaux2026blockgen}.

\subsection{The Denoising Framework at Inference}\label{sec:denoising}

\paragraph{Canvas initialization.} To generate a canvas, we iteratively refine it over a maximum of $N$ denoising steps, with step size $\Delta t = 1 / N$. Let $x_t \in \Vcal^C$ denote the corrupted canvas at step $t$. We initialize the process at $t=1$ with a canvas $x_1$ of uniformly random tokens from the vocabulary $\Vcal$. 

\paragraph{The decoder forward pass.} At each step $t$, we apply the transformer with the shared weights $\theta$ as a decoder, written as $\dec$, to predict the probability distribution of the clean tokens. The decoder takes three inputs: the current noisy canvas $x_t$, the context KV cache $H$, and a continuous \textit{self-conditioning} signal $z_t \in \mathbb{R}^{C \times d}$ that feeds the model's previous predictions back into itself \citep{chen2023analog, strudel2022self, jo2026loopholing}. Using bidirectional attention across the canvas tokens and cross-attention to the KV cache, the decoder outputs the unnormalized logits $L_t$:
\begin{equation} \label{eq:denoising-forward-pass}
    L_t = \dec(x_t, z_t, H) \in \mathbb{R}^{C \times V}.
\end{equation}

\paragraph{The denoising iteration.} At each iteration, we compute the logits $L_t$ and evaluate the clean token probabilities $\hat{p}_0$, update the self-conditioning signal for the next step, and sample the refined canvas:
\begin{equation} \label{eq:denoising-iteration}
\begin{aligned} 
    \hat{p}_0 &= \softmax \big(L_t/\tau_t\big) \in [0, 1]^{C \times V}, \\
    z_{t-\Delta t} &= \operatorname{FFW}(\hat{p}_0 E) \in \mathbb{R}^{C \times d},
    \\
    x_{t-\Delta t} &\sim \operatorname{Step}(x_t, \hat{p}_0).
\end{aligned}
\end{equation}
Here, $E \in \mathbb{R}^{V \times d}$ is the token embedding matrix and $\operatorname{FFW}$ is a standard feedforward network. The time-dependent temperature $\tau_t > 0$ can sharpen the model's predictions before we calculate the final transition probabilities. A transition mapping $\step$, as described in Equation~\eqref{eq:transition-mapping}, computes the categorical distribution used to sample the updated canvas $x_{t-\Delta t}  \in \Vcal^C$, which, along with the new self-conditioning signal $z_{t-\Delta t}$, is then fed into the next denoising step.
Section~\ref{sec:sampling} details our specific choices for transition mapping $\step$ and the temperature $\tau_t$.

\paragraph{Multinomial diffusion.} Our denoising iteration (Equation~\ref{eq:denoising-iteration}) uses multinomial (or uniform) diffusion rather than masked diffusion \citep{hoogeboom2021argmax, austin2021structured}. Because all tokens can transition between one another, the model can continuously correct its own errors: tokens accepted during earlier denoising steps ($t' > t$) within the current canvas can still be revised. However, we note that tokens from previously generated canvases are permanently frozen.

\paragraph{Remark on interpretability.} Recent interpretability analyses demonstrate that while the self-conditioning signal $z_t$ introduces a continuous latent space injection into the sequence, these intermediate self-conditioning vectors map robustly to an interpretable token bottleneck, preserving the model's algorithmic transparency~\citep{engels2026how, asaria2026neither}.

\subsection{The DiffusionGemma Sampler}
\label{sec:sampling}

In contrast to AR generation, which relies on rigid heuristics like temperature, top-$p$, and top-$k$, diffusion modeling introduces a vastly richer sampling design space that includes discrete predictor-corrector mechanisms and multi-step solvers \citep{lezama2023predictor, deschenaux2026diffusion, liu2026nisampling, yao2026accelerating, ren2025fast}. Framing text generation as an iterative temporal process allows us to decouple the inference-time algorithm from the underlying architecture, granting fine-grained control over the trade-off between computational cost and generation quality. 

While we described the overall approach of canvas denoising in Section~\ref{sec:denoising}, this subsection details a specific instantiation: the entropy-bounded sampler \citep{ben2026accelerated} with temperature annealing and adaptive stopping (Algorithm~\ref{alg:sampling}). While this is our default and recommended sampler, DiffusionGemma is modular and is not strictly bound to this exact sampling configuration for high-quality inference.

\begin{algorithm}[t!]
\caption{DiffusionGemma Sampling with Entropy-Bounded Denoising
and Adaptive Stopping}
\label{alg:sampling}
\small
\begin{algorithmic}[1]
\Require Context $h$; maximum number of denoising steps $N=48$
\Require Entropy budget $b=0.1$;
         stopping threshold $e_{\mathrm{stop}}=0.005$
\Require Temperature-schedule endpoints
         $\tau_{\max}=0.8$ and $\tau_{\min}=0.4$

\Statex \textbf{Initialization}
\State $H \gets \enc(h)$
    \Comment{Causal-attention KV cache}
\State $\Delta t \gets 1/N$
\State $x_1 \sim \unif(\Vcal^C)$,
    \quad $z_1 \gets \boldsymbol{0}$
\State $\hat{x}_0^{\mathrm{prev}} \gets \bot$

\Statex \textbf{Denoising}
\For{$n=1,\ldots,N$}
    \State $t \gets 1-(n-1)\Delta t$

    \State $L_t \gets \dec(x_t,z_t,H)$,
        \quad $L_t\in\mathbb{R}^{C\times V}$
    \State $\hat{x}_{0\mid t} \gets \arg\max(L_t)$
        \Comment{Current deterministic prediction}

    \State $\tau_t \gets
        (\tau_{\max}-\tau_{\min})t+\tau_{\min}$
    \State $\hat{p}_0 \gets
        \softmax(L_t/\tau_t)$

    \State $e^i \gets
        \operatorname{Entropy}(\hat{p}_0^i)$
        for all $i\in\{1,\ldots,C\}$
    \State $\bar{e} \gets
        \dfrac{1}{C}\sum_{i=1}^{C}e^i$

    \If{$n>1
        \land \bar{e}\leq e_{\mathrm{stop}}
        \land \hat{x}_{0\mid t}=\hat{x}_0^{\mathrm{prev}}$}
        \State \Return $\hat{x}_{0\mid t}$
            \Comment{Adaptive stopping}
    \EndIf

    \State $\hat{x}_0^{\mathrm{prev}}
        \gets \hat{x}_{0\mid t}$

    \If{$n<N$}
        \State $z_{t-\Delta t} \gets
            \operatorname{FFW}(\hat{p}_0E)$
            \Comment{$E$ is the embedding matrix}

        \State $x_D \sim \hat{p}_0$

        \State Let $(\pi_1,\ldots,\pi_C)$ order positions
            by non-decreasing entropy
            \Comment{Break ties by position index}
        \Statex \hspace{\algorithmicindent}
            $e^{\pi_1}\leq e^{\pi_2}\leq\cdots\leq e^{\pi_C}$

        \State $k \gets
            \displaystyle\max\left\{
                m\in\{1,\ldots,C\}:
                \sum_{j=1}^{m-1}e^{\pi_j}\leq b
            \right\}$
        \State $U \gets \{\pi_1,\ldots,\pi_k\}$

        \State For each $i\in\{1,\ldots,C\}$, set
        \Statex \hspace{\algorithmicindent}
            $\displaystyle
            x_{t-\Delta t}^i \gets
            \begin{cases}
                x_D^i,
                    & i\in U,\\
                \tilde{x}^i,\quad
                    \tilde{x}^i\sim\unif(\Vcal),
                    & i\notin U.
            \end{cases}$
    \EndIf
\EndFor

\State \Return $\hat{x}_{0\mid t}$
\end{algorithmic}
\end{algorithm}

\paragraph{Entropy-bounded token refinement.} After producing the marginal distributions $p_{\theta}(x_0 \mid x_t, z_t, H)$ with the denoiser and applying temperature scaling (detailed below), tokens are sampled from the resulting probability distribution. We employ an entropy-bounded sampler \citep{ben2026accelerated}, whereby tokens are accepted in rank order from lowest to highest entropy (similar to the MaskGIT decoding scheme by \citet{Changetal2022}), ensuring that their mutual information bound remains strictly below a predefined error tolerance threshold ($b=0.1$). Once the threshold is attained, all other tokens are renoised uniformly at random, maintaining these uncommitted positions as a uniform prior to force local exploration during the next forward pass.

\paragraph{Temperature annealing.} To balance rate of convergence and linguistic diversity, token probabilities are artificially sharpened via tempering. A temperature $\tau_t < 1$ is annealed linearly from an initial value of $\tau_{\max}=0.8$ down to $\tau_{\min}=0.4$ across the fractional denoising timescale $t \in [0, 1]$. While static temperature adjustments are traditionally used to truncate the unreliable tail of token distributions and prevent text degeneration, this dynamic annealing ensures that the model explores diverse token possibilities in early, highly-noised states and aggressively commits to high-confidence sequences as the semantic structure crystallizes \citep{xiao2024edt}.  

\begin{figure*}[!t]
    \centering
    \includegraphics[width=1.0\linewidth]{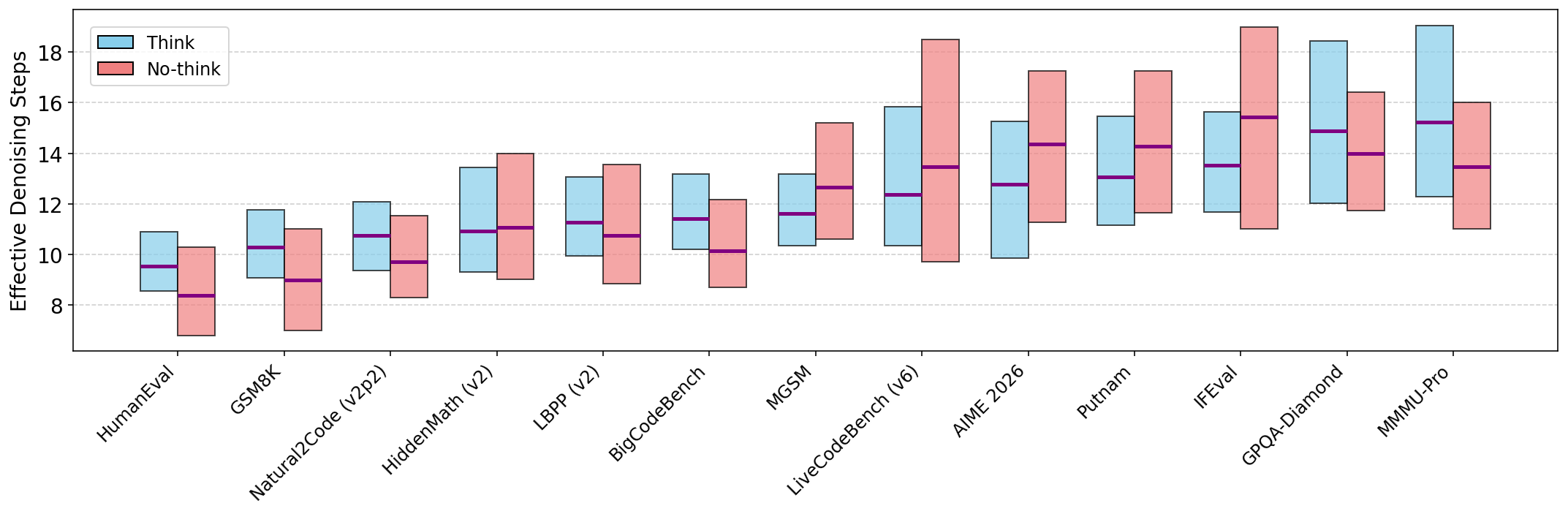}
    \caption{\textbf{Adaptive stopping enables DiffusionGemma to dynamically adjust its number of denoising steps to task complexity and domain.} We report median, first and third quartiles of the effective denoising steps (Equation~\ref{eq:effective_denoising_steps}). See Table~\ref{tab:dg_td_latency} for full benchmarks and latency metrics.}
    \label{fig:boxplot_evals_effective_denoising_steps}
\end{figure*}

\paragraph{Adaptive stopping heuristic.} To optimize inference efficiency, the sampler dynamically halts the denoising process based on the model's step-wise uncertainty, strictly capping iterations at $N$. This early termination is triggered when two conditions are simultaneously satisfied:
\begin{itemize}[topsep=0pt]
\item \textit{Confident predictions:} The mean predictive entropy across the entire canvas falls below a predefined threshold ($e_{\mathrm{stop}}=0.005$).
\item \textit{Stable predictions:} The deterministic sequence predictions (i.e., the most likely tokens) from two consecutive denoising steps are identical.
\end{itemize}
By successfully bypassing redundant refinement steps, this mechanism allows the model to dynamically scale its inference-time compute to the complexity of the prompt. As illustrated in Figure~\ref{fig:boxplot_evals_effective_denoising_steps}, instead of exhausting the maximum $N=48$ budget, the model averages approximately $12$ effective denoising steps (defined in Section~\ref{sec:metrics}) across the downstream evaluations shown in the figure. This yields a $4\times$ reduction in overall latency without sacrificing generation quality. Furthermore, the boxplots reveal distinct convergence profiles across domains: the model tends to deploy fewer steps for structured tasks like code and more for natural language. Moreover, it is not only the domain, but also the task complexity that dictates this behavior; for instance, harder code problems (e.g., LiveCodeBench) naturally require more steps to converge than easier ones (e.g., HumanEval).

\subsection{Inference Efficiency Metrics} \label{sec:metrics}

Because adaptive stopping turns the number of denoising steps into a random variable, we introduce a set of metrics to quantify denoising inference efficiency. Let $K$ be the total number of canvases generated, $N_k \leq N$ be the number of denoising steps executed for the $k$-th canvas, and $C_k \leq C$ be the number of valid tokens produced in that canvas (i.e., all $C$ tokens, or up to the first end-of-sequence token if one is generated). First, we define the Total Tokens as the sum of all valid tokens generated across the entire sequence:
\begin{equation}\label{eq:total_tokens}
\text{Total Tokens} \triangleq \sum_{k=1}^K C_k.
\end{equation}
We define the \textbf{Total Denoising Steps} as the absolute number of denoising steps executed across the entire generation:
\begin{equation}\label{eq:total_denoising_steps}
   \text{Total Denoising Steps} \triangleq \sum_{k=1}^{K} N_k.
\end{equation}
Since each denoising step incurs a fixed computational cost, the Total Denoising Steps is the primary driver of end-to-end generation latency. However, because this absolute metric inherently scales with the total sequence length, we also evaluate a normalized measure of model efficiency. We define the~\textbf{Effective Denoising Steps} as the token-weighted average of steps across all canvases:
\begin{equation}\label{eq:effective_denoising_steps}
   \text{Effective Denoising Steps} \triangleq \frac{1}{\text{Total Tokens}} \sum_{k=1}^K N_k C_k.
\end{equation}
We weight by the number of valid tokens to prevent the metric from being downwardly biased by the final canvas, which tends to require fewer denoising steps when it is only partially filled. Note that without adaptive stopping, the Effective Denoising Steps metric always reduces to the denoising budget $N$. Finally, we calculate the \textbf{Tokens Per Forward (TPF)}:
\begin{equation}\label{eq:tpf}
   \text{TPF} \triangleq \frac{\text{Total Tokens}}{\text{Total Denoising Steps} + K - 1},
\end{equation}
where the $K-1$ term accounts for the single additional forward pass required between canvases to encode the newly generated clean tokens and append their key-value pairs to the KV cache.

\subsection{Retained Autoregressive Capability} \label{sec:retained_ar_capability}

Because DiffusionGemma shares the exact same transformer architecture as Gemma 4, the final DiffusionGemma weights can be seamlessly loaded back into the original architecture to perform standard AR generation using causal attention, exactly as the base model does. As demonstrated in Section~\ref{sec:experiments}, Table~\ref{tab:model_comparison}, the model maintains robust capabilities in this setting; its performance scores in AR mode land squarely between those of DiffusionGemma's primary text diffusion mode and the baseline Gemma 4 checkpoint from which DiffusionGemma was initialized.

\section{Supervised Finetuning} \label{sec:sft}

\begin{figure}[!t]
    \centering
    \includegraphics[width=0.9\linewidth]{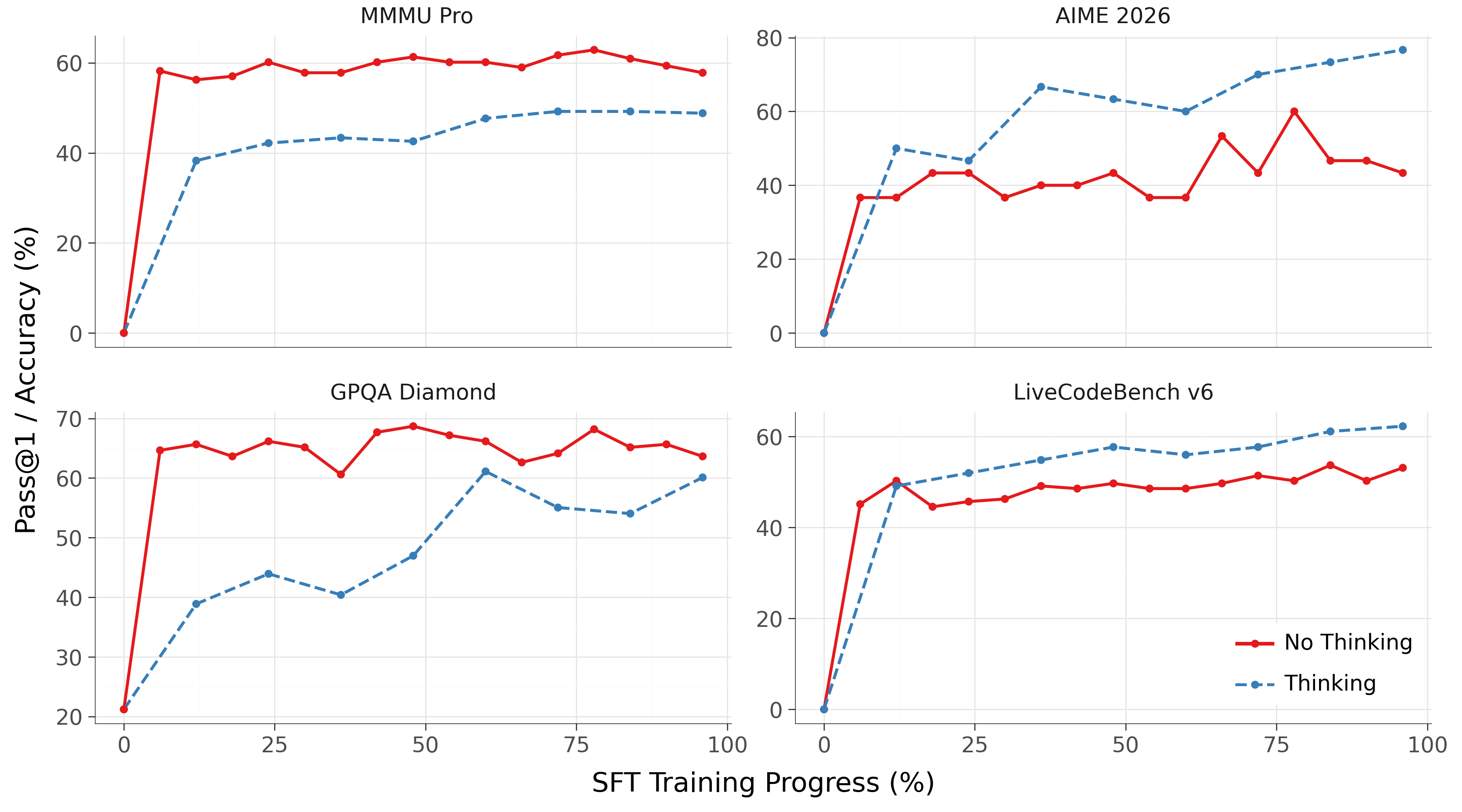}
    \caption{\textbf{Evolution of downstream performance during SFT.} Prior to SFT, the model is incapable of denoising text. Only a moderate amount of SFT is needed to achieve good performance in non-thinking mode, however extended SFT is crucial for the model to learn thinking behaviour. Results use entropy-bounded sampling with adaptive stopping and a maximum of $N=192$ denoising steps.}
    \label{fig:sft}
\end{figure}

\begin{figure}[!t]
    \centering
    \begin{minipage}[c]{0.6\linewidth}
        \includegraphics[width=0.95\linewidth]{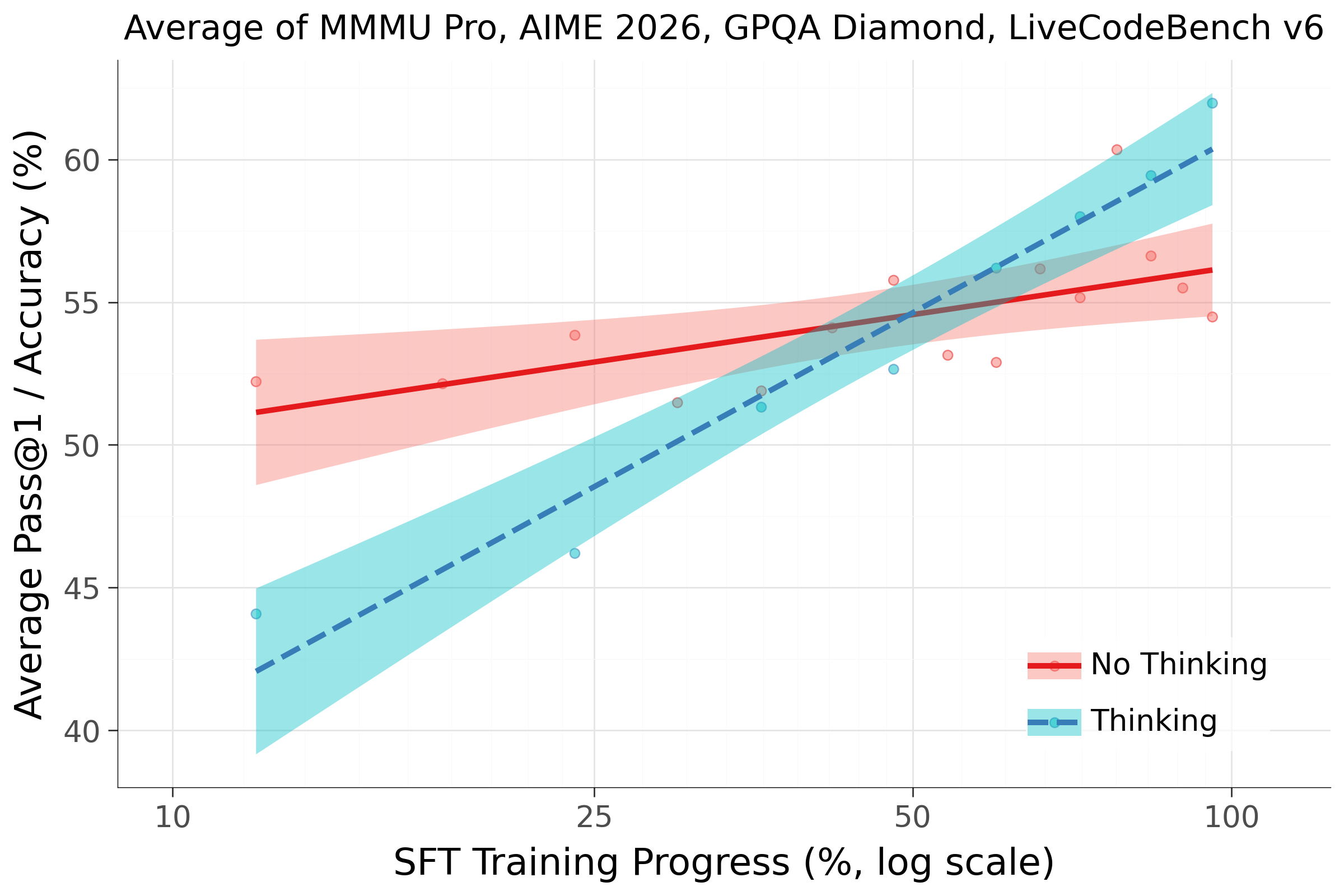}
    \end{minipage}\hfill
    \begin{minipage}[c]{0.35\linewidth}
        \caption{\textbf{Downstream performance during SFT improves log-linearly with training progress.} The log-linear trend for thinking performance starts off at a lower point yet exhibits a steeper slope than non-thinking mode. Results use the EntropyBounded sampler with adaptive stopping and a maximum of $N=192$ denoising steps.}
        \label{fig:average_sft}
    \end{minipage}
\end{figure}

We start from the publicly released Gemma 4 26B A4B~\citep{gemmateam2026gemma4technicalreport} checkpoint and run an extended finetuning phase where the model adapts to predicting blocks of 256 tokens from noisy inputs. We use a block-diagonal attention mask, enabling bidirectional attention within each block without allowing the model to condition on other denoising blocks. For a given canvas, the model conditions on the prompt and previous (uncorrupted) tokens via the encoder KV cache. We use discrete \emph{multinomial diffusion} as our corruption process and uniformly sample noisy tokens from the vocabulary~\citep{hoogeboom2021argmax, austin2021structured}. For a given canvas, we sample a noise level $t \sim \unif[0, 1]$ and noise each token in the canvas with probability $t$. Given a clean context of prompt and previous canvases $H$ (encoded through the KV cache), a self-conditioning signal $z_t$, and a noisy canvas $x_t$, the model is trained to minimize the cross-entropy loss between its predictions and the ground-truth canvas extracted from the training data~\citep{gat2024discrete}:
\begin{equation}
    L(\theta) = - \sum_{i=1}^C \log p_{\theta}(x_0^i \mid x_t, z_t, H),
\end{equation}
where superscript $i$ denotes indexing along the canvas dimension; $p_\theta$ is parameterized via a softmax transformation over the neural network's output logits. As shown in Figures~\ref{fig:sft} and~\ref{fig:average_sft}, denoising performance improves rapidly within the initial steps of training, after which it settles into a log-linear performance improvement trend. Thinking performance benefits from extended SFT, as the model initially struggles with maintaining coherent reasoning traces, often collapsing into stuttering or cycles---a common challenge for internalizing reasoning in language models~\citep{zelikman2024quiet}.

\section{Sampler Distillation \& Reinforcement Learning} \label{sec:sdrl}

Following the SFT stage, the model achieves strong generation quality when using a high number of denoising steps. However, its performance on advanced reasoning and coding tasks is somewhat poorer than the baseline AR model. More critically, when operating in the few-step regime required for ultra-low latency inference, generation quality collapses. To address this, we target a dual improvement: pushing the model's intelligence while simultaneously compressing its denoising trajectory.

Traditionally, achieving these two goals requires a decoupled, multi-stage pipeline that treats reward-driven alignment and sampler distillation as distinct phases. We bypass this with a unified \textit{online learning} stage, coined \textbf{sampler distillation \& reinforcement learning} (\sdrl), which optimizes both axes concurrently. Relying on a joint objective, a single gradient update drives:

\begin{enumerate}[topsep=0pt]
    \item \textbf{Reward maximization}: Elevating absolute generation quality and alignment, analogous to standard RL for AR and diffusion models \citep{black2024training, liu2025flow, zheng2025diffusionnft, zhao2025d1, ma2025reinforcement, wallace2023diffusion, fan2023dpok, clark2024directly, dong2023raft, lee2023aligning}.
    \item \textbf{Sampler distillation}: Mapping this high-quality generation to the few-step regime, addressing a compression challenge unique to iterative diffusion frameworks \citep{salimans2022progressive, song2023consistency, deschenaux2025beyond, fu2025learnable, luo2023latent, sauer2024adversarial, liu2023instaflow, yin2024distribution, hoogeboom2026beyond}.
\end{enumerate}

\paragraph{Training setup.} We adapt the data distribution used by the Gemma~4 RL recipe \citep{gemmateam2026gemma4technicalreport} for our use case. Encompassing both thinking and non-thinking modes, our setup targets improvements across a variety of capabilities such as helpfulness, mathematical reasoning, coding and instruction-following. Initialized from the SFT weights, the model acts as an online teacher that generates denoising trajectories (using a sampler configured with high maximum denoising steps and mild temperature annealing) to establish a high-quality reference. The \sdrl joint objective uses these trajectories to simultaneously maximize reward and drive sampler distillation in order to compress the model's highest-quality output into the few-step regime.

\begin{figure*}[!t]
    \centering
    \includegraphics[width=0.8\linewidth]{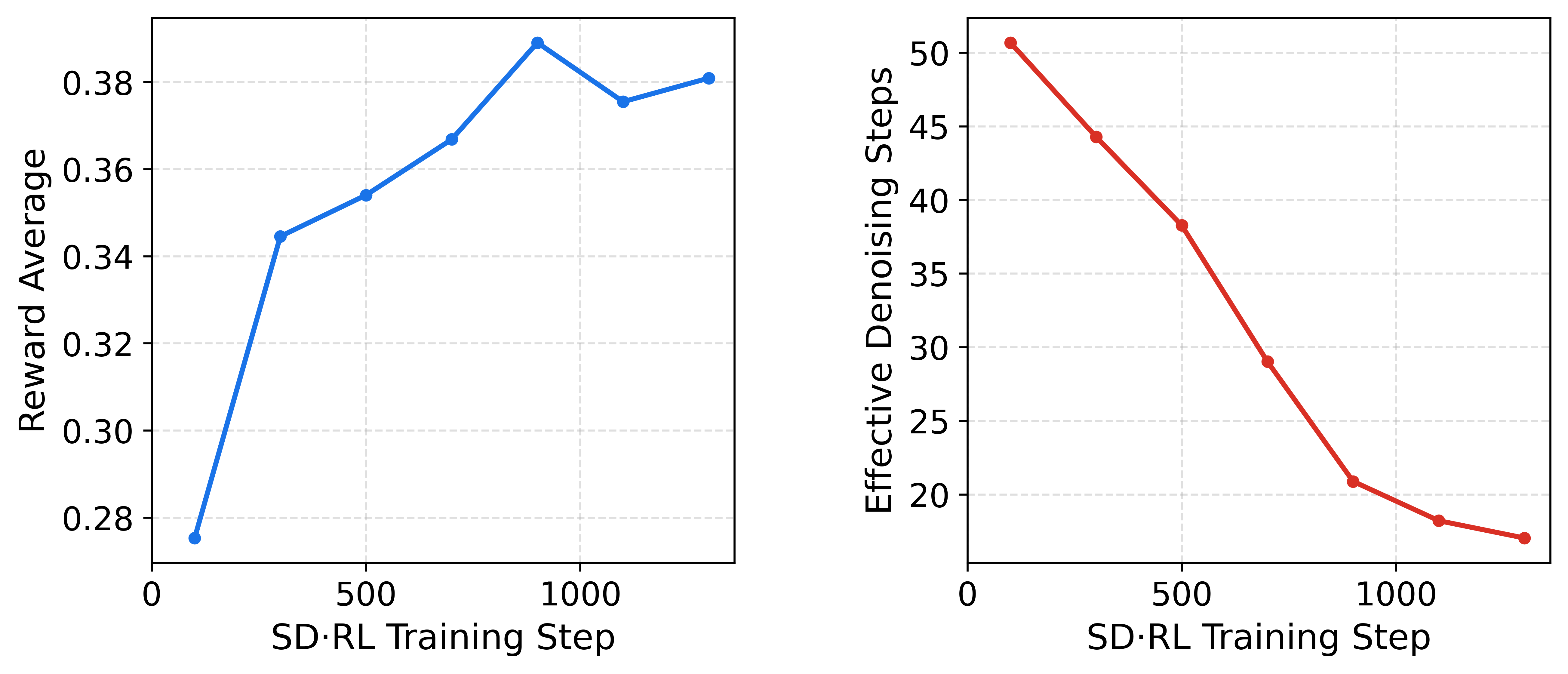}
    \caption{
        \textbf{\sdrl training simultaneously increases average reward and reduces the effective denoising steps of the online teacher} (training metrics shown in buckets of 200 steps). Both effects together push the quality-speed Pareto frontier of the model.
    }
    \label{fig:rl-training-metrics}
\end{figure*}

\paragraph{Training dynamics \& implicit curriculum effect.} Over the course of \sdrl training, two synergistic dynamics emerge, as shown in Figure~\ref{fig:rl-training-metrics}. First, the online teacher's average reward steadily increases, reflecting improved fundamental capabilities. Second, facilitated by the adaptive stopping mechanism, the online teacher progressively requires fewer effective denoising steps to achieve these high rewards. This acceleration occurs because the \sdrl objective systematically reduces the predictive entropy of the model. Crucially, the interplay between the reward objective and adaptive stopping induces a curriculum learning effect. Early in training, high predictive entropy delays the adaptive stopping trigger. As the model's confidence improves and entropy drops, adaptive stopping triggers earlier. This seamlessly shifts the training distribution toward ever shorter denoising trajectories, allowing the algorithm to dynamically pace its own sampler distillation. Consequently, and in contrast to RL for AR models, prolonging the \sdrl phase remains highly beneficial even after the reward metric plateaus; continued entropy reduction translates directly into further inference speedups.

\begin{figure*}[!t]
    \centering
    \includegraphics[width=1.0\linewidth]{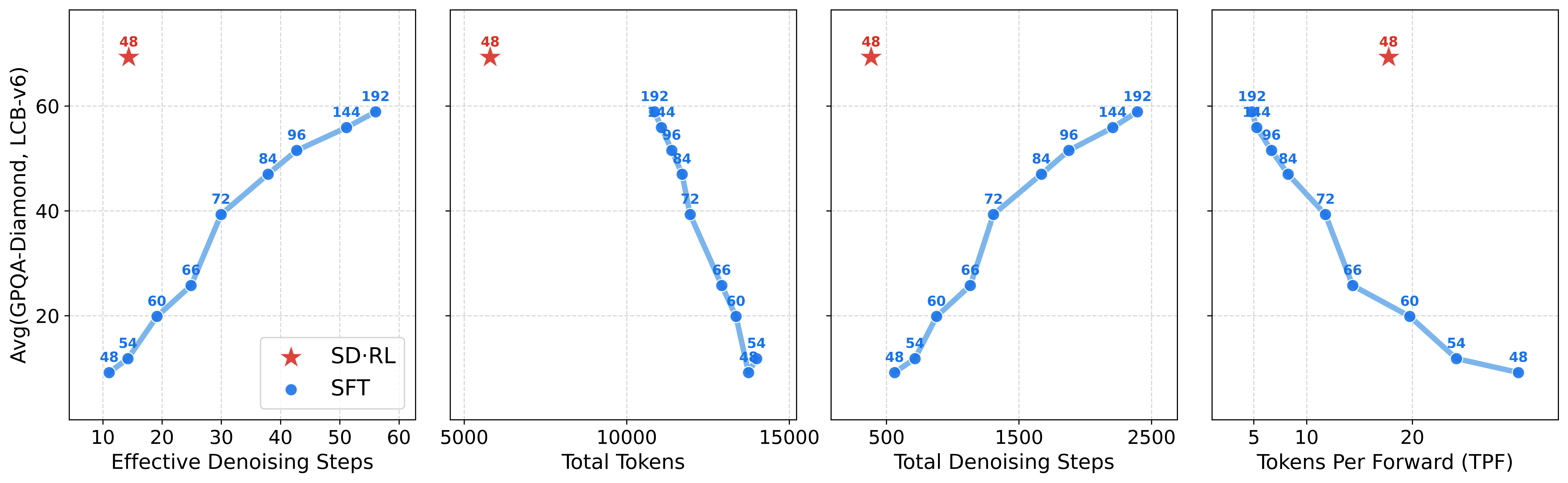}
    \caption{\textbf{\sdrl significantly advances the quality-speed Pareto frontier.} The \sdrl configuration uses the DiffusionGemma sampler with a maximum of $N=48$ denoising steps. The SFT frontier is derived by sweeping $N$ from 48 to 192 (note that higher $N$ yields milder temperature annealing). Both quality (y-axis) and inference efficiency (x-axes, see Section~\ref{sec:metrics}) are calculated as the average between GPQA-Diamond and LiveCodeBench-v6 in thinking mode, averaged over 3 seeds.}
    \label{fig:pareto-rl-vs-sft}
\end{figure*}

\paragraph{Improved speed-to-intelligence Pareto frontier.} As illustrated in Figure~\ref{fig:pareto-rl-vs-sft}, \sdrl training significantly expands the Pareto frontier established by the SFT checkpoint. On the quality axis, it yields a 10-point improvement on the combined GPQA-Diamond and LiveCodeBench-v6 score. On the efficiency axis, it quadruples the TPF from 5 to nearly 20, unlocking ultra-low latency inference.

As shown in Figure~\ref{fig:pareto-rl-vs-sft}, evaluating the SFT checkpoint with the default DiffusionGemma sampler (maximum of $N=48$ denoising steps) results in poor downstream accuracies and artificially low effective denoising steps. This counterintuitive behavior stems from the SFT model frequently degenerating into repetitive token loops in this restricted-step regime. Once it falls into a loop, its predictive entropy collapses, which prematurely triggers the adaptive stopping mechanism. Figures~\ref{fig:gpqa-rl-example-biology} and \ref{fig:gpqa-rl-example-quantum} (Appendix~\ref{sec:samples_before_after_sdrl}) provide generated samples on the GPQA-Diamond benchmark that illustrate how these degeneracies manifest: the SFT model begins with a valid, logical reasoning trace but suddenly collapses into a token repetition loop. In contrast, the samples post-\sdrl avoid these degenerative traps to sustain coherent reasoning at minimal latency.

\begin{figure*}[t]
    \centering
    \includegraphics[width=0.6\linewidth]{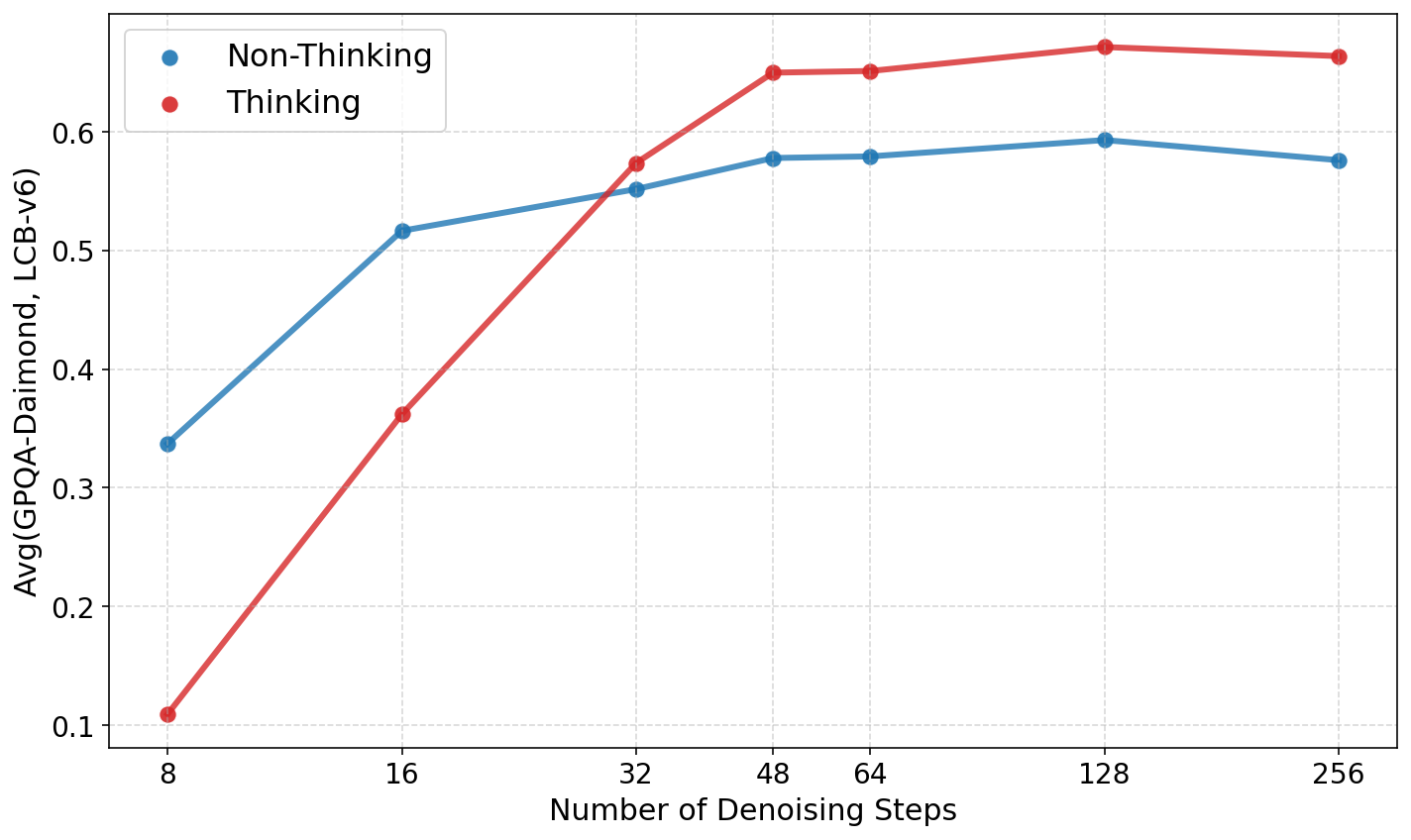}
    \caption{\textbf{Performance vs.\,number of denoising steps $N$, without adaptive stopping or temperature annealing.} Performance is measured as the average score over GPQA-Diamond, LiveCodeBench-v6 (3 seeds). To remove confounding effects of early stopping and temperature annealing, we disabled adaptive stopping and use temperature $\tau_t = 1$.}
    \label{fig:score_vs_dns}
\end{figure*}

\paragraph{\sdrl specialization in the few-step regime.} For the SFT checkpoint, downstream performance scales consistently with the maximum number of denoising steps $N$ up to 192, at which point denoising is roughly any-order autoregressive. Following \sdrl, the scaling behavior improves faster in $N$ and plateaus earlier: performance improves steadily up to $N=48$, but exhibits diminishing returns thereafter (Figure~\ref{fig:score_vs_dns}). This early saturation emerges because the \sdrl objective explicitly specializes the model for the few-step regime by aggressively minimizing predictive entropy.

\paragraph{Emergent conciseness.} 
An emergent property of our \sdrl optimization is that it encourages the model to produce concise, token-efficient outputs. This contrasts with RL for AR models, which often maximizes reward by inducing longer reasoning traces. Our final checkpoint produces generations nearly $2\times$ shorter than the SFT checkpoint (Figure~\ref{fig:pareto-rl-vs-sft}). While this means the model forgoes some capability gains typically associated with extended reasoning, it acts as a multiplier on inference speed gains. Compounding fewer total tokens with fewer effective denoising steps per canvas directly drives exceptionally low end-to-end latency: DiffusionGemma requires less than $5\%$ of the total forward passes used by the Gemma 4 AR baseline across our eval suite.

\section{Inference Optimizations} \label{sec:inference_optimizations}

The preceding sections focused on maximizing the number of tokens produced per forward pass through our SFT and \sdrl pipeline. We now turn to the complementary axis: minimizing the wall-clock cost of each forward pass through targeted GPU-level optimizations. Fundamentally, the throughput advantage of text diffusion over AR decoding is dictated by two competing quantities: on one hand, each denoising step decodes $\mathrm{TPF}$ tokens, and thus requires $\mathrm{TPF}$ times fewer forward passes than an AR baseline. On the other hand, because each forward pass processes a canvas of 256 tokens, each such step is $r$ times slower, and so the overall throughput relative to the AR baseline becomes $\mathrm{TPF}/r$. The forward pass overhead can be substantially reduced by hardware optimizations.

\paragraph{Low batch size serving.} While text diffusion inference requires more floating-point operations (FLOPs) per generated token than AR models, it relies on significantly fewer forward passes. Because LLM serving on modern hardware accelerators is typically memory-bound---largely dictated by KV cache capacity and memory bandwidth~\citep{kwon2023efficient, pope2023efficiently}---this reduction in memory transfers creates a latency advantage that outweighs the higher computational cost. As a result, text diffusion is highly effective in low-batch-size scenarios, leveraging available compute capacity to minimize per-request latency. For simplicity, our analysis focuses on the single-request inference throughput (a batch size of 1) of DiffusionGemma, comparing it directly to its AR counterpart, Gemma 4 26B A4B. Reference inference implementations are available in HuggingFace Transformers~\citep{diffusiongemma2026hf} and vLLM~\citep{vllm2026diffusiongemma}.

\begin{figure}[htbp]
    \centering
    \includegraphics[width=1\linewidth]{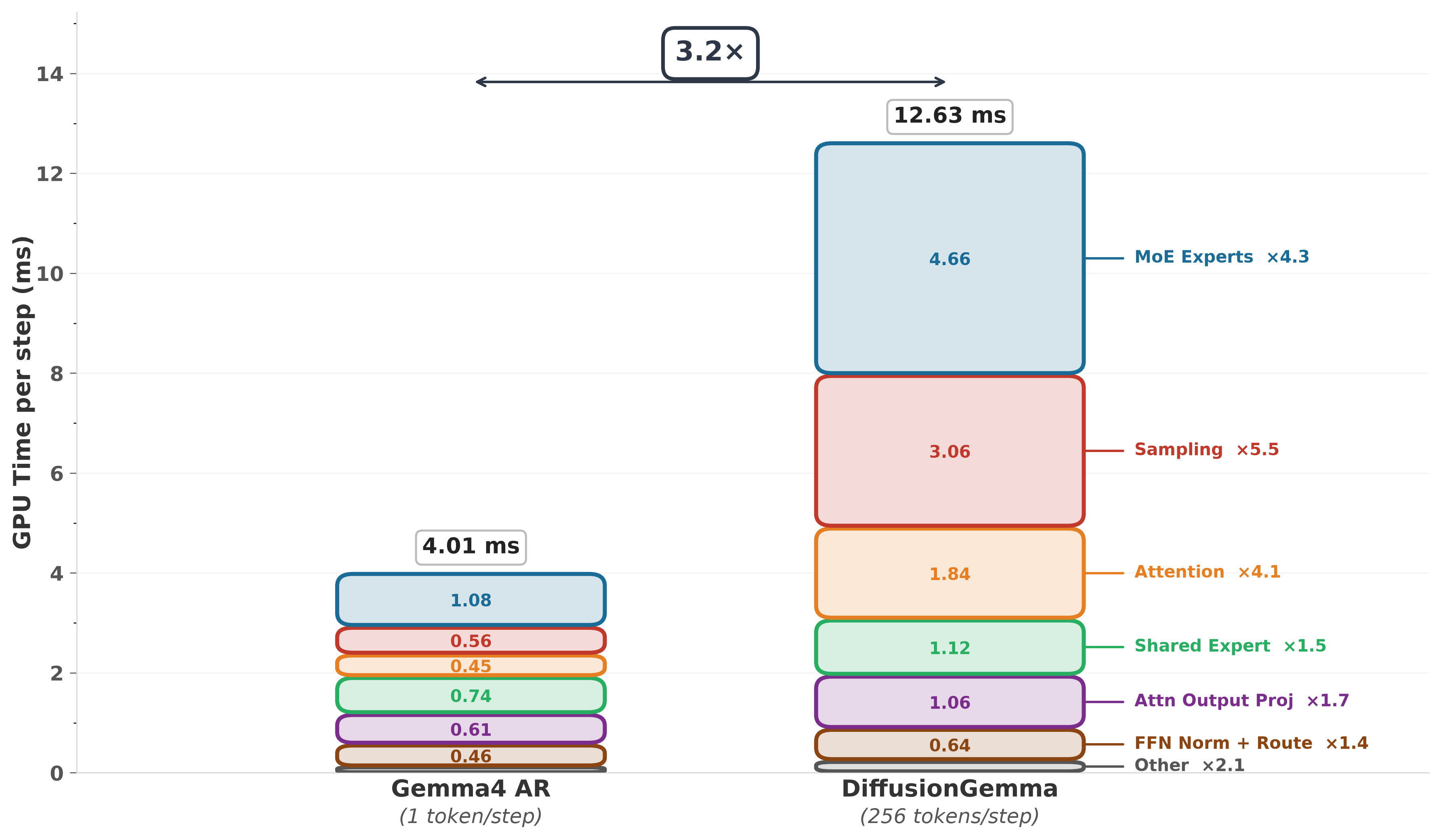}
    \caption{
        \textbf{Per-step GPU time breakdown: DiffusionGemma processes 256 tokens per step with only a 3.2$\times$ increase in per-step latency compared to single-token AR generation.} Serving a single request (batch size 1) on a H100, FP8 precision, with 4096 input tokens, 1024 output tokens.
    }
    \label{fig:inference-comparison}  
\end{figure}

\paragraph{GPU time breakdown.} Figure~\ref{fig:inference-comparison} presents the per-step GPU kernel time breakdown for both models; we focus on GPU kernel time rather than end-to-end latency to isolate model-specific bottlenecks. For each model individually, end-to-end latency exceeds GPU kernel time by approximately 1\,ms due to detokenization and other CPU-side
serving overhead, with this gap being similar for both the Gemma 4 AR and DiffusionGemma models. Each DiffusionGemma step processes $256\times$ more tokens, yet is only $3.2\times$ slower than the single-token AR step. The time difference can be mainly attributed to three operations: mixture-of-experts (MoE), sampling, and attention. Other operations (e.g., shared expert, attention output projection, etc.) are at most $2\times$ slower. 

\begin{itemize}[topsep=0pt]
    \item \textbf{MoE.} When serving a single request, the MoE layer computation is memory-bound, with its running time dominated by the transfer of expert weights from high-bandwidth memory, a well-documented bottleneck when serving sparse MoE models~\citep{rajbhandari2022deepspeedmoe, huang2024efficient}. For the Gemma 4 AR model, only 8 unique experts are activated per token per MoE layer. For the DiffusionGemma model, on average, approximately 84 unique experts are activated per canvas of 256 tokens per MoE layer, as measured on the PG-19 benchmark~\citep{rae2020compressive}.\footnote{Dataset available at \url{https://github.com/google-deepmind/pg19}.} Activating more experts per forward pass results in a $4.3\times$ slower MoE kernel. It is the same type of penalty incurred by verification in speculative decoding, but scaled to larger token parallelism. For a dense architecture, this overhead would be eliminated, reducing the per-step feed-forward network slowdown to less than $2\times$. 
    \item \textbf{Sampling.} Beyond AR's single-token softmax-and-sample, text diffusion sampling requires additional operations, most notably a self-conditioning embedding matmul and softmax over the full canvas of 256 tokens (see Algorithm~\ref{alg:sampling}). These operations are performed in full precision with a vocabulary dimension of 262k. Consequently, text diffusion sampling takes 3.06ms while AR sampling takes only 0.56ms. We implement sampling using standard PyTorch primitives optimized with \texttt{torch.compile}, rather than hand-written GPU kernels, to facilitate easier extensibility by the community.
    \item \textbf{Attention.} Unlike AR, DiffusionGemma uses bidirectional attention over a canvas of 256 tokens. As such, we cannot utilize the fast single-token decoding attention available to AR models, but we can take advantage of the highly-optimized FlashAttention-4 kernel \citep{dao2022flashattention, zadouri2026flashattention4}. Under these optimizations, the attention operation is $4\times$ slower for the DiffusionGemma model than the Gemma 4 AR model. 
\end{itemize}

\paragraph{Eliminating CPU-GPU synchronization.} There is additional complexity when serving DiffusionGemma: denoising steps depend on adaptive stopping and there are two types of forward passes (the denoising step and the KV cache update) that require different attention masks and that can occur within the same batch. To minimize request latency, it is important that this complexity is handled solely by GPU operations without triggering any additional CPU-GPU synchronization. This is achieved by extending asynchronous scheduling to the text diffusion models and introducing a per-sequence causal attention flag (see \citealp{vllm2026diffusiongemma} for more details).

\paragraph{Throughput for batch size of 1 serving.} The decoding throughput for text diffusion models is calculated by the following formula:
\begin{equation}\label{eq:tps}
\text{TPS} \triangleq \dfrac{\text{TPF}}{t_{\text{fwd}}},
\end{equation}
where TPS is Tokens Per Second, TPF is Tokens Per Forward (see Equation~\ref{eq:tpf}), and $t_{\text{fwd}}$ is the time needed for a single denoising step, which varies with context length (due to attention). On an H100 GPU (FP8 precision), a single denoising step of DiffusionGemma takes $t_{\text{fwd}}=13.56\text{ms}$ on average (end-to-end; the per-step GPU time is 12.63ms, see Figure~\ref{fig:inference-comparison}) when serving a single request with 4096 input tokens and 1024 output tokens. TPF varies depending on the task. Assuming a TPF of 19.74 (the average TPF across the 7 benchmarks reported in Table~\ref{tab:model_comparison}), the average decoding throughput of the model is 1456 TPS---a $7.1\times$ improvement over the Gemma 4 AR model (204 TPS) and a $4.8\times$ improvement over the AR model with MTP (303 TPS), on the same device setup.

\begin{figure}[!t]
    \centering
    \includegraphics[width=1\linewidth]{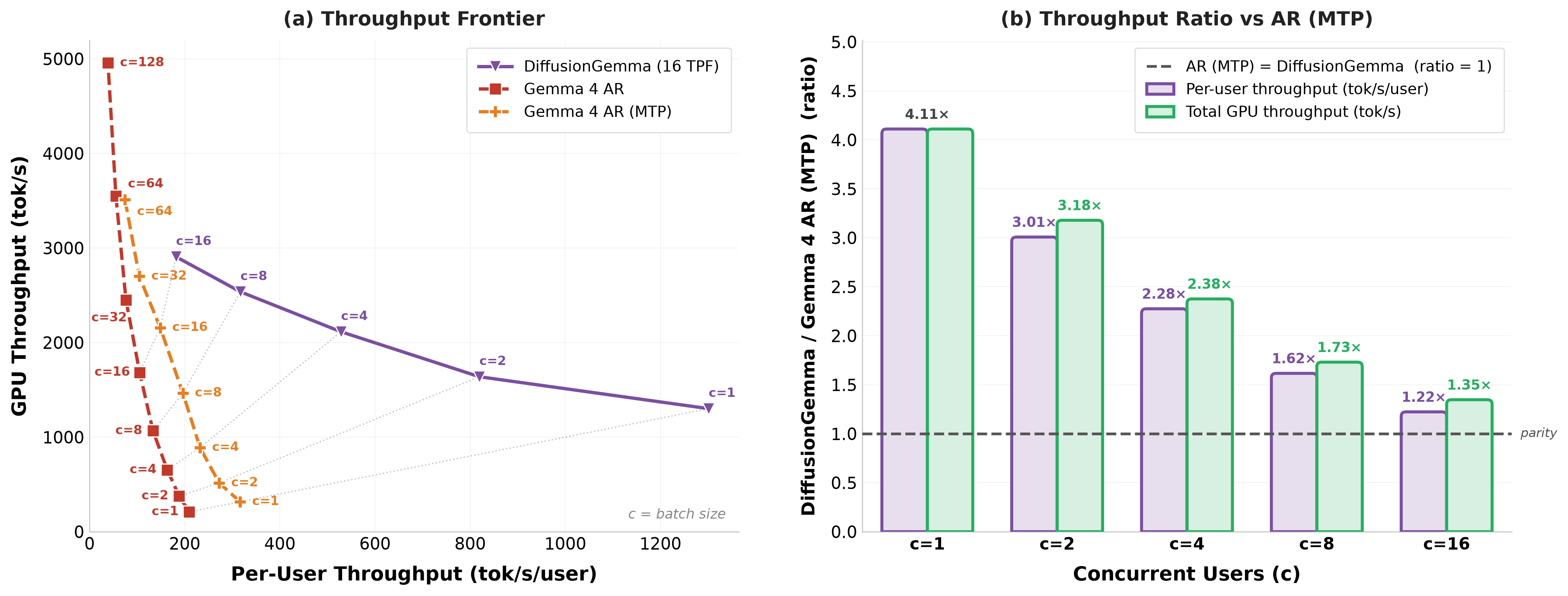}
    \caption{
        \textbf{Trade-off between total and per-user throughput of the Gemma 4 AR model (with and without MTP) and DiffusionGemma.} In the low batch size regime, DiffusionGemma offers substantially higher TPS per user \emph{and} higher total throughput. It is only at moderate batch sizes (around 32 concurrent requests) that AR models begin to have a throughput advantage. All models are run on an H100 with FP8 precision using the PG-19 benchmark (4096 input tokens, 1024 output tokens); the Gemma 4 AR (MTP) model uses a draft length of 4.
    }
    \label{fig:inference-throughput} 
\end{figure}

\paragraph{Multi-user throughput.} Figure~\ref{fig:inference-throughput} shows the trade-off between total and per-user throughput of the Gemma 4 AR model (with MTP) and DiffusionGemma depending on the number of concurrent users (i.e., batch size). Importantly, these results are without targeted optimization for batch sizes larger than 1: current kernel selection is suboptimal, and sampling has not been tuned to scale with batch size; for example, applying top-$k$ truncation to the sampling step is expected to yield significant throughput gains at higher batch sizes with negligible impact on output quality. Despite this, DiffusionGemma offers substantially higher TPS per user \emph{and} higher total throughput than the Gemma 4 AR (MTP) model in the low batch size regime, with AR models beginning to gain a throughput advantage only at moderate batch sizes (around 32 concurrent requests).

\paragraph{Toward real traffic throughput.} Compared to the AR equivalent, DiffusionGemma changes the compute characteristics of two transformer layers: for the attention layer, it performs TPF$\times$ fewer transfers of KV cache; for the MoE layer, it performs proportionally more FLOPs (scaling with the number of effective denoising steps). This has the potential to address one of the challenges in modern LLM serving: memory-bound attention limiting compute utilization in FFW/MoE layers (\citealp{zhu2024nanoflow, tang2024quest, zhu2025megascaleinfer}), which is particularly relevant for agentic workflows with long context. DiffusionGemma effectively trades data movement for compute, which is favorable on modern GPU hardware where compute-to-bandwidth ratios continue to grow; a thorough empirical analysis under realistic traffic conditions is beyond the scope of this work.

\section{Experimental Results} \label{sec:experiments}

We evaluate DiffusionGemma's capabilities and inference efficiency across four operational modes: text diffusion (TD) versus autoregressive (AR) generation, each evaluated with and without thinking enabled; unless otherwise specified, TD with thinking is enabled. We benchmark the model against its AR initialization (Gemma 4 26B A4B), contemporary open-weight text diffusion models (LLaDA 2.1 Flash 100B and Nemotron Diffusion 14B), and the proprietary Mercury 2 API.

We consider a diverse suite of benchmarks spanning core domains such as mathematical reasoning, code generation, general knowledge, multimodal understanding, and instruction following alongside agentic capabilities. Specifically, for mathematical reasoning, we utilize AIME \citep{dekoninck2026matharena}, GSM8K \citep{cobbe2021training}, MGSM \citep{shi2022language}, Putnam \citep{tsoukalas2024putnambench}, and HiddenMath (internal). Coding performance is measured against LiveCodeBench-v6 \citep{jain2025livecodebench}, Codeforces \citep{quan2025codeelo}, HumanEval \citep{chen2021evaluating},  BigCodeBench \citep{zhuo2025bigcodebench}, LBPP (v2) \citep{matton2024leakage}, and Natural2Code (internal). Broad and expert-level general knowledge is measured against GPQA-Diamond \citep{rein2023gpqa}, BIG-Bench \citep{suzgun2022challenging}, MMMLU \citep{openai2024mmmlu}, and MMLU-Pro \citep{wang2024mmlupro}, while multimodal reasoning is evaluated on MMMU-Pro \citep{yue2025mmmu}. Finally, strict instruction following is assessed via IFEval \citep{zhou2023instruction}, and agentic task completion is evaluated through the Tau-bench suite, encompassing the Retail, Airline, and Telecom environments \citep{yao2024tau}.

\begin{figure*}[t]
    \centering
    \includegraphics[
        width=\textwidth,
        height=0.70\textheight,
        keepaspectratio
    ]{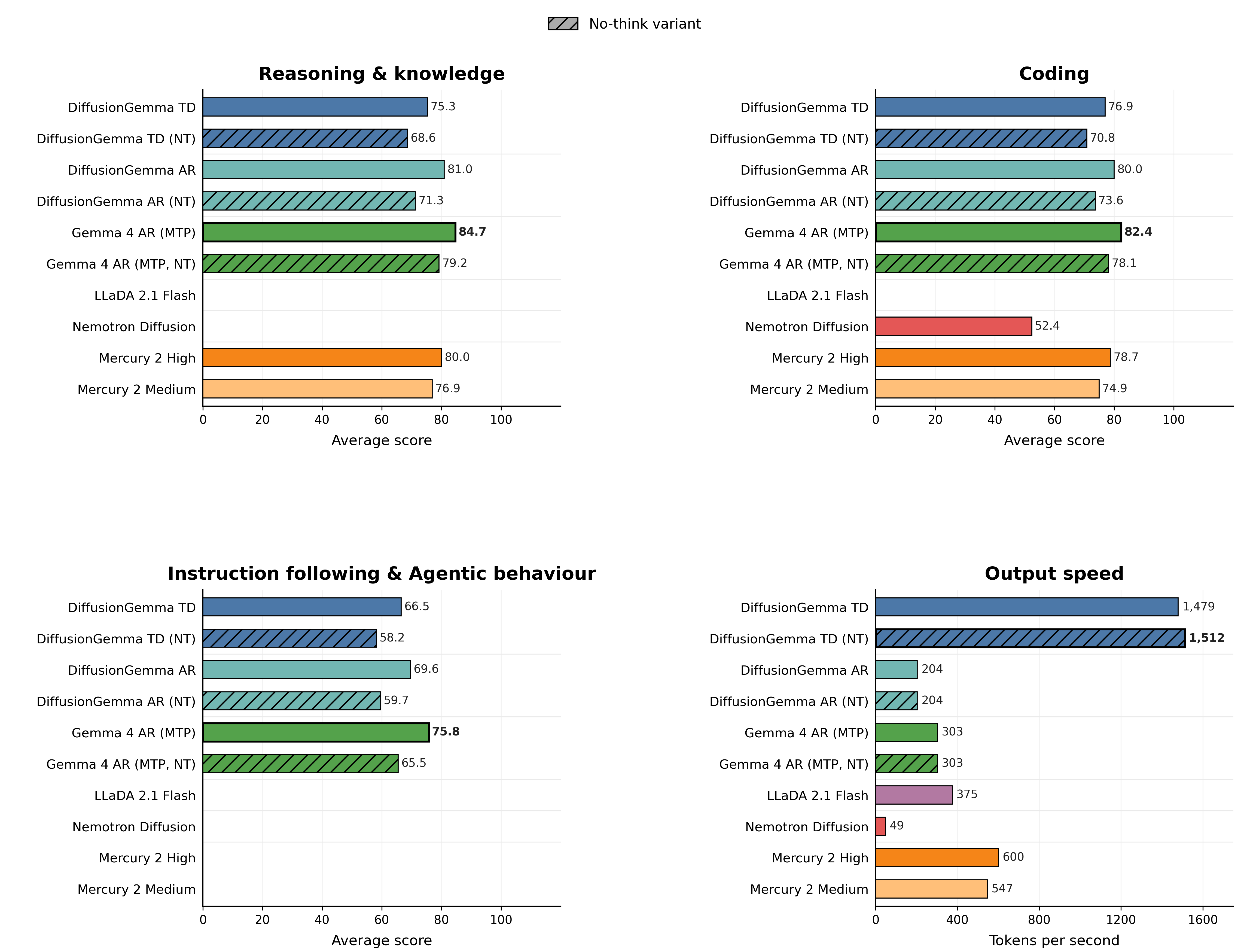}
    \caption{
        \textbf{Performance by capability area\textsuperscript{*}
        (unweighted mean of 0--100 scores across the area) and output speed.} We omit models for which we do not have complete coverage. Non-thinking variants are indicated using striped bars. Output speed is averaged over the 7 benchmarks for which we have full coverage, see Table~\ref{tab:model_comparison}.
    }
    \label{fig:benchmark_summary}
    \vspace{2pt}
    \parbox{\textwidth}{
        \footnotesize
        \raggedright
        \textsuperscript{*}
        Benchmarks are grouped into:
        reasoning and knowledge
        (\textit{AIME 2026}, \textit{GPQA Diamond}, \textit{BigBench EH},
        \textit{GSM8K}, \textit{MGSM}, \textit{MMMLU}, \textit{Putnam},
        \textit{MMLU-Pro}, and \textit{HiddenMath});
        coding
        (\textit{LiveCodeBench V6}, \textit{HumanEval},
        \textit{BigCodeBench}, \textit{LBPP}, and \textit{Natural2Code});
        and instruction following and agentic behaviour
        (\textit{IFEval}, \textit{Tau2 Retail}, \textit{Tau2 Airline},
        and \textit{Tau2 Telecom}).
    }
\end{figure*}

The complete per-benchmark results for all models and inference configurations are reported in Table~\ref{tab:model_comparison}. Table~\ref{tab:dg_td_latency} provides a complementary analysis of DiffusionGemma's decoding efficiency, reporting Tokens Per Forward (TPF; Equation~\ref{eq:tpf}), Tokens Per Second (TPS; Equation~\ref{eq:tps}), effective denoising steps (Equation~\ref{eq:effective_denoising_steps}), total generated tokens (Equation~\ref{eq:total_tokens}), and end-to-end generation latency, excluding prefill time. For a higher-level comparison, Figure~\ref{fig:benchmark_summary} groups the benchmarks into three capability areas---reasoning and knowledge, coding, and instruction following and agentic behaviour---and reports the unweighted mean of the 0--100 benchmark scores within each area, together with output throughput. A model is shown for a capability area only if it completed every constituent benchmark; missing bars therefore indicate incomplete benchmark coverage rather than a score of zero. Hatched bars denote the no-think variants.

DiffusionGemma sets a new performance frontier for text diffusion models, substantially outperforming existing open-weight diffusion baselines while increasing TPF by approximately an order of magnitude. In terms of quality it is  highly competitive with Mercury~2, a closed-weight text-diffusion model, while reaching roughly 1,500 output tokens per second on a single previous-generation H100 GPU, a roughly $2.5\times$ speedup over Mercury~2.

Relative to the AR model used for initialization, DiffusionGemma in text-diffusion (TD) mode trades some absolute benchmark performance for substantially greater decoding speed. Although the conversion reduces performance across the three capability areas, TD mode delivers nearly $5\times$ the output throughput of the original Gemma~4 AR baseline under heavily optimized MTP serving: 1,479 tokens per second compared with 303 tokens per second. The two-stage training pipeline nevertheless retains support for autoregressive decoding. When run in standard left-to-right AR mode, DiffusionGemma recovers part of the performance gap observed in TD mode and narrows the capability gap to the original baseline, albeit at lower throughput. This dual-mode capability could enable requests to be routed dynamically according to latency requirements and task complexity.

\begin{landscape} 

\definecolor{benchmarkhighlight}{RGB}{232,244,248} 
\begin{table}[t!]
\centering
\footnotesize
\setlength{\tabcolsep}{3pt}
\begin{tabular}{lcccccccccc}
\toprule
 & \multicolumn{8}{c}{\textbf{Open-weight Models}} & \multicolumn{2}{c}{\textbf{Closed-weight Model}} \\
\cmidrule(r){2-9} \cmidrule(l){10-11}
 & \multicolumn{4}{c}{DiffusionGemma} & \multicolumn{2}{c}{Gemma 4} & LLaDA 2.1 Flash & Nemotron Diffusion & \multicolumn{2}{c}{Mercury 2} \\
 & \multicolumn{4}{c}{26B A4B} & \multicolumn{2}{c}{26B A4B} & 100B & 14B & \multicolumn{2}{c}{Unknown} \\
\cmidrule(r){2-5} \cmidrule(r){6-7} \cmidrule(r){8-8} \cmidrule(r){9-9} \cmidrule(l){10-11}
Mode & TD & TD (No-think) & AR & AR (No-think) & AR (MTP) & AR (MTP, No-think) & TD (S Mode) & TD (Diffusion Mode) & High & Medium \\
\midrule
AIME 2026 & 69.1 & 50.8 & 84.2 & 57.5 & 88.3 & 80.0 & 80.0 & 40.0 & 91.7 & 82.5 \\
GPQA Diamond & 73.2 & 64.6 & 79.8 & 67.2 & 82.3 & 73.7 & 68.7 & 47.0 & 75.2 & 66.7 \\
LiveCodeBench-V6 & 69.1 & 60.6 & 71.4 & 58.3 & 77.1 & 72.6 & 39.4 & 28.6 & 79.4 & 74.9 \\
Codeforces ELO & 1429 & 959 & 1569 & 1059 & 1718 & 1529 & 718 & - & 1986 & 1629 \\
BigBench EH & 47.6 & 40.0 & 59.1 & 42.2 & 64.8 & 56.2 & - & - & 48.9 & 43.8 \\
GSM8K & 96.3 & 95.8 & 96.6 & 96.1 & 96.7 & 96.4 & 45.0 & - & 96.5 & 95.8 \\
MGSM & 84.8 & 80.7 & 87.9 & 84.3 & 92.9 & 91.5 & 6.8 & 69.3 & 91.9 & 91.2 \\
MMMLU & 81.5 & 76.3 & 82.2 & 78.0 & 86.3 & 78.0 & - & - & 81.9 & 80.6 \\
MMMU Pro & 54.3 & 66.0 & 63.3 & 66.7 & 73.8 & 72.5 & - & - & - & - \\
Putnam & 67.4 & 57.1 & 74.7 & 59.7 & 81.0 & 72.9 & - & 45.8 & 73.6 & 73.6 \\
HumanEval & 94.5 & 92.7 & 98.2 & 97.6 & 98.8 & 97.6 & 90.2 & 86.0 & 98.2 & 98.2 \\
BigCodeBench & 46.0 & 41.9 & 47.7 & 45.9 & 50.2 & 48.1 & - & 33.5 & 47.6 & 45.3 \\
LBPP & 81.0 & 68.9 & 86.3 & 74.1 & 89.5 & 77.3 & 45.7 & 40.7 & 89.2 & 85.0 \\
IFEval & 97.4 & 94.5 & 97.2 & 95.7 & 98.7 & 97.8 & - & 72.1 & 97.0 & 94.5 \\
Tau2 Retail & 71.5 & 57.5 & 75.4 & 61.0 & 85.5 & 79.0 & - & - & - & - \\
Tau2 Airline & 69.0 & 49.0 & 72.0 & 50.0 & 76.0 & 51.0 & - & - & - & - \\
Tau2 Telecom & 28.1 & 32.0 & 33.8 & 32.0 & 43.0 & 34.2 & - & - & - & - \\
MMLU-Pro & 77.6 & 77.9 & 78.8 & 79.1 & 82.6 & 82.6 & - & - & 77.6 & 75.5 \\
\rowcolor{benchmarkhighlight}
Natural2Code & 94.0 & 90.1 & 96.2 & 92.3 & 96.3 & 94.7 & 86.9 & 73.3 & 79.1 & 71.3 \\
\rowcolor{benchmarkhighlight}
HiddenMath & 80.6 & 74.3 & 85.4 & 77.5 & 87.2 & 81.6 & - & 44.3 & 82.7 & 82.3 \\
\midrule
Output Speed (TPS) & 1479 & 1512 & 204 & 204 & 303 & 303 & 375 & 49 & 600 & 547 \\
Tokens Per Forward (TPF) & 19.74 & 18.76 & 1.00 & 1.00 & 1.40 & 1.40 & 4.63 & 1.79 & - & - \\
Average Total Tokens & 4,001 & 829 & 5,184 & 1,025 & 7,207 & 1,816 & 4,371 & 941 & 3,882 & 1,222 \\
\bottomrule
\end{tabular}
\caption{\textbf{Comparison of model performance, TPS and TPF across various benchmarks.} TPS excludes prefill time; ``$-$'' denotes missing data. For speed measurements: DiffusionGemma and Gemma~4 are measured on 1$\times$~H100 (FP8, batch size~1); Nemotron~14B on 1$\times$~H100 (bfloat16, batch size~1); LLaDA~2.1 Flash~100B on 8$\times$~B200 (bfloat16, batch size~1); Mercury~2 via its public API (see Appendix~\ref{sec:competitor-eval-methodology} for speed estimation). TPS, TPF and total tokens are averaged over the 7 benchmarks for which we have full coverage: AIME 2026, GPQA Diamond, LiveCodeBench-v6, MGSM, HumanEval, LBPP, and Natural2Code. TPS and TPF for Gemma 4 (MTP) are measured using SPEED-Bench~\citep{abramovich2026speed}. The Natural2Code and HiddenMath rows are highlighted as they are proprietary, unleaked evals.
}
\label{tab:model_comparison}
\end{table}

\begin{table}[t!]
\centering
\small
\setlength{\tabcolsep}{2.5pt}
\begin{tabular}{l|cc|cc|cc|cc|cc|cc|cc}
\toprule
 & \multicolumn{2}{c}{\textbf{Score} ($\uparrow$)} & \multicolumn{2}{c}{\textbf{TPF} ($\uparrow$)} & \multicolumn{2}{c}{\textbf{TPS} ($\uparrow$)} & \multicolumn{2}{c}{\textbf{Effective DNS} ($\downarrow$)} & \multicolumn{2}{c}{\textbf{Total Forwards} ($\downarrow$)} & \multicolumn{2}{c}{\textbf{Total Tokens}} & \multicolumn{2}{c}{\textbf{E2E Time (s)} ($\downarrow$)} \\
\cmidrule(lr){2-3} \cmidrule(lr){4-5} \cmidrule(lr){6-7} \cmidrule(lr){8-9} \cmidrule(lr){10-11} \cmidrule(lr){12-13} \cmidrule(lr){14-15}
\textbf{Benchmark} & \textbf{Think} & \textbf{No-Think} & \textbf{Think} & \textbf{No-Think} & \textbf{Think} & \textbf{No-Think} & \textbf{Think} & \textbf{No-Think} & \textbf{Think} & \textbf{No-Think} & \textbf{Think} & \textbf{No-Think} & \textbf{Think} & \textbf{No-Think} \\
\midrule
AIME 2026 & 69.1 & 50.8 & 19.3 & 16.7 & 1365.4 & 1333.0 & 12.6 & 14.1 & 390.6 & 91.1 & 6,445 & 1,309 & 4.72 & 0.98 \\
GPQA Diamond & 73.2 & 64.6 & 16.7 & 16.5 & 1207.8 & 1330.2 & 15.1 & 13.4 & 443.4 & 48.1 & 5,647 & 726 & 4.68 & 0.55 \\
LiveCodeBench-V6 & 69.1 & 60.6 & 18.5 & 16.9 & 1278.3 & 1333.4 & 13.8 & 14.0 & 581.8 & 195.7 & 7,534 & 1,847 & 5.89 & 1.39 \\
Codeforces ELO & 1429 & 959 & 15.1 & 14.0 & 950.5 & 1040.3 & 17.1 & 18.5 & 959.6 & 521.1 & 11,622 & 4,279 & 12.23 & 4.11 \\
BigBench EH & 47.6 & 40.0 & 20.8 & 17.7 & 1390.2 & 1415.1 & 11.9 & 12.8 & 434.9 & 68.1 & 9,062 & 1,233 & 6.52 & 0.87 \\
GSM8K & 96.3 & 95.8 & 23.2 & 24.1 & 1866.2 & 1966.4 & 9.1 & 7.5 & 43.4 & 13.2 & 883 & 298 & 0.47 & 0.15 \\
MGSM & 84.8 & 80.7 & 19.0 & 16.8 & 1526.7 & 1367.5 & 11.5 & 11.9 & 63.6 & 19.5 & 1,085 & 291 & 0.71 & 0.21 \\
MMMU Pro & 54.3 & 66.0 & 17.7 & 15.4 & 1351.3 & 1255.1 & 13.8 & 15.3 & 191.6 & 31.6 & 3,178 & 472 & 2.35 & 0.38 \\
Putnam & 67.4 & 57.1 & 18.1 & 16.0 & 1330.8 & 1282.4 & 13.4 & 14.4 & 303.8 & 77.4 & 4,725 & 1,103 & 3.55 & 0.86 \\
HumanEval & 94.5 & 92.7 & 23.0 & 24.3 & 1838.2 & 1981.2 & 9.4 & 8.0 & 55.6 & 14.3 & 1,174 & 305 & 0.64 & 0.15 \\
BigCodeBench & 46.0 & 41.9 & 19.6 & 19.4 & 1560.1 & 1579.0 & 11.3 & 10.1 & 77.3 & 21.8 & 1,410 & 394 & 0.90 & 0.25 \\
LBPP & 81.0 & 68.9 & 20.5 & 19.2 & 1509.8 & 1545.4 & 11.6 & 10.9 & 264.3 & 79.4 & 4,730 & 859 & 3.13 & 0.56 \\
IFEval & 97.4 & 94.5 & 17.2 & 9.0 & 1368.3 & 732.1 & 13.0 & 14.4 & 100.8 & 24.0 & 1,464 & 239 & 1.07 & 0.33 \\
Natural2Code & 94.0 & 90.1 & 21.1 & 21.0 & 1682.7 & 1706.5 & 10.5 & 9.6 & 70.5 & 24.7 & 1,391 & 465 & 0.83 & 0.27 \\
HiddenMath & 80.6 & 74.3 & 21.0 & 19.4 & 1591.2 & 1564.7 & 11.3 & 11.5 & 206.4 & 49.7 & 3,501 & 844 & 2.20 & 0.54 \\
\bottomrule
\end{tabular}
\caption{\textbf{Performance and latency metrics of DiffusionGemma TD in Thinking vs. No-Thinking mode.} We report accuracy/score, Tokens Per Forward (TPF, higher is faster), Tokens Per Second (TPS), Effective Denoising Steps (DNS, lower is faster), Total Forwards, Total Tokens, and End-to-End Latency per sample in seconds (excluding prefill time). The latency metrics are averaged across all samples within each benchmark.}
\label{tab:dg_td_latency}
\end{table}

\end{landscape}

\section{Open-Source Downstream SFT}\label{sec:downstream_sft}

Alongside DiffusionGemma, we release an open-source finetuning toolkit that allows practitioners to adapt the model to their own domain-specific datasets. We build on top of Hackable Diffusion~\citep{hackable_diffusion2026github}, a modular open-sourced research toolbox for generative modeling.
We provide Low-Rank Adaptation (LoRA) recipes~\citep{hu2021loralowrankadaptationlarge} to enable finetuning on consumer hardware.

\paragraph{Finetuning procedure.} Our open-source SFT toolkit includes both a causal encoder and a diffusion decoder objective. Training sequences are of length $P+KC$ where $P$ is the number of prompt tokens, $K$ is the number of canvases and $C$ is the canvas size. To compute the total loss, the encoder first processes all tokens in the entire sequence, populates a KV cache $H$, and provides next token predictions that are fed into a standard cross-entropy loss for the encoder. The decoder loss is computed by uniformly sampling a canvas $k$ from the $K$ canvases available. For canvas $k$, we evaluate a denoising cross-entropy loss using the decoder's predicted logits given the current noisy state, $x_t$, the KV cache $H$ for the prompt and any prior canvases in the sequence (i.e., canvas $1$, $2$, \ldots, $k-1$); for $50\%$ of the data-points in a batch, the decoder is also conditioned on a self-conditioning state $z_t$ computed from a previous forward pass. The other $50\%$ have $z_t = \textbf{0}$.
 Formally, we write the losses as
\begin{equation}
   L_{\text{encoder}}(\theta) = - \frac{1}{P+KC} \sum_{j=1}^{P+KC} \log p_\theta(x^j \mid x^{1:j-1}) , \quad  \quad L_{\text{decoder}}(\theta) = - \frac{1}{C} \sum_{i=1}^C \log p_\theta(x_0^{i} \mid x_t, z_t, H).
\end{equation}
For the encoder loss, $j$ indexes along all tokens in the sequence (prompt and canvases); for the decoder loss, $i$ indexes along tokens in the given canvas. The final loss is the sum of the two losses.

\paragraph{LoRA for parameter-efficient finetuning.} LoRA is applied to all linear operations (attention projections, MLP gates, MoE routers, and the self-conditioning feedforward block). This allows us to achieve strong downstream performance while training only a small fraction of the model's parameters, using 2$\times$ A100 80GB GPUs. All of our training details are reported in Appendix~\ref{sec:sft-hd-appendix}.

\begin{figure}[tbp]
    \centering
    \begin{minipage}[b]{0.45\textwidth}
        \centering
        \includegraphics[width=\linewidth]{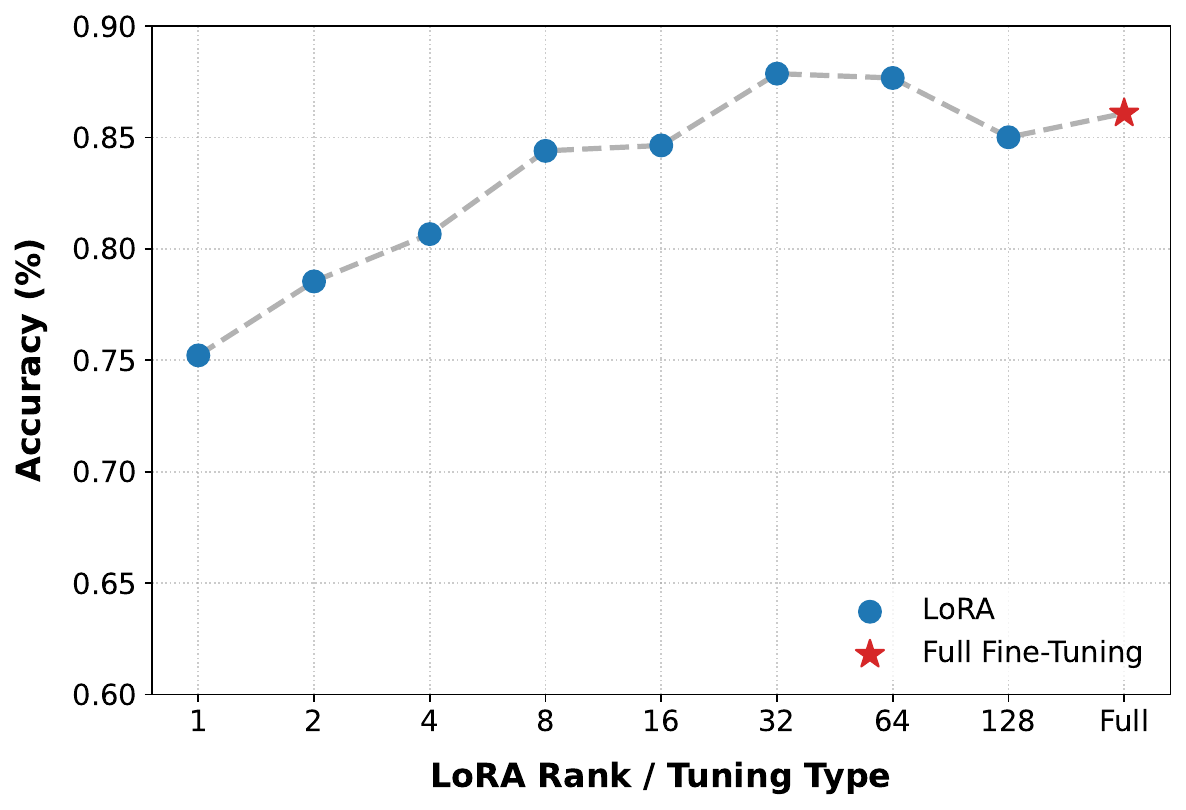}
        \captionof{figure}{
            \textbf{Sudoku Downstream SFT performance.} The x-axis is the LoRA rank and the y-axis is the accuracy of the model.
        }
        \label{fig:sft-sudoku} 
    \end{minipage}
    \hfill 
    \begin{minipage}[b]{0.50\textwidth}
        \centering
        \vspace{5pt} 
        \small 
        \begin{tabular}{lcc}
        \toprule
        Model & Denoising Steps & Accuracy (\%)  \\
        \midrule
        DiffusionGemma & 40.65 & 0.00 \\
        + LoRA finetuning & \textbf{10.72} & \textbf{84.40} \\
        \bottomrule
        \end{tabular}
        \captionof{table}{\textbf{Sudoku accuracy of the finetuned model with LoRA rank 8.}
        The base model fails to produce a correct Sudoku grid before finetuning. Finetuning the model leads to accuracy above $80\%$, while using significantly less effective denoising steps due to lower predictive entropy.}
        \label{tab:sudoku_steps_vs_acc}
    \end{minipage}
\end{figure}

\paragraph{Case study: Sudoku puzzle solving.} Solving Sudoku puzzles is a compelling testbed for discrete diffusion due to the non-autoregressive nature of the task. We finetune DiffusionGemma on an open-source dataset of Sudoku puzzles.\footnote{\url{https://www.kaggle.com/datasets/rohanrao/sudoku}} With full finetuning, the sampler (Algorithm~\ref{alg:sampling}) achieves >85\% puzzle-level accuracy evaluated on a held out set of $4096$ puzzles (Figure~\ref{fig:sft-sudoku}). Decreasing the LoRA rank trades-off accuracy and compute. In Table~\ref{tab:sudoku_steps_vs_acc}, we report performance of the original model as well as the finetuned model. See Appendix~\ref{sec:sft-hd-appendix} for additional results on PubMedQA~\citep{jin2019pubmedqa}.

\section{Practical Advantages of Text Diffusion}\label{sec:advantages}

In this manuscript, we have primarily emphasized low latency as the principal advantage of text diffusion over standard AR language modeling \citep{ghazvininejad2019mask, gong2022diffuseq, li2022diffusionlm, dieleman2022continuous}. However, the architectural paradigm of text diffusion offers several benefits that extend beyond computational efficiency. Here, we demonstrate the practical advantages through concrete examples and qualitative analysis of samples generated by the model. To isolate the intrinsic capabilities of the architecture, we disable explicit thinking modes in both DiffusionGemma and the baseline Gemma AR model. This allows us to observe and evaluate the underlying generative process.

\subsection{Bidirectional Reasoning and Self-Correction}
AR next-token prediction is inherently causal; during the generation of a given token, the model can only attend to the preceding context. It fundamentally lacks the capacity to condition upon tokens that have yet to be generated. By contrast, text diffusion operates with full bidirectional attention across the canvas. This non-causal property allows tokens at arbitrary positions across the canvas to attend simultaneously to both past and future representations---a critical capability for complex planning and reasoning tasks where early decisions depend on eventual outcomes \citep[e.g.,][]{papadopoulos2024arrows, kitouni2024factorization, zhangli2024reverse, nagarajan2025roll}. Consequently, locally within a canvas, future tokens can directly influence the formation of earlier tokens.

Because text diffusion employs an iterative refinement process, it possesses a built-in mechanism for self-correction. Any premature commitments generated in earlier denoising steps can be revised during subsequent denoising iterations \citep{reid2022diffuser, han2023ssdlm}. While we have previously demonstrated the efficacy of bidirectional reasoning and self-correction in structured logic puzzles like Sudoku, we demonstrate these mechanics in two reasoning examples.

\begin{figure}[htbp]
\begin{tcolorbox}[
  colback=gray!5,
  colframe=gray!50,
  fontupper=\small,
  title={\sffamily\bfseries Prompt}
]
``Solve $7 \times (\sqrt{111 + 10} - 3) - 9^2$. Give the answer first
then provide the reasoning, be concise.''

\medskip
\begin{tcolorbox}[colback=red!3, colframe=red!25, boxrule=0.4pt, left=4pt, right=4pt]
\textbf{Gemma~4 26B A4B:}
$-1$

\textbf{Reasoning:}
\begin{enumerate}[nosep, leftmargin=1.5em]
  \item Inside parentheses: $111 + 10 = 121$.
  \item Square root: $\sqrt{121} = 11$.
  \item Inside parentheses continued: $11 - 3 = 8$.
  \item Multiplication: $7 \times 8 = 56$.
  \item Exponent: $9^2 = 81$.
  \item Subtraction: $56 - 81 = -25$.
\end{enumerate}
Correction: Wait, recalculating: $56 - 81 = -25$.

Correct Answer: $-25$
\end{tcolorbox}

\smallskip
\begin{tcolorbox}[colback=green!3, colframe=green!30, boxrule=0.4pt, left=4pt, right=4pt]
\textbf{DiffusionGemma:}
$-25$

\textbf{Reasoning:}
\begin{enumerate}[nosep, leftmargin=1.5em]
  \item Simplify inside the square root: $\sqrt{111 + 10} = \sqrt{121} = 11$.
  \item Solve inside the parentheses: $(11 - 3) = 8$.
  \item Multiply: $7 \times 8 = 56$.
  \item Calculate the exponent: $9^2 = 81$.
  \item Final subtraction: $56 - 81 = -25$.
\end{enumerate}
\end{tcolorbox}

\smallskip
{\footnotesize \textbf{Analysis:} Constrained by causal generation, Gemma~4 commits to the incorrect answer ``$-1$'' in its very first tokens. As it autoregressively generates its reasoning, it discovers its own error and is forced to append a correction. Conversely, DiffusionGemma produces the correct answer ``$-25$'' immediately, preventing logical contradictions. Note that both models are operating in non-thinking mode.}
\end{tcolorbox}

\vspace{1em}
\centering
\includegraphics[clip, viewport=0bp 0bp 500bp 230bp, width=0.9 \linewidth]{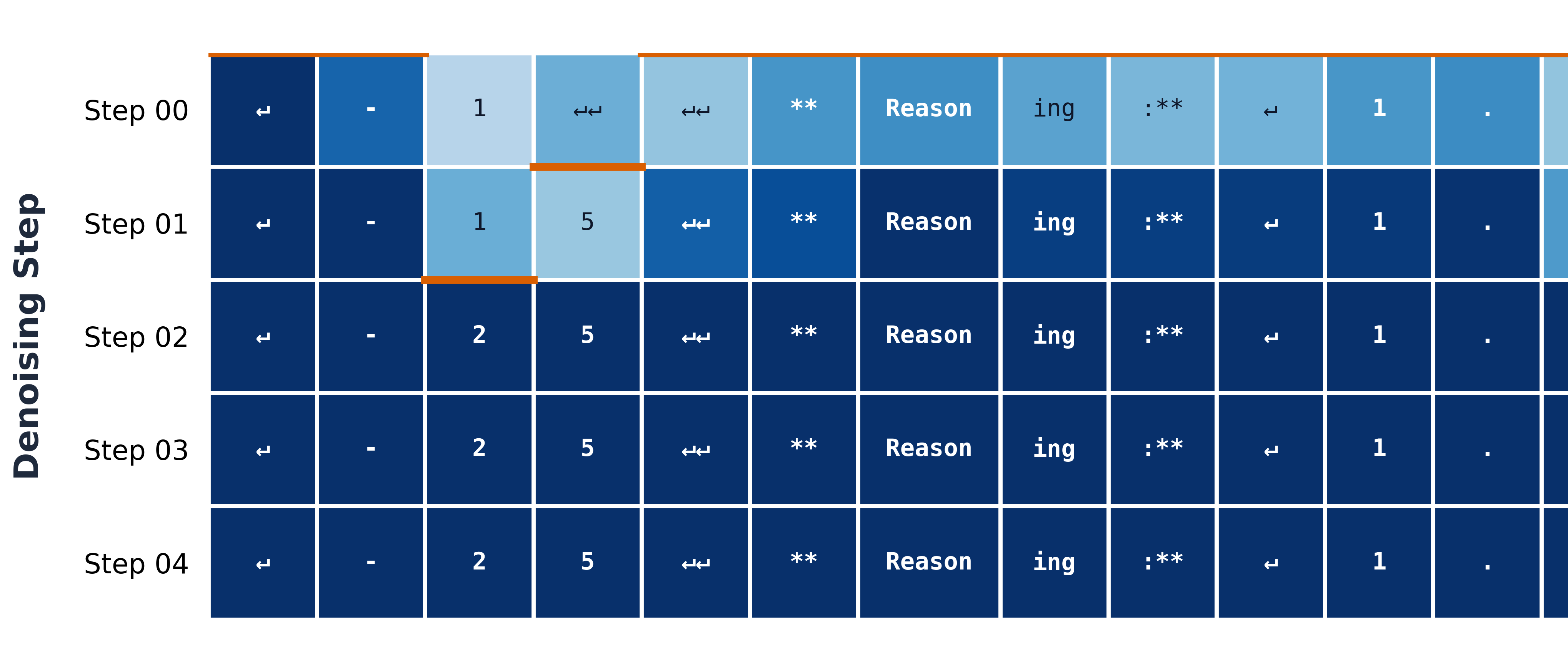}
\caption{\textbf{Denoising trace for the math reasoning problem (zoomed in to the first 12 tokens), where color intensity indicates model confidence}. DiffusionGemma initially makes the same causal estimation mistake as the AR model, assigning high probability to $-1$. However, through iterative bidirectional refinement, it suggests $-15$ in the next step before fully converging to the correct answer $-25$. The parallel denoising process allows the final answer and reasoning steps to inform one another. Denoising is completed in just 5 steps.}
\label{fig:math_reasoning}
\end{figure}

Consider the multi-step arithmetic problem detailed in Figure \ref{fig:math_reasoning}. Constrained by causal generation, the AR Gemma model must commit to the first token of the final answer before articulating the intermediate calculations. This forces an incorrect initial prediction ($-1$) that the model must later correct. DiffusionGemma, conversely, leverages its bidirectional attention canvas to simultaneously evolve the final answer and its underlying logic. As visualized in the accompanying denoising trace, the parallel diffusion process explores incorrect intermediate states but uses the emerging reasoning tokens to course-correct, seamlessly converging on the correct answer ($-25$). See Appendix~\ref{app:advantages_bidir} for the second example of logical reasoning and correction via bidirectional attention and denoising.

\subsection{Dynamic and Adaptive Computation}

Unlike AR models, which inherently allocate a fixed amount of computation per generated token, text diffusion facilitates dynamic test-time compute. Text diffusion models can autonomously calibrate the computational effort expended on a given prompt, trading increased inference time for enhanced generative performance based on the inherent difficulty of the task \citep{ye2024diffusion}. DiffusionGemma automatically adapts to task difficult via adaptive stopping. As illustrated in Figure~\ref{fig:boxplot_evals_effective_denoising_steps}, tasks of varying complexity organically elicit different numbers of effective denoising steps. This allows the model to conserve compute on easier queries while expending more compute on complex reasoning. In Appendix~\ref{app:advantages_adaptive} we concretely illustrate the adaptive behavior of text diffusion by contrasting a structurally “hard” generation task against a structurally “easy” task and show that the easy task requires fewer denoising steps, as well as showing how the information propagates differently across the canvas of tokens in these two cases.

Furthermore, the maximum number of denoising steps serves as an explicit configuration parameter to manage the latency-quality tradeoff. A lower step count forces the model to traverse the reverse process more coarsely, yielding faster results with a corresponding reduction in output precision. Expanding the step count allows for a finer resolution during generation, maximizing quality while extending the required compute time. Figure~\ref{fig:score_vs_dns} plots this continuous relationship, demonstrating how the model can be dynamically calibrated to suit varied operational constraints.

\subsection{Structured and Constrained Outputs}
In many practical applications, the desired output adheres to a rigid, highly structured format (e.g., JSON schemas) or exhibits strong lexical dependence on the input prompt, such as in optical character recognition (OCR), code editing tasks, or fine-grained syntactic control \citep{chen2023textdiffuser, li2022diffusionlm}. Because text diffusion generates and refines all tokens in parallel, it seamlessly exploits these structural priors to accelerate convergence. AR models lack this capability; constrained by strict sequential decoding, they must incur the same $O(N)$ computational cost to generate $N$ tokens, even when those tokens consist of rigid boilerplate or verbatim copies of the input context. By contrast, DiffusionGemma can identify and lock in predictable syntactic structures across the entire sequence simultaneously. We highlight this efficiency through two real-world examples: strict JSON extraction (Appendix~\ref{app:advantages_structure}, Figures \ref{fig:json-prompt} and \ref{fig:json-denoising}) and Python code debugging (Appendix~\ref{app:advantages_structure}, Figures \ref{fig:python-prompt} and \ref{fig:bug-fix}). Both cases clearly illustrate that the highly constrained output allows the diffusion process to converge in merely two to three steps, drastically reducing latency compared to sequential decoding.

\section{Limitations \& Known Issues} \label{sec:limitations}

In the previous section, we discussed new emergent properties of text diffusion and advantages of our approach relative to AR language modeling. While DiffusionGemma establishes a new Pareto frontier for generative text modeling and inference efficiency, the current experimental release has these known limitations:

\begin{itemize}[topsep=0pt]

    \item \textbf{Performance gap relative to the AR baseline:} Lower absolute performance than its AR initialization (Gemma 4 26B A4B) stems from several practical constraints: bypassing native diffusion pretraining to warm-start from AR weights; relying on a comparatively short SFT phase due to compute budget constraints; using an online learning algorithm (\sdrl) that explicitly targets ultra-low latency, intrinsically trading off asymptotic performance; and inheriting architectural, optimization, and data-mixture decisions from the AR baseline that may be suboptimal for the discrete diffusion paradigm.

    \item \textbf{Generation length and conciseness:} As mentioned in Section~\ref{sec:sdrl}, our final checkpoint produces highly concise outputs. While this emergent brevity acts as a multiplier for inference speed, it precludes the model from leveraging the quality improvements typically unlocked by longer, more elaborate reasoning traces.

    \item \textbf{Occasional token stuttering:} In rare instances, the model's output degenerates into repetitive loops or localized stuttering (e.g., endlessly repeating a common token like ``the the the''). While our \sdrl training successfully mitigates the vast majority of such instances, this uncommon artifact remains a direct consequence of operating in an ultra-low latency regime, where the aggressively reduced number of denoising steps can occasionally compromise the robustness of the generation process.
    
    \item \textbf{Occasional omission of closing thought tags in multimodal tasks:} When processing multimodal prompts, the model does not always reliably generate a closing thought tag (even when the reasoning is correct). This can artificially drag down performance in thinking mode on specific benchmarks; for example, on MMMU-Pro, the thinking score drops below the non-thinking score (54.3 vs.\ 66.0). This was identified too late in the pipeline to apply a fix for this release.

    \item \textbf{Throughput limits at high batch sizes:} DiffusionGemma excels at low batch sizes by trading memory-bandwidth costs for compute, outperforming Gemma~4~AR (with MTP) in both per-user and total throughput for up to $\sim$32 concurrent users (Figure~\ref{fig:inference-throughput}). Beyond this point, the higher per-token compute cost causes AR models to gain a throughput advantage. As noted in Section~\ref{sec:inference_optimizations}, these results are obtained without targeted batch-size optimization; a thorough empirical analysis under realistic traffic conditions remains future work.

\end{itemize}

\section{Conclusion} \label{sec:conclusion}
DiffusionGemma demonstrates a practical, compute-efficient path to ultra-fast text generation. By finetuning the existing Gemma 4 26B A4B AR model to perform text diffusion through our two-stage training pipeline (SFT and \sdrl), it establishes a new Pareto frontier for the speed-to-intelligence tradeoff---achieving around 1,500 tokens per second on a single H100 while retaining highly competitive reasoning and multimodal capabilities. By releasing DiffusionGemma as an experimental open-weight model---alongside reference implementations in HuggingFace Transformers and vLLM---we aim to empower the open-source community to push the boundaries of text diffusion. We hope researchers and practitioners will build upon this approach, whether by finetuning for specialized tasks, exploring novel sampling algorithms, or further optimizing inference efficiency.

\bibliography{main}

\newpage

\appendix

\phantomsection
\section*{Contributions and Acknowledgments (listed alphabetically)}
\label{sec:contributions}

\noindent\textbf{Core contributors (workstream leads marked with ‘*’)}

\begin{multicols}{3}
\noindent
Adrien Ali Taïga \\ 
James Assiene \\ 
Daniele Calandriello* \\ 
Rahma Chaabouni \\ 
João Gante*  \\ 
Tamara von Glehn* \\ 
Nate Keating \\ 
Chris Knutsen \\ 
Martin Kukla* \\ 
Tianlin Liu \\ 
Ivan Lobov* \\ 
Ofir Nabati \\ 
João Gabriel Oliveira \\ 
Nicolas Perez-Nieves* \\ 
Nastasia Prutianova \\ 
Bobak Shahriari* \\ 
Jean Tarbouriech* \\ 
Pavel Tyletski \\ 
Çağlar Ünlü \\ 
Cindy Wu
\end{multicols}

\noindent\textbf{Contributors}

\begin{multicols}{3}
\noindent
Glenn Cameron \\ 
Jerome Connor \\ 
Sertan Girgin \\ 
Maarten Grootendorst \\ 
Alon Levkovitch \\ 
Eliya Nachmani \\ 
Omar Sanseviero \\ 
Piotr Stanczyk
\end{multicols}

\noindent\textbf{Finetuning framework}

\begin{multicols}{3}
\noindent
Quentin Berthet \\ 
Andrew Campbell \\ 
Clément Crepy \\ 
Valentin De Bortoli \\ 
Arnaud Doucet \\ 
Romuald Elie \\ 
Alexandre Galashov \\ 
Klaus Greff \\ 
Alexis Jacq \\ 
David Ruhe \\ 
Yu-Han Wu
\end{multicols}

\noindent\textbf{Leads}

\begin{multicols}{3}
\noindent
Sebastian Flennerhag \\ 
Brendan O'Donoghue \\ 
George Scrivener \\ 
Shantanu Thakoor
\end{multicols}

\noindent\textbf{Acknowledgements}

\begin{multicols}{3}
\noindent
Sander Dieleman \\
Lucas Dixon \\
Johan Ferret \\
Parnian Kassraie \\
Preethi Lahoti \\
Gaël Liu \\
Sarah Perrin \\
Angéline Pouget \\
Louis Rouillard \\
Pier Giuseppe Sessa \\
Danilla Sinopalnikov \\
Gemma Team \\
\end{multicols}

\noindent\textbf{Sponsors}

\begin{multicols}{3}
\noindent
Olivier Bachem \\ 
Jeff Dean \\ 
Zoubin Ghahramani \\ 
Raia Hadsell \\ 
Demis Hassabis \\ 
Prateek Jain \\ 
Armand Joulin \\ 
Koray Kavukcuoglu \\ 
Marc’Aurelio Ranzato \\ 
Oriol Vinyals
\end{multicols}

\newpage

\section{Related work} \label{sec:related_work}

\subsection{Continuous Diffusion for Text}

While discrete diffusion remains highly effective, recent advances indicate a resurgence of continuous and hybrid approaches. Flow-matching frameworks like Embedded Language Flows have demonstrated that by abandoning per-step token supervision and remaining in an unrestricted continuous embedding space until a final discretization step, continuous models can substantially outperform discrete baselines \citep{hu2026elf}. Building on this continuous paradigm, recent advances in flow maps for discrete data have shown that by aligning training dynamics with the geometry of the probability simplex, these generative trajectories can be compressed into single-step mappings, achieving high-quality parallel language generation in one or a few steps \citep{lee2026flowmap, potaptchik2026discrete}. Similarly, models utilizing pretrained contextual autoencoders (e.g., Cosmos, TEncDM) map text into compressed, smooth latent spaces where Gaussian diffusion operates efficiently without rounding errors, providing a continuous manifold that lends itself well to advanced guidance techniques \citep{he2024manifold, meshchaninov2025cosmos, shabalin2025tencdm}, while recent approaches like hyperspherical flows avoid Gaussian corruption by rotating token embeddings on a hypersphere to better match the geometric structure of language \citep{deschenaux2026hyperspherical}. Other hybrid frameworks, such as CANDI \citep{pynadath2025candi}, address the ``temporal dissonance'' of applying Gaussian noise to discrete data by decoupling discrete and continuous corruption, allowing the model to simultaneously learn conditional structure and continuous geometry. Concurrently, Score Entropy Discrete Diffusion has bridged these domains by applying continuous-time Markov chains to learn the probability ratios of discrete data distributions \citep{lou2024discrete}. These breakthroughs suggest that the initial failures of continuous text diffusion may have been artifacts of sub-optimal spatial geometries and restrictive training objectives rather than inherent modality limitations.

\subsection{Speculative Decoding}

Speculative decoding~\citep{leviathan2023fast,chen2023accelerating,xia-etal-2024-unlocking} reduces serving latency by utilizing a smaller draft model to propose token sequences that are verified in parallel by a larger target model. The overall latency depends on the generation time of both the draft and target models. AR drafters~\citep{li2026eagle} are fundamentally constrained by sequential generation: increasing draft quality requires more parameters, which increases per-token-latency and diminishes end-to-end gains. Parallel drafters such as Medusa~\citep{cai2024medusa} circumvent this by generating many tokens at once, but do so independently, yielding suboptimal acceptance rates. Diffusion-based drafters recover inter-token dependencies by instead modeling the joint distribution over draft tokens. TiDAR~\citep{liu2025tidar} uses a single model for diffusion-based drafting, and AR verification---a configuration that DiffusionGemma also supports---while DFlash~\citep{chen2026dflash} employs a separate model as the drafter. However, it exhibits suffix decay, with acceptance rates declining at later draft positions~\citep{cheng2026dspark}. DSpark~\citep{cheng2026dspark} addresses this by adding a lightweight AR module on top of the parallel backbone. In contrast, DiffusionGemma operates as a standalone diffusion model, eliminating the verification bottleneck and scaling to longer generated canvases than draft-then-verify approaches.

\section{Samples Before and After \texorpdfstring{\sdrl}~Training}\label{sec:samples_before_after_sdrl}

To illustrate the impact of our \sdrl phase, Figures~\ref{fig:gpqa-rl-example-biology}, \ref{fig:gpqa-rl-example-quantum} contrast representative generations from the SFT model and the final checkpoint. They show how \sdrl training resolves the severe repetitive looping observed in the SFT baseline, allowing the model to complete complex reasoning traces.

\definecolor{promptbg}{gray}{0.95}
\definecolor{promptframe}{gray}{0.5}
\definecolor{afterbg}{HTML}{F0FAF0}
\definecolor{afterframe}{HTML}{2E7D32}
\definecolor{beforebg}{HTML}{FFF5F5}
\definecolor{beforeframe}{HTML}{C62828}
\definecolor{thinkbg}{HTML}{FAFAFA}
\definecolor{thinkframe}{gray}{0.75}
\mdfdefinestyle{promptbox}{%
  backgroundcolor=promptbg,
  linecolor=promptframe,
  linewidth=0.8pt,
  innertopmargin=8pt,
  innerbottommargin=8pt,
  innerleftmargin=8pt,
  innerrightmargin=8pt,
  skipabove=6pt,
  skipbelow=4pt
}
\mdfdefinestyle{afterbox}{%
  backgroundcolor=afterbg,
  linecolor=afterframe,
  linewidth=1pt,
  innertopmargin=8pt,
  innerbottommargin=8pt,
  innerleftmargin=8pt,
  innerrightmargin=8pt,
  skipabove=6pt,
  skipbelow=4pt
}
\mdfdefinestyle{beforebox}{%
  backgroundcolor=beforebg,
  linecolor=beforeframe,
  linewidth=1pt,
  innertopmargin=8pt,
  innerbottommargin=8pt,
  innerleftmargin=8pt,
  innerrightmargin=8pt,
  skipabove=6pt,
  skipbelow=4pt
}
\mdfdefinestyle{thinkbox}{%
  backgroundcolor=thinkbg,
  linecolor=thinkframe,
  linewidth=0.5pt,
  innertopmargin=6pt,
  innerbottommargin=6pt,
  innerleftmargin=6pt,
  innerrightmargin=6pt,
  skipabove=4pt,
  skipbelow=4pt
}

\begin{figure*}[t]
\scriptsize
\begin{mdframed}[style=promptbox]
\noindent\textbf{Prompt}
You perform a high-throughput experiment on white lupine to find genes contributing to resistance to the fungal disease anthracnose. You receive three candidate genes of unknown function~-- G1, G2, and G3. You create three knock-out mutants (g1, g2, g3) and double-mutants (g1g2, g1g3, g2g3). You know that at least one of these genes is a transcription factor acting upstream of other gene(s). After testing with the pathogen, you receive the following results (100\% = wild-type resistance):

\vspace{2pt}
\begin{tabular}{@{}lr@{\hspace{6pt}}lr@{\hspace{6pt}}lr@{}}
\toprule
Genotype & Resistance & Genotype & Resistance & Genotype & Resistance \\
\midrule
g1 & 75\% & g1g3 & 10\% & g1g2 & 0\% \\
g2 & 0\%  & g2g3 & 0\%  &      &     \\
g3 & 50\% &      &      &      &     \\
\bottomrule
\end{tabular}
\vspace{2pt}

\noindent Which conclusion can you draw?

(A)~G2 is a TF, G1\&G3 show pleiotropy, G1 epistatic towards G3

(B)~G2 is a TF, G1\&G3 show gene redundancy, G1 epistatic towards G3

(C)~G2 is a TF, G1\&G3 share a promoter, G3 epistatic towards G1

(D)~G1 is a TF, G2\&G3 show pleiotropy, G2 epistatic towards G1
\hfill\textbf{\textit{Correct answer:}~(B)}
\end{mdframed}
\vspace{2pt}
\noindent
\begin{minipage}[t]{0.49\textwidth}
\begin{mdframed}[style=beforebox]
\noindent\textbf{Before \sdrlfootnotesize} \hfill {\colorbox{beforebg}{\textcolor{beforeframe}{\textbf{$\boldsymbol\times$~Incorrect (empty)}}}}

\noindent\textit{Thinking (truncated):}
\begin{mdframed}[style=thinkbox]
\begin{itemize}[leftmargin=12pt, labelsep=4pt, itemsep=0pt, parsep=0pt, topsep=0pt]
\item Wild-type (control): 100\% resistance.
  \begin{itemize}[leftmargin=12pt, labelsep=4pt, itemsep=0pt, parsep=0pt, topsep=0pt]
  \item g1 mutant: 75\% resistance (slight decrease).
  \item g2 mutant: 0\% resistance (complete loss).
  \item g3 mutant: 50\% resistance (moderate decrease).
  \item g1g3 double mutant: 10\% resistance (significant decrease, more than g1 or g3 alone).
  \item g2g3 double mutant: 0\% resistance (same as g2).
  \item g1g2 double mutant: 0\% resistance (same as g2). [...]
  \item If G2 is a transcription factor (TF) that activates G1 and G3:
    \begin{itemize}[leftmargin=12pt, labelsep=4pt, itemsep=0pt, parsep=0pt, topsep=0pt]
    \item If G2 is G, G1 and G3 are targets.
    \item If G2 is knocked out, G1 and G3 are not expressed.
    \item So, g2 would have the same phenotype as g2g1 and g2g3.
    \item This matches the data: g2 (0\%), g2g1 (0\%), g2g3 (0\%).
    \end{itemize}
  \item Now, let's look at G1 and G3 again:
    \begin{itemize}[leftmargin=12pt, labelsep=4pt, itemsep=0pt, parsep=0pt, topsep=0pt]
    \item g1: 75\%
    \item g3: 50\%
    \item g1g3: 10\%
    \item If G1 and G3 act in parallel pathways ( pathways,, loss111111111111111111111 G11 G G11 G G G1 G G1 G G G G G G G1 G1 G G1 G G G G G G G1 G G G G G G111 G111 G G G1111 G111111111 G1111 G G1111111111111111111
    \end{itemize}
  \end{itemize}
\end{itemize}
{\itshape [\ldots repeated 1 tokens until the max tokens budget.]}
\end{mdframed}
\vspace{2pt}
\noindent\textit{Response:}
\begin{mdframed}[style=thinkbox]
\textit{[empty --- no response produced]}
\end{mdframed}
\end{mdframed}
\end{minipage}%
\hfill
\begin{minipage}[t]{0.49\textwidth}
\begin{mdframed}[style=afterbox]
\noindent\textbf{After \sdrlfootnotesize} \hfill {\colorbox{afterbg}{\textcolor{afterframe}{\textbf{\checkmark~Correct}}}}

\noindent\textit{Thinking (truncated):}
\begin{mdframed}[style=thinkbox]
\begin{itemize}[leftmargin=12pt, labelsep=4pt, itemsep=0pt, parsep=0pt, topsep=0pt]
\item Wild-type (control): 100\% resistance.
  \begin{itemize}[leftmargin=12pt, labelsep=4pt, itemsep=0pt, parsep=0pt, topsep=0pt]
  \item g1: 75\% resistance (slightly reduced).
  \item g2: 0\% resistance (completely sensitive).
  \item g3: 50\% resistance (moderately reduced).
  \item g1g3: 10\% resistance (highly sensitive).
  \item g2g3: 0\% resistance (completely sensitive).
  \item g1g2: 0\% resistance (completely sensitive).
  \end{itemize}
\end{itemize}
g2 has 0\% resistance. This means G2 is essential for the resistance. g1g3 has 10\%, much lower than both g1 (75\%) and g3 (50\%). This suggests that G1 and G3 work in parallel pathways (redundancy).

\noindent If G2 is a TF upstream of G1 and G3, then knocking out G2 (g2) would prevent the expression of both G1 and G3. This would explain why g2, g2g3, and g1g2 all have 0\% resistance.

\vspace{2pt}
\noindent\textit{[...re-examines epistasis definitions, checks all options, confirms (B)...]}
\vspace{2pt}

\noindent Given the options, (B) is the only one that correctly identifies G2 as a transcription factor and G1/G3 as redundant.
\end{mdframed}
\vspace{2pt}
\noindent\textit{Response:}
\begin{mdframed}[style=thinkbox]
To determine the correct conclusion regarding the genes G1, G2, and G3, we analyze the experimental resistance data provided:
\begin{enumerate}[leftmargin=12pt, labelsep=4pt, itemsep=0pt, parsep=0pt, topsep=2pt]
\item \textbf{Wild-type (control):} 100\% resistance.
\item \textbf{g1 mutant:} 75\% resistance (G1 contributes slightly to resistance). \textit{[...]}
\end{enumerate}

\noindent\textbf{Analysis of the interactions:}
\begin{itemize}[leftmargin=12pt, labelsep=4pt, itemsep=0pt, parsep=0pt, topsep=2pt]
\item \textbf{Transcription Factor:} If G2 is a transcription factor upstream of both G1 and G3, knocking out G2 would prevent the expression of both G1 and G3, \textit{[...]}
\end{itemize}

\vspace{2pt}
\noindent\textbf{Final Answer: (B)}
\end{mdframed}
\end{mdframed}
\end{minipage}
\caption{\textbf{Example from GPQA-Diamond (Biology) showing model behavior with the DiffusionGemma sampler before and after \sdrl training}. The SFT model begins with \emph{correct} reasoning---identifying G2 as an essential transcription factor and recognizing G1/G3 redundancy---but degenerates into a repetitive token loop (the digit `1'), producing an empty final response. After \sdrl, the model completes the analysis cleanly and outputs the correct answer.}
\label{fig:gpqa-rl-example-biology}
\end{figure*}

\begin{figure*}[t]
\scriptsize
\begin{mdframed}[style=promptbox]
\noindent\textbf{Prompt}
\vspace{2pt}\noindent
A quantum mechanical particle of mass $m$ moves in two dimensions in the following potential, as a function of $(r, \theta)$:
\[
V(r, \theta) = \tfrac{1}{2}\,k\,r^2 + \tfrac{3}{2}\,k\,r^2\cos^2(\theta)
\]
Find the energy spectrum.

\vspace{2pt}
\noindent
\begin{tabular}{@{}ll@{}}
(A) & $E = (n_x + 3n_y + \tfrac{3}{2})\,\hbar\sqrt{k/m}$ \\[2pt]
(B) & $E = (2n_x + n_y + \tfrac{3}{2})\,\hbar\sqrt{k/m}$ \\[2pt]
(C) & $E = (3n_x + 2n_y + \tfrac{1}{2})\,\hbar\sqrt{k/m}$ \\[2pt]
(D) & $E = (2n_x + 3n_y + \tfrac{1}{2})\,\hbar\sqrt{k/m}$
\end{tabular}

\hfill\textbf{\textit{Correct answer:} (B)}
\end{mdframed}
\vspace{2pt}
\noindent
\begin{minipage}[t]{0.49\textwidth}
\begin{mdframed}[style=beforebox]
\noindent\textbf{Before \sdrlfootnotesize} \hfill {\colorbox{beforebg}{\textcolor{beforeframe}{\textbf{$\boldsymbol\times$~Incorrect (empty)}}}}

\noindent\textit{Thinking (truncated):}
\begin{mdframed}[style=thinkbox]
The potential is given in polar coordinates $(r, \theta)$ as:
$V(r, \theta) = \frac{1}{2}kr^2 + \frac{3}{2}kr^2\cos^2(\theta)$.
We know:

\quad $x = r\cos(\theta)$,

\quad $y = r\sin(\theta)$,

\quad $r^2 = x^2 + y^2$,

\quad $\cos^2(\theta) = x^2/r^2$.

\smallskip
Substitute these into the potential expression:

\quad $V(x,y) = \frac{1}{2}k(x^2+y^2) + \frac{3}{2}kx^2 = 2kx^2 + \frac{1}{2}ky^2$

\textit{[...]}

Plugging in the values for $\omega_x$ and $\omega_y$:

\textit{[...]}

The calculated energy spectrum $E = (2n_x + n_y + \tfrac{3}{2})\,\hbar\sqrt{k/m}$, which matches option (B).

\medskip
\textit{[\ldots re-verifies answer three additional times, all correctly arriving at (B) \ldots]}

\medskip
\noindent Let's re-read the options and
\quad .\qquad = ( = =xx\_nn\_\_33/ \qquad ))))
\smallskip
)))))))222\_\_\_\_y +//)) ))) ))))))))))))))2 2 2222\\
)))))))))))))))))))))))))))))))))))))))))))))))))\\
{\itshape [\ldots repeated closing-parenthesis until max token budget.]}
\end{mdframed}
\vspace{2pt}
\noindent\textit{Response:}
\begin{mdframed}[style=thinkbox]
\textit{[empty --- no response produced]}
\end{mdframed}
\end{mdframed}
\end{minipage}%
\hfill
\begin{minipage}[t]{0.49\textwidth}
\begin{mdframed}[style=afterbox]
\noindent\textbf{After \sdrlfootnotesize} \hfill {\colorbox{afterbg}{\textcolor{afterframe}{\textbf{\checkmark~Correct}}}}

\noindent\textit{Thinking (truncated):}
\begin{mdframed}[style=thinkbox]
The potential is $V(r, \theta) = \frac{1}{2}kr^2 + \frac{3}{2}kr^2\cos^2(\theta)$.

\quad The coordinates are polar $(r, \theta)$.

\quad $x = r\cos\theta$

\quad $y = r\sin\theta$

\quad $r^2 = x^2 + y^2$

\quad $\cos\theta = \frac{x}{r}$

\quad $V(r, \theta) = \frac{1}{2}k(x^2+y^2) + \frac{3}{2}k(x^2+y^2)(\frac{x}{r})^2$

\textit{[\ldots]}

The Hamiltonian is separable into two one-dimensional harmonic oscillators:

\quad $H = H_x + H_y$

\textit{[\ldots]}

For a 1D harmonic oscillator $H = -\frac{\hbar^2}{2m}\frac{d^2}{dx^2} + \frac{1}{2}m\omega^2x^2$, the energy eigenvalues are:
    $E = (n+\frac{1}{2})\hbar\omega$, where $n = 0, 1, 2, ...$

\textit{[\ldots]}

The calculated $E = (2n_x+n_y+3/2)\hbar\sqrt{k/m}$ matches option (B).
   
\textit{[\ldots verifies calculation twice, confirms match with option (B).]}
\end{mdframed}
\smallskip
\noindent\textit{Response (excerpt):}
\begin{mdframed}[style=thinkbox]
To find the energy spectrum of the quantum mechanical particle of mass $m$ in the given potential $V(r, \theta) = \frac{1}{2}kr^2 + \frac{3}{2}kr^2\cos^2(\theta)$, we first convert the potential into Cartesian coordinates $(x, y)$.

1.  \textbf{Convert the potential to Cartesian coordinates:}
    The relationship between polar and Cartesian coordinates is:

\textit{[\ldots]}



2.  \textbf{Identify the Hamiltonian:}
    The Hamiltonian for a particle in two dimensions is: \textit{[\ldots]}


3.  \textbf{Calculate the energy eigenvalues for each oscillator:}
    The energy levels of a 1D harmonic oscillator $\frac{1}{2}m\omega^2x^2$ are $E = (n+\frac{1}{2})\hbar\omega$.

\textit{[\ldots]}

4.  \textbf{Find the total energy:}

\textit{[\ldots]}

\quad $E = (2n_x + n_y + \frac{3}{2})\hbar\sqrt{\frac{k}{m}}$

\textbf{Final Answer: (B)}
\end{mdframed}
\end{mdframed}
\end{minipage}
\caption{\textbf{Example from GPQA-Diamond (2D harmonic oscillator energy spectrum) showing model behavior with the DiffusionGemma sampler before and after \sdrl training}. The SFT model begins with a \emph{correct} derivation---converting to Cartesian coordinates, identifying the separable Hamiltonian, and computing the correct energy spectrum---but degenerates into a repetitive token loop (closing parentheses), producing an empty final response. After \sdrl, the model completes the same derivation cleanly and outputs the correct answer.}
\label{fig:gpqa-rl-example-quantum}
\end{figure*}

\section{Additional Open-Source Downstream Finetuning Results}
\label{sec:sft-hd-appendix}

In Section \ref{sec:downstream_sft}, we describe our finetuning strategy as well as our main results on Sudoku puzzle solving. We now  provide full training recipes as well as additional results for PubMedQA~\citep{jin2019pubmedqa}. 

\paragraph{Sudoku solving trace summary.} In Figure~\ref{fig:sudoku-rl-example}, we present one trace summary for a successful solving of a Sudoku puzzle.

\paragraph{Practical recipe summary.} Table~\ref{tab:sft_recipes} summarizes the key hyperparameters used in Sudoku and PubMedQA. Full training and evaluation code is available open-source via the Hackable Diffusion adapter. For our LoRA strategy to update all the linear layers, using rank 8 for Sudoku solving problem, we only finetune 8M parameters. 

\paragraph{Case study: PubMedQA.} To demonstrate applicability beyond structured reasoning, we finetune DiffusionGemma on PubMedQA~\citep{jin2019pubmedqa}, a biomedical question-answering benchmark where the model must read a medical research abstract and produce both a categorical answer (\emph{yes}, \emph{no}, or \emph{maybe}) and a detailed explanatory paragraph. The model is evaluated only on the long task using BLEU score against reference explanations, showing that DiffusionGemma can be effectively adapted to domain-specific natural language generation tasks with minimal data and compute.

\begin{table}[H]
\centering
\begin{tabular}{lccc}
\toprule
Model & Effective Denoising Steps & Accuracy (\%) & BLEU \\
\midrule
DiffusionGemma & 18.09 & 75.6 & 10.76 \\
+ LoRA finetuning & 31.57 & 76.62 & 20.67 \\
\bottomrule
\end{tabular}
\caption{\textbf{PubMedQA performance of the finetuned model with LoRA rank 4.} Finetuning leads to a slight increase in accuracy on a model with a good base performance.}
\label{tab:pubmedqa_steps_vs_perf}
\end{table}

\begin{table}[h]
\centering
\small
\begin{tabular}{lccc}
\toprule
\textbf{Hyperparameter} & \textbf{Sudoku (LoRA)} & \textbf{Sudoku (Full)} & \textbf{PubMedQA} \\
\midrule
LoRA rank & 8 & --- & 4 \\
Canvas size & 256 & 256 & 128 \\
Number of canvases & 1 & 1 & 2 \\
Prompt length & 256 & 256 & 1024 \\
Batch size & 2 & 8 & 2 \\
Peak learning rate & $3 \times 10^{-4}$ & $1.125\times 10^{-4}$ & $1.0 \times 10^{-4}$ \\
End learning rate & $3 \times 10^{-5}$ & $1.125\times 10^{-5}$ & $1.0 \times 10^{-5}$ \\
Training steps & 8{,}000 & 2{,}000 & 2{,}000 \\
Optimizer & Adam & Adafactor & Adam \\
LR schedule & Cosine with warmup & Cosine with warmup & Cosine with warmup \\
Warmup iterations & 400 & 100 & 100 \\
Weight decay & $10^{-4}$ & $10^{-4}$ & $10^{-4}$ \\
Min.\ hardware & 2$\times$ A100 80GB & 8$\times$ A100 80GB & 2$\times$ A100 80GB \\
\bottomrule
\end{tabular}
\caption{\textbf{Summary of downstream finetuning hyperparameters for the Sudoku and PubMedQA recipes.} Both LoRA recipes use adapters applied to all linear layers and train for 2{,}000 steps. A full-weight alternative is also provided for Sudoku, using Adafactor to manage memory on 8 GPUs.}
\label{tab:sft_recipes}
\end{table}

\definecolor{promptbg}{gray}{0.95}
\definecolor{promptframe}{gray}{0.5}
\definecolor{afterbg}{HTML}{F0FAF0}
\definecolor{afterframe}{HTML}{2E7D32}
\definecolor{beforebg}{HTML}{FFF5F5}
\definecolor{beforeframe}{HTML}{C62828}
\definecolor{thinkbg}{HTML}{FAFAFA}
\definecolor{thinkframe}{gray}{0.75}

\mdfdefinestyle{promptbox}{%
  backgroundcolor=promptbg,
  linecolor=promptframe,
  linewidth=0.8pt,
  innertopmargin=8pt,
  innerbottommargin=8pt,
  innerleftmargin=8pt,
  innerrightmargin=8pt,
  skipabove=6pt,
  skipbelow=4pt
}
\mdfdefinestyle{afterbox}{%
  backgroundcolor=afterbg,
  linecolor=afterframe,
  linewidth=1pt,
  innertopmargin=8pt,
  innerbottommargin=8pt,
  innerleftmargin=8pt,
  innerrightmargin=8pt,
  skipabove=6pt,
  skipbelow=4pt
}
\mdfdefinestyle{beforebox}{%
  backgroundcolor=beforebg,
  linecolor=beforeframe,
  linewidth=1pt,
  innertopmargin=8pt,
  innerbottommargin=8pt,
  innerleftmargin=8pt,
  innerrightmargin=8pt,
  skipabove=6pt,
  skipbelow=4pt
}
\mdfdefinestyle{thinkbox}{%
  backgroundcolor=thinkbg,
  linecolor=thinkframe,
  linewidth=0.5pt,
  innertopmargin=6pt,
  innerbottommargin=6pt,
  innerleftmargin=6pt,
  innerrightmargin=6pt,
  skipabove=4pt,
  skipbelow=4pt
}


\begin{figure*}[htbp]
\scriptsize
\begin{mdframed}[style=promptbox]
\noindent\textbf{Prompt}

\vspace{2pt}\noindent
\textbf{System:} Solve the following Sudoku puzzle. Empty cells are represented by 0. Output ONLY the solved puzzle immediately as a 9x9 grid of numbers separated by spaces. Do not include \#\#\#\#, explanations, or any other text.

\vspace{4pt}\noindent
\textbf{User:} 4 0 0 0 0 1 0 0 6 0 2 1 8 0 9 0 0 0 0 9 0 4 0 2 0 0 0 0 0 0 0 4 8 6 0 0 0 0 5 0 2 7 4 1 9 9 0 7 5 1 0 3 0 0 6 1 3 2 9 0 5 0 0 5 0 9 0 8 0 0 6 4 2 8 4 6 7 0 1 9 0
\end{mdframed}
\vspace{2pt}
\noindent
\begin{minipage}[t]{0.49\textwidth}
\begin{mdframed}[style=beforebox]
\noindent\textbf{Before Downstream Finetuning} \hfill {\colorbox{beforebg}{\textcolor{beforeframe}{\textbf{$\boldsymbol\times$~Incorrect }}}}

\noindent\textit{Thinking (truncated):}
\begin{mdframed}[style=thinkbox]
\texttt{<|turn>model} \\
\texttt{<|channel>thought} \\
\texttt{<channel|>}4 8 3 7 5 1 9 2 6 \\
7 2 1 8 6 9 5 3 4 \\
6 9 5 4 3 2 8 1 7 \\
1 4 8 9 9 6 3 7 5 \\
3 5 6 2 7 4 1 9 8 \\
9 7 2 5 1 3 4 6 1 \\
2 3 9 1 4 5 7 8 6 \\
5 6 7 1 2 8 4 4 3 \\
8 1 4 6 7 6 2 9 5 \\
\\
Wait, let me re-solve that carefully as the input string was a bit fragmented. \\
Input: \\
4 0 0 | 0 0 1 | 0 0 6 \\
0 2 1 | 8 0 9 | 0 0 0 \\
0 9 0 | 4 0 2 | 0 0 0 \\
0 0 0 | 0

{\itshape [\ldots truncated]}
\end{mdframed}
\vspace{2pt}
\noindent\textit{Response:}
\begin{mdframed}[style=thinkbox]
\textit{[empty --- answer cut]}
\end{mdframed}
\end{mdframed}
\end{minipage}%
\hfill
\begin{minipage}[t]{0.49\textwidth}
\begin{mdframed}[style=afterbox]
\noindent\textbf{After Downstream Finetuning} \hfill {\colorbox{afterbg}{\textcolor{afterframe}{\textbf{\checkmark~Correct}}}}

\noindent\textit{Thinking:}
\begin{mdframed}[style=thinkbox]
\textit{[empty --- successfully suppressed thought channel]}
\end{mdframed}
\vspace{2pt}
\noindent\textit{Response:}
\begin{mdframed}[style=thinkbox]
4 5 8 7 3 1 9 2 6 \\
3 2 1 8 6 9 7 4 5 \\
7 9 6 4 5 2 8 3 1 \\
1 3 2 9 4 8 6 5 7 \\
8 6 5 3 2 7 4 1 9 \\
9 4 7 5 1 6 3 8 2 \\
6 1 3 2 9 4 5 7 8 \\
5 7 9 1 8 3 2 6 4 \\
2 8 4 6 7 5 1 9 3
\end{mdframed}
\end{mdframed}
\end{minipage}
\caption{\textbf{Example evaluating formatting rule-following on a Sudoku puzzle before and after Open-Source Downstream Finetuning.} Before Open-Source Downstream Finetuning, the model fails to follow the strict negative constraints (``Output ONLY the solved puzzle immediately''), improperly attempting to build an erroneous grid in its thought channel and failing to provide a final response. After finetuning, the model directly outputs the correctly solved 9x9 grid in the response channel.}
\label{fig:sudoku-rl-example}
\end{figure*}

\section{Prompt Formatting} \label{sec:prompt_formatting}

To facilitate complex, multi-turn interactions, multimodal inputs, and agentic workflows, the model employs a structured Jinja2-based chat template. This template serializes the conversation history, system instructions, and tool schemas into a standardized string format using specialized control tokens. Explicit formatting ensures the model can accurately distinguish between user inputs, internal reasoning, tool invocations, and system-level contexts. What follows is a summary of the main features, but please refer to the implementation\footnote{\url{https://huggingface.co/google/diffusiongemma-26B-A4B-it/blob/main/chat_template.jinja}} for full details.

\subsection{BOS and EOS Special Tokens}

As in other Gemma models, every conversation must start with a special BOS token, which has no text rendering but corresponds to token integer 2. The BOS special token must either be manually prepended to the tokenized input or usually by the tokenizer via an \inlinecode{add\_bos=True} keyword argument. At the other end, when a block-AR generation completes, it ends in the usual special token \inlinecode{<turn|>} followed by padding with the EOS token. Similarly, the EOS token corresponds to the integer 1 but has no text rendering when detokenized.

\subsection{Conversational Structuring}

The template strictly delineates conversation turns using opening and closing tags. \inlinecode{<|turn>} and \inlinecode{<turn|>} tokens encapsulate individual messages, and the opening token is appended with the specific role (e.g., \inlinecode{<|turn>system\textbackslash n}). The model expects four roles: \inlinecode{system}, \inlinecode{user}, \inlinecode{model}, and \inlinecode{tool} for tool call responses.

The template natively supports multimodal routing by parsing content arrays for specific media types, injecting \inlinecode{<|image|>} tokens into the context stream where appropriate. In the forward pass of the model, these tokens are replaced by the input image features, as opposed to the corresponding token embedding.

\subsection{Thinking Channels}

As with the Gemma 4 models, DiffusionGemma supports thinking mode, which can be enabled by adding the thinking token \inlinecode{<|think|>} to the system instruction. When thinking is enabled, the model will emit an internal reasoning channel (which could technically still be empty) followed by the final answer:
\begin{lstlisting}
<|channel>thought
# DiffusionGemma's internal reasoning goes here.
<channel|>
# DiffusionGemma's final answer goes here
\end{lstlisting}
Importantly, even when the \inlinecode{<|think|>} token is not present in the system instruction, the model will still emit an empty thought channel as follows:
\begin{lstlisting}
<|channel>thought
<channel|>
[final answer]
\end{lstlisting}
For multi-turn conversations, do \textbf{not} include previous hidden thoughts in the conversation history. Only include the final assistant response before the next user turn.

\subsection{Tool and Function Calling Serialization}
A significant portion of the template is dedicated to parsing and serializing JSON-like tool schemas into a compact, token-efficient format. All JSON-like schemas, e.g. tool definitions or tool responses, rely on the custom delimiter \inlinecode{<|"|>} (as opposed to a plain \inlinecode{"}) for clarity.

\begin{itemize}[topsep=0pt]
    \item \inlinecode{<|tool>} and \inlinecode{<tool|>}: Used within the \textbf{system prompt} to define available function schemas, including their descriptions, parameters, and required arguments. 
    
    \item \inlinecode{<|tool\_call>} and \inlinecode{<tool\_call|>}: Model requests to invoke tools are formatted as \inlinecode{<|tool\_call>call:function\_name\{arguments\}<tool\_call|>}. 
    
    \item \inlinecode{<|tool\_response>} and \inlinecode{<tool\_response|>}: Results returned from external tools are appended to the context window wrapped by these tokens, allowing the model to seamlessly integrate external data into its subsequent turns.
\end{itemize}

\section{Mercury 2 Speed Estimation} \label{sec:competitor-eval-methodology}

\paragraph{Estimation data.} As Mercury 2 \citep{inception2026mercury2} is a closed source model, we do not have direct access to generate speed metrics such as TPF and TPS. We take a black-box approach and estimate TPS by querying OpenRouter's API \footnote{\url{https://openrouter.ai}} with the default maximum sequence (input+output) length of 50,000. We ran two independent rounds of queries — the first on July 9--10, 2026 (with a small number of retries on July 11), the second round between July 25--26, 2026 — and obtained consistent results across both measurements. Each API response includes the following metadata:
\begin{itemize}[topsep=0pt]
    \item the number of input tokens (prompt tokenization);
    \item how many prompt tokens were cached (KV-cache hits vs.\ fresh prefill);
    \item total output tokens (split into thinking tokens and answer tokens);
    \item the wall-clock time for the request (including network latencies and other overheads).
\end{itemize}

Response metadata does not include a breakdown of generation speed. We estimate generate TPS via a least-squares model, detailed below.

\paragraph{Estimation model.}
We estimate per-token generation speed by fitting the following model
via non-negative least-squares (NNLS):
\begin{equation}
    \text{Total Time} = \alpha \cdot \text{effective\_prefill\_tokens}
    + \beta \cdot \text{total\_output\_tokens} + \gamma,
    \label{eq:nnls-speed}
\end{equation}
where $\text{effective\_prefill\_tokens} = \text{prompt\_tokens} - \text{cached\_tokens}$ is the number of tokens requiring fresh computation, $\alpha$ is the per-token prefill time, $\beta$ is the per-token generation time, and $\gamma$ captures constant per-request overhead (network latency, etc.). From the fitted $\beta$, we obtain the generation speed as $1/\beta$ TPS. The non-negativity constraint reflects that processing tokens, generating tokens, and per-request overhead can only add time. In practice, we disable caching across queries.

In order to capture any potential per-task variation in generation speed (e.g. from adaptive computation methods), we fit Equation~\eqref{eq:nnls-speed} independently per benchmark, before taking an arithmetic average. Similarly, for open weight models we compute aggregate TPS measurements by first measuring per-task generative speed and then taking the arithmetic average.  

\paragraph{Speed estimates.}

\begin{figure}[!t]
    \centering
    \includegraphics[width=0.9\linewidth]{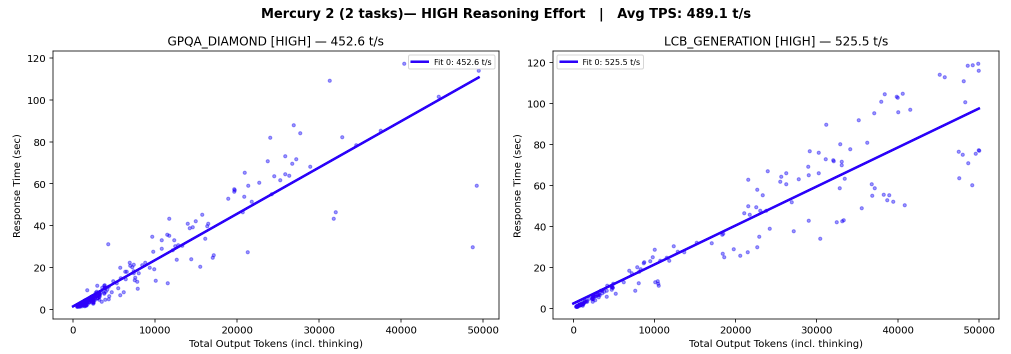}
    \caption{\textbf{Mercury 2 token throughput (two benchmarks, \texttt{high} reasoning effort).} Including outliers where the model produces no answer tokens (i.e.,\ outputting thinking tokens only). Evaluated on GPQA Diamond and LiveCodeBench-v6; used to compute the speed reported in Figure~\ref{fig:final-pareto}. The average of per-task TPS estimates yields 489~TPS.}
    \label{fig:mercury_speed_2evals}
\end{figure}

Our speed measurements for Figure~\ref{fig:final-pareto} are based on the GPQA-Diamond and LiveCodeBench-v6 datasets. For Mercury 2, we estimate the generative speed per task using the above methodology. Figure~\ref{fig:mercury_speed_2evals} shows the corresponding analysis; we obtain a TPS of 452.6 for GPA Diamond and 525.5 TPS for LiveCodeBench-v6, which translates into an average speed of 489~TPS.

\begin{figure}[!t]
    \centering
    \includegraphics[width=0.9\linewidth]{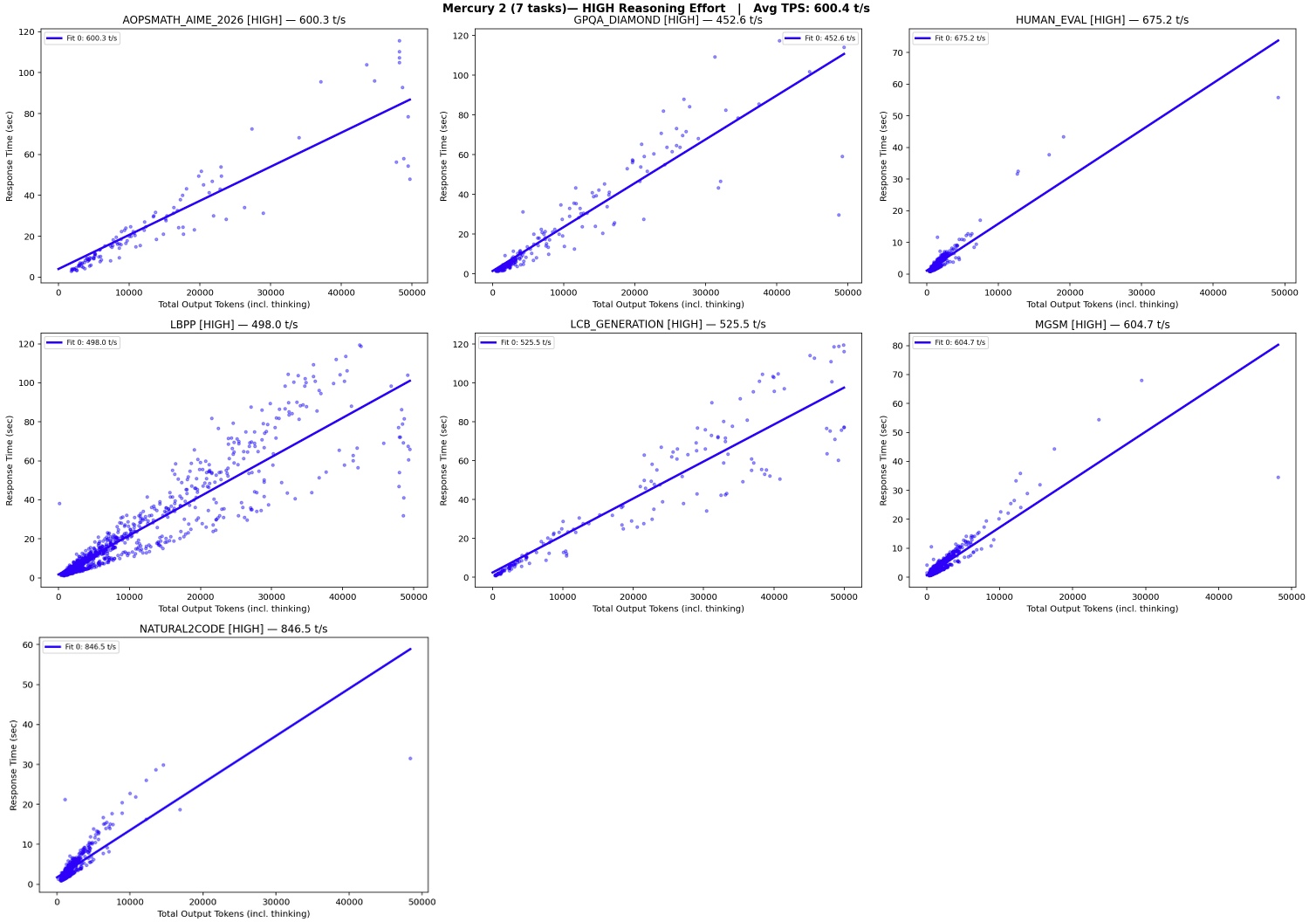}
    \caption{\textbf{Mercury 2 token throughput (seven benchmarks, \texttt{high} reasoning effort).} Including outliers where the model produces no answer tokens (i.e.,\ outputting thinking tokens only). Used to compute the \texttt{high} effort speed reported in Table~\ref{tab:model_comparison}. Average TPS estimates yields 600~TPS.}
    \label{fig:mercury_speed_7evals_high}
\end{figure}

\begin{figure}[!t]
    \centering
    \includegraphics[width=0.9\linewidth]{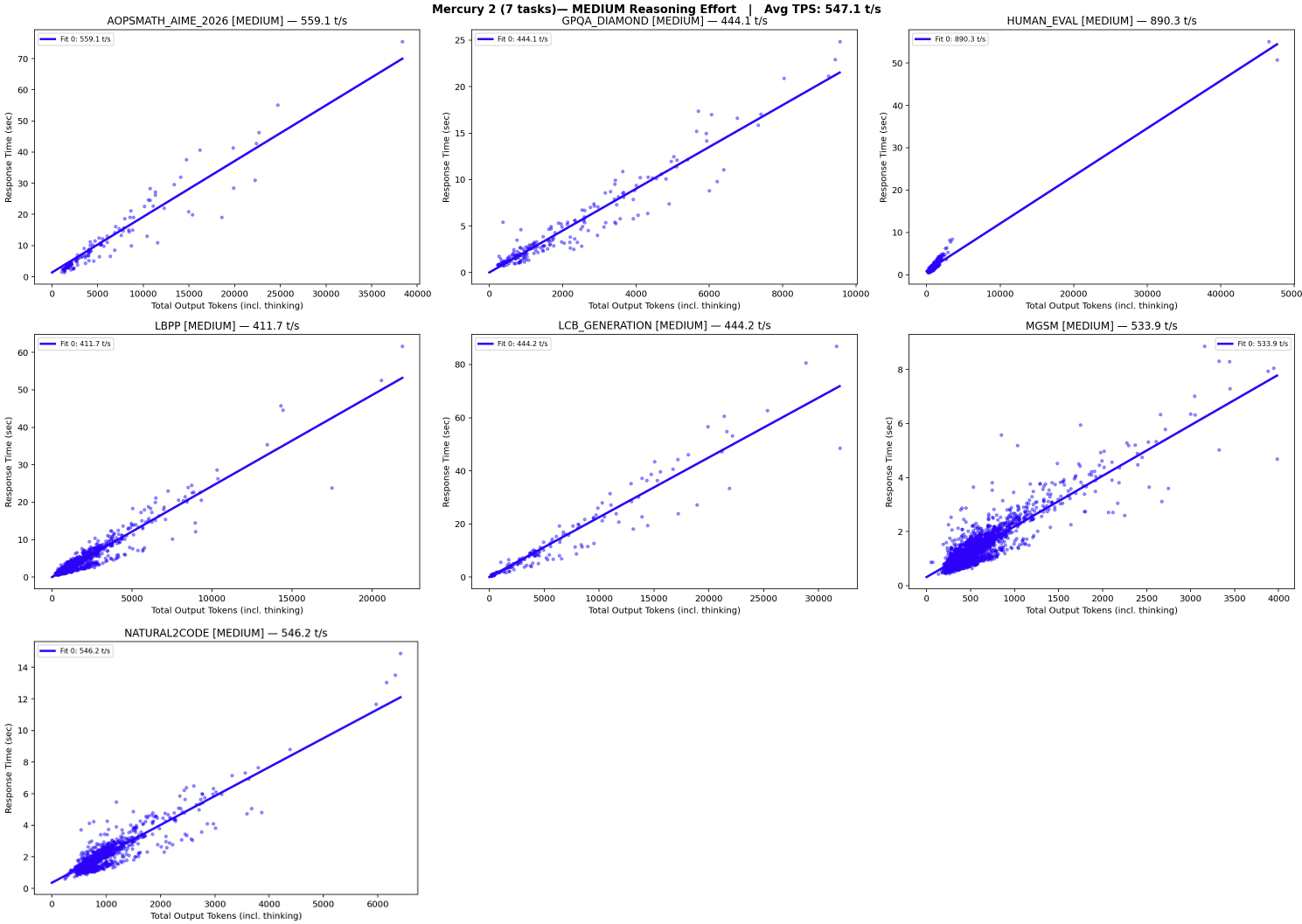}
    \caption{\textbf{Mercury 2 token throughput (seven benchmarks, \texttt{medium} reasoning effort).} Including outliers where the model produces no answer tokens (i.e.,\ outputting thinking tokens only). Used to compute the \texttt{medium} effort speed reported in Table~\ref{tab:model_comparison}. Average TPS estimates yields 547~TPS.}
    \label{fig:mercury_speed_7evals_medium}
\end{figure}

Our speed measurements for Table~\ref{tab:model_comparison} are based on seven benchmarks where we have measurements for all models: AIME 2026, GPQA Diamond, HumanEval, LBPP, LiveCodeBench-v6, MGSM, and Natural2Code. Again, for Mercury 2 we fit our speed estimation model on each task separately first before taking an arithmetic average. Figures~\ref{fig:mercury_speed_7evals_high} and~\ref{fig:mercury_speed_7evals_medium} report the corresponding analysis; we obtain average TPS estimates of 600~TPS for \texttt{high} reasoning effort and 547~TPS for \texttt{medium} reasoning effort. We note that some evals (notably HumanEval and Natural2Code) have outliers that generate 50,000 tokens at extremely high speeds. These give a slightly favorable bias to our generation speed estimates and also explain why \texttt{high} reasoning effort have a higher TPS than \texttt{medium} reasoning effort. These outliers represent corrupted outputs where the model gets stuck in a thinking loop and exhaust the maximum generation length without returning a valid response.

Inception reports $\sim$1000~TPS on NVIDIA Blackwell GPUs\footnote{\url{https://www.inceptionlabs.ai/blog/introducing-mercury-2}, retrieved July 28, 2026.} while Artificial Analysis\footnote{\url{https://artificialanalysis.ai/models/mercury-2}, retrieved July 28, 2026.} reports a median of $\sim$987~TPS on undisclosed hardware. The difference to our estimate can be due to a number of reasons---in particular different hardware, serving optimizations, and/or different datasets used for measurement (this may have a large impact due to adaptive computation). It is worth noting that Artificial Analysis also reports substantial variance. Our estimate reflects the average speed a user would experience via the OpenRouter API.


\newpage

\newcommand{\nothink}[1]{\textcolor{gray}{#1}}

\begin{landscape}
\section{Denoising trajectory sampled from DiffusionGemma}
\label{appendix:denoising-trajectory}
\begin{figure}[H]
    \centering
    \includegraphics[width=1.0\linewidth]{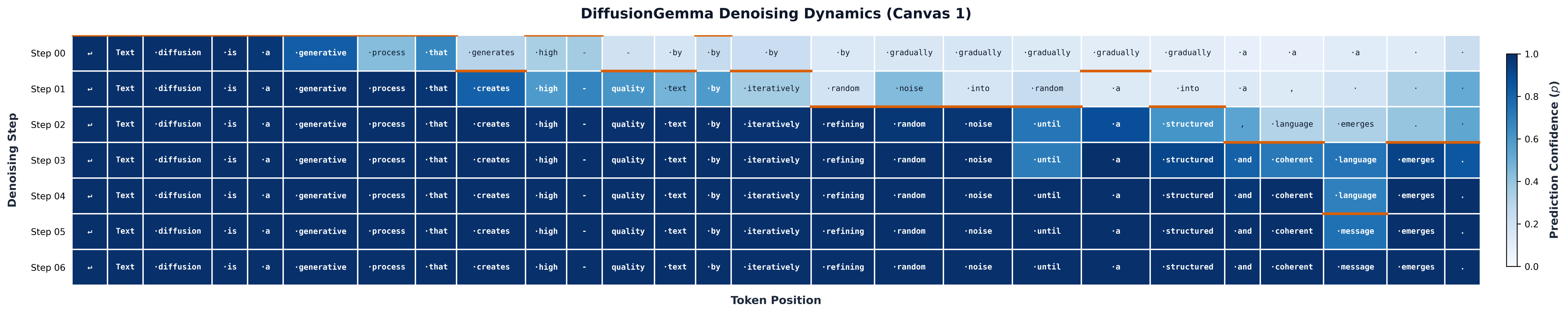}
    \caption{\textbf{Example denoising trajectory sampled from DiffusionGemma}. Prompt: ``Please write a short sentence describing text diffusion.'' The color intensity denotes model confidence. The final response is ``Text diffusion is a generative process that creates high-quality text by iteratively refining random noise until a structured and coherent message emerges.'' which requires 7 denoising steps (forward passes) to generate.}
    \label{fig:text_diffusion_sample}
\end{figure}
\end{landscape}

\section{More on Practical Advantages of Text Diffusion}

\subsection{Bidirectional Reasoning and Self-Correction}\label{app:advantages_bidir}

\begin{figure}[htbp]
\begin{tcolorbox}[
  colback=gray!5,
  colframe=gray!50,
  fontupper=\small,
  title={\sffamily\bfseries Prompt}
]
``A frog is trying to cross a 25-foot pond. Each hop covers exactly 5
feet and takes 1 minute. However, the lily pad located at exactly the
20-foot mark is magical: if the frog lands on it, it instantly
teleports the frog backward to the 5-foot mark. Is it possible to
reach the 25-foot pad?''

\medskip
\begin{tcolorbox}[colback=red!3, colframe=red!25, boxrule=0.4pt, left=4pt, right=4pt]
\textbf{Gemma~4 26B A4B:}
Yes, it is possible. The frog can reach the 25-foot mark by avoiding
the 20-foot mark. Since the frog's hops are exactly 5 feet, it would
land on the 5, 10, 15, 20, and 25-foot marks. To avoid the 20-foot
mark, the frog would need a different jump distance, but based on the
strict rule of ``exactly 5 feet'' per hop, it is impossible because
the frog is forced to land on the 20-foot mark to reach the 25-foot
mark.
\end{tcolorbox}

\smallskip
\begin{tcolorbox}[colback=green!3, colframe=green!30, boxrule=0.4pt, left=4pt, right=4pt]
\textbf{DiffusionGemma:}
No. To reach 25 feet, the frog must land on the 20-foot mark, which
teleports it back to 5 feet, creating an infinite loop.
\end{tcolorbox}

\smallskip
{\footnotesize \textbf{Analysis:} Gemma~4 commits to ``Yes'' without foreseeing the mathematical trap, resulting in a response that contradicts its own opening statement. DiffusionGemma converges cleanly to the correct ``No'' since the response and the reasoning can co-evolve over denoising. Note that both models are operating in non-thinking mode.}
\end{tcolorbox}

\vspace{1em}
\centering
\includegraphics[clip, viewport=0bp 0bp 500bp 230bp, width=\linewidth]{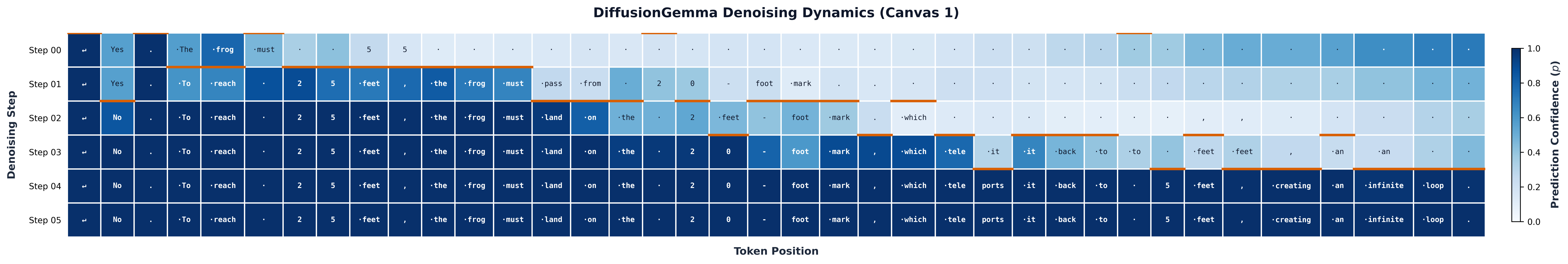}
\caption{\textbf{Denoising trace for the frog puzzle (zoomed to first 12 tokens), where color intensity indicates model confidence}. DiffusionGemma initially assigns high probability to the intuitive but incorrect ``Yes.'' After several denoising steps, bidirectional attention allows the model to realize the task is impossible and revises its initial prediction to ``No''. Denoising is completed in 6 steps.}
\label{fig:frog}
\end{figure}

Consider the frog crossing puzzle in Figure \ref{fig:frog}. This example highlights the necessity of self-correction in logical deduction. The prompt introduces a deceptive mathematical trap---an infinite loop---that the AR baseline blindly falls into, generating a contradictory ``Yes''. DiffusionGemma's parallel generation exhibits a distinct self-correction trajectory. While it initially assigns high probability to the intuitive but incorrect ``Yes'', the bidirectional propagation of the logical constraints across the sequence enables the model to revise its stance and cleanly output ``No'' before the final tokens are sampled.

\newpage
\subsection{More on Dynamic and Adaptive Computation}\label{app:advantages_adaptive}
\begin{figure}[htbp]
\begin{tcolorbox}[
  colback=gray!5,
  colframe=gray!50,
  fontupper=\small,
  title={\sffamily\bfseries Prompt}
]
``Generate a sequence of 20 tokens (0s and 1s) based on the following strict rules. To generate the next token, look at the two immediately preceding tokens: If the previous two tokens are 1, 1, the next token is 0. If the previous two tokens are 1, 0, the next token is 1. If the previous two tokens are 0, 1, the next token is 1. If the previous two tokens are 0, 0, the next token is 0. The starting sequence is: 0 1 Continue the sequence step-by-step until you reach exactly 20 tokens. Only emit the tokens, do not emit any other text.''

\medskip
\begin{tcolorbox}[colback=green!3, colframe=green!30, boxrule=0.4pt, left=4pt, right=4pt]
\textbf{DiffusionGemma:}
0 1 1 0 1 1 0 1 1 0 1 1 0 1 1 0 1 1 0 1
\end{tcolorbox}
\end{tcolorbox}

\vspace{1em}
\centering
\includegraphics[width=\linewidth]{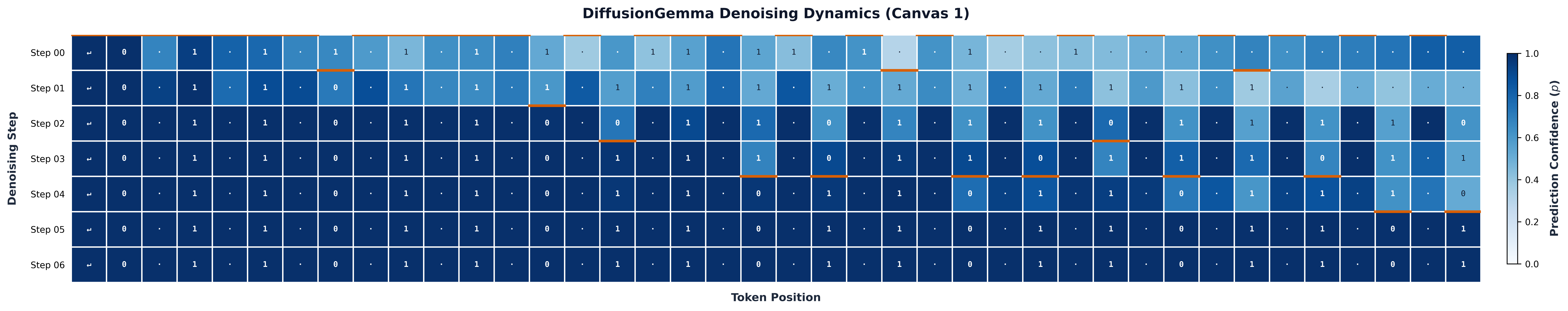}
\caption{\textbf{Denoising trace for the sequential dependency task}. Because there is a strict sequential dependency in the target output, the tokens are organically resolved in roughly a left-to-right ordering. The model adaptively allocates more compute to this ``hard'' task, taking 7 denoising steps to resolve the entire sequence of 40 tokens.}
\label{fig:sequential}
\end{figure}

\begin{figure}[htbp]
\begin{tcolorbox}[
  colback=gray!5,
  colframe=gray!50,
  fontupper=\small,
  title={\sffamily\bfseries Prompt}
]
``I will provide an input sequence of 20 binary digits. I want you to output a new sequence of the exact same length based on the following strict rules. To generate the output token at position n, look at the input sequence at position n and the input sequence at position n$-$1. If the two input tokens are 1, 1, the output token is 0. If the two input tokens are 1, 0, the output token is 1. If the two input tokens are 0, 1, the output token is 1. If the two input tokens are 0, 0, the output token is 0. For the very first token, assume the "previous" input token was a 0. Input sequence: 0 1 1 0 1 0 0 1 1 1 0 0 1 0 1 1 1 0 1 0. Only emit the tokens, do not emit any other text.''

\medskip
\begin{tcolorbox}[colback=green!3, colframe=green!30, boxrule=0.4pt, left=4pt, right=4pt]
\textbf{DiffusionGemma:}
0 1 0 1 1 1 0 1 0 0 1 0 1 1 1 0 0 1 1 1
\end{tcolorbox}

\smallskip
{\footnotesize \textbf{Analysis:} Unlike the sequential task, this prompt asks the model to apply rules over a static input string. Because the causal dependency on prior \textit{outputs} is removed, DiffusionGemma can successfully resolve all local rules in parallel.}
\end{tcolorbox}

\vspace{1em}
\centering
\includegraphics[width=\linewidth]{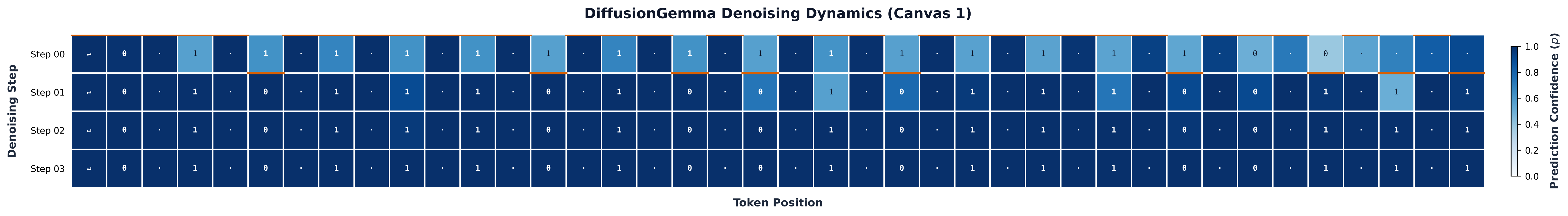}
\caption{\textbf{Denoising trace for the convolutional dependency task}. Because the rules depend entirely on the statically provided input sequence rather than the model's own dynamic output, bidirectional attention allows DiffusionGemma to evaluate the ``easy'' task in parallel, only requiring 4 denoising steps to converge.}
\label{fig:convolution}
\end{figure}

To concretely demonstrate the adaptive behavior of text diffusion, we contrast a structurally ``hard'' generation task against an ``easy'' task in Figures \ref{fig:sequential} and \ref{fig:convolution}. Both tasks involve generating a sequence of binary digits governed by identical local logical rules. However, in the structurally hard variant (Figure \ref{fig:sequential}), each new token strictly depends on the two immediately preceding \emph{generated} tokens. This causal dependency requires sequential reasoning, forcing the model to resolve the logic in a predominantly left-to-right manner. Consequently, the diffusion process adaptively expends more computational effort, requiring 7 denoising steps to fully resolve the sequence. It is crucial to note, however, that even on this structurally difficult, sequential problem, DiffusionGemma successfully decodes the entire sequence of 40 tokens (20 binary digits and 20 delimiting spaces) in only 7 denoising steps---a substantial acceleration compared to the 40 discrete sequential steps an AR model would require.

Conversely, the ``easy'' convolutional variant (Figure \ref{fig:convolution}) asks the model to apply the exact same logic rules over a statically provided \emph{input} string. Because the causal dependency on the model's own dynamic output is removed, the task lacks sequential structure. DiffusionGemma immediately leverages its bidirectional attention to independently resolve all local rules in parallel, allowing the entire output sequence to converge simultaneously in merely 4 denoising steps.

\subsection{Structured and Constrained Outputs}\label{app:advantages_structure}

\begin{figure}[H]
\begin{tcolorbox}[
  colback=gray!5,
  colframe=gray!50,
  fontupper=\small\ttfamily,
  title={\normalfont\sffamily\bfseries Prompt}
]
You are a precise data extraction assistant. Your task is to extract
information from the user's text and return it strictly as a JSON
object matching the schema below. Do not include any conversational
text, explanations, or markdown formatting. Output only the raw JSON
object.

\medskip
\normalfont\small\textbf{Schema:}

{\ttfamily\small
\setlength{\parindent}{1em}

\{\\
\hspace*{1em}"type": "object",\\
\hspace*{1em}"properties": \{\\
\hspace*{2em}"first\_name": \{ "type": "string" \},\\
\hspace*{2em}"last\_name":~~\{ "type": "string" \},\\
\hspace*{2em}"age":~~~~~~~~\{ "type": "integer" \},\\
\hspace*{2em}"hobbies":~~~~\{ "type": "array",\\
\hspace*{5.5em}"items": \{ "type": "string" \} \},\\
\hspace*{2em}"is\_student": \{ "type": "boolean" \}\\
\hspace*{1em}\},\\
\hspace*{1em}"required": ["first\_name", "last\_name",\\
\hspace*{5.5em}"age", "hobbies", "is\_student"]\\
\}
}

\medskip
\normalfont\small\textbf{Input text:} ``Hi there, my name is Arthur Pendelton. I just
turned 34 last week. In my free time, I really enjoy woodworking,
baking sourdough bread, and cycling. I currently work full-time as an
accountant, having finished my master's degree five years ago.''
\end{tcolorbox}
\caption{Prompt used for the structured JSON extraction example detailed below.}
\label{fig:json-prompt}

\vspace{0.5em}

\centering
\begin{subfigure}[t]{0.48\textwidth}
  \centering
  \includegraphics[width=\textwidth]{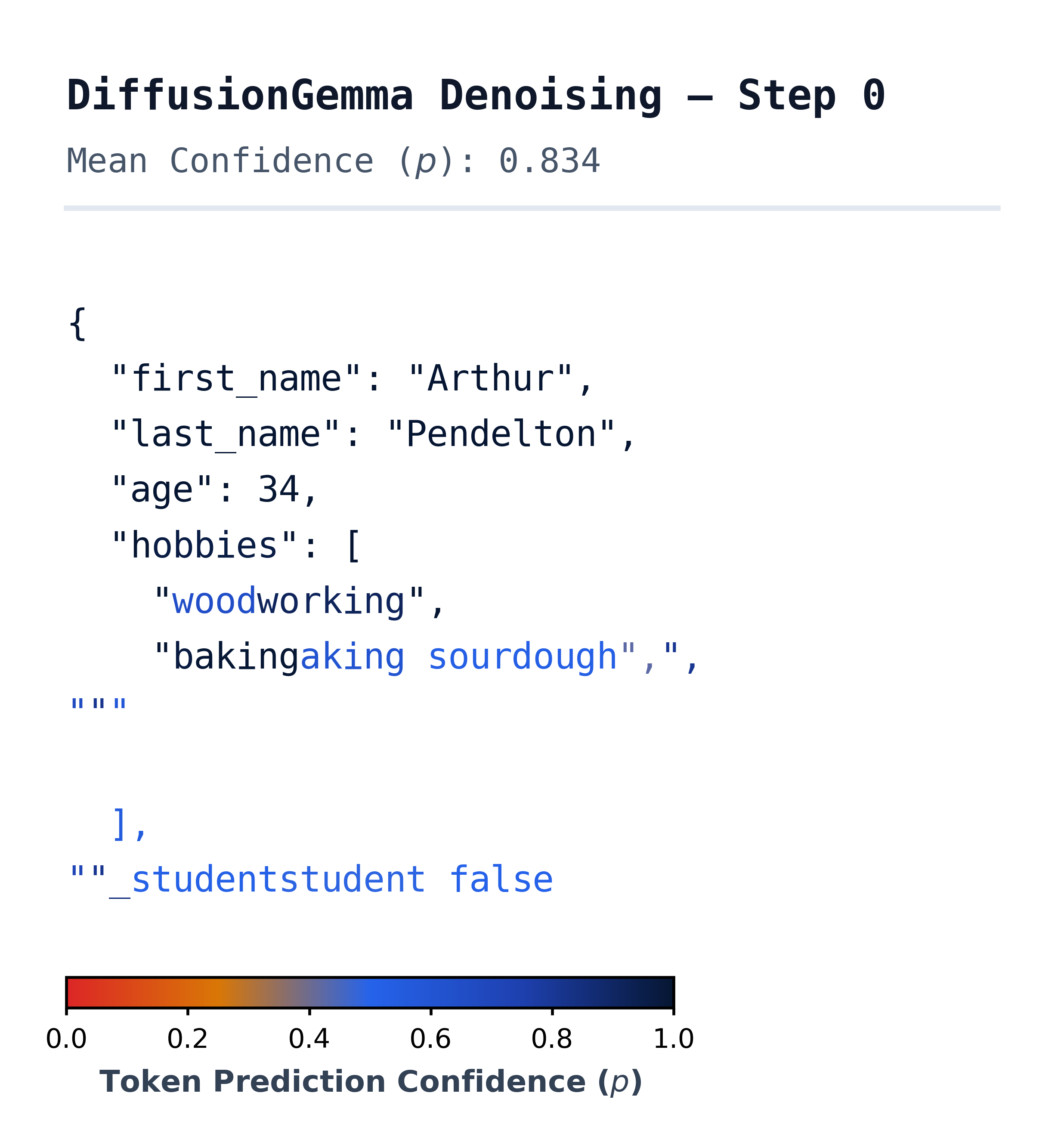}
\end{subfigure}
\hfill
\begin{subfigure}[t]{0.48\textwidth}
  \centering
  \includegraphics[width=\textwidth]{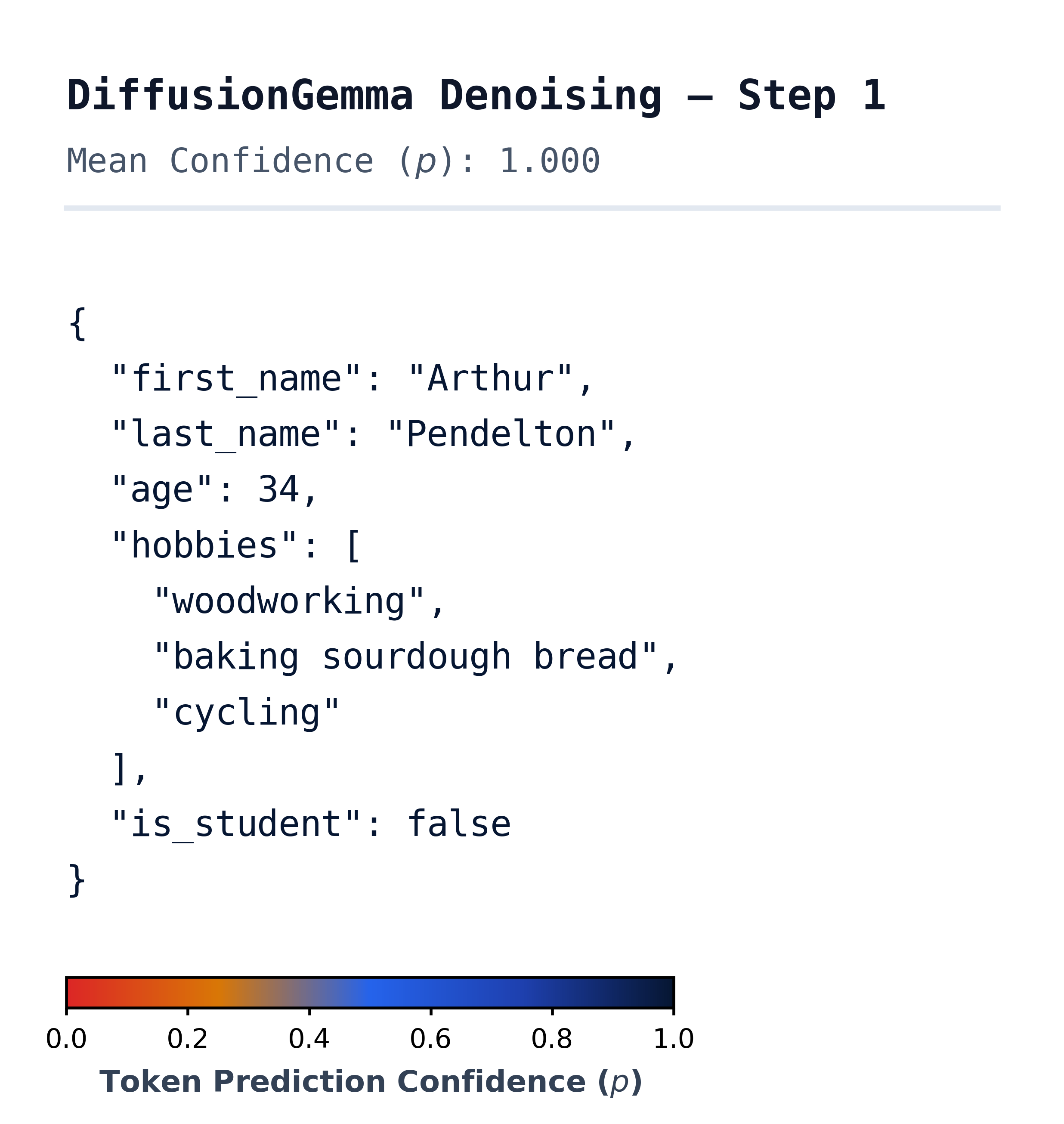}
\end{subfigure}
\caption{\textbf{DiffusionGemma structured generation.} Per-token prediction confidence across denoising steps for a JSON output task. Color encodes model confidence $p$. Because the target output is heavily dictated by the provided JSON schema, text diffusion leverages parallel generation to converge almost instantaneously. The initial canvas (left) is largely correct and exhibits an average confidence exceeding 83\% per token. By the second denoising step (right), the output has fully converged to 100\% confidence, allowing the final response to be returned in just two steps.}
\label{fig:json-denoising}
\end{figure}

\begin{figure}[htbp]
\begin{tcolorbox}[
  colback=gray!5,
  colframe=gray!50,
  fontupper=\small\ttfamily,
  title={\normalfont\sffamily\bfseries Prompt}
]
Please fix this buggy python code. Output only the fixed code, nothing else:

\medskip

def get\_even\_metrics(numbers):\\
\hspace*{2em}"""Returns the total count of even numbers and a list of them."""\\
\hspace*{2em}\# Find all even numbers\\
\hspace*{2em}evens = (n for n in numbers if n \% 2 == 0)\\
\\
\hspace*{2em}\# Calculate metrics\\
\hspace*{2em}count = sum(1 for \_ in evens)\\
\hspace*{2em}even\_list = list(evens)\\
\\
\hspace*{2em}return count, even\_list
\end{tcolorbox}
\caption{Prompt used for the buggy Python code editing example.}
\label{fig:python-prompt}

\vspace{1.5em}

\centering
\begin{subfigure}[t]{0.49\textwidth}
  \vspace{0pt}
  \centering
  \includegraphics[width=\textwidth]{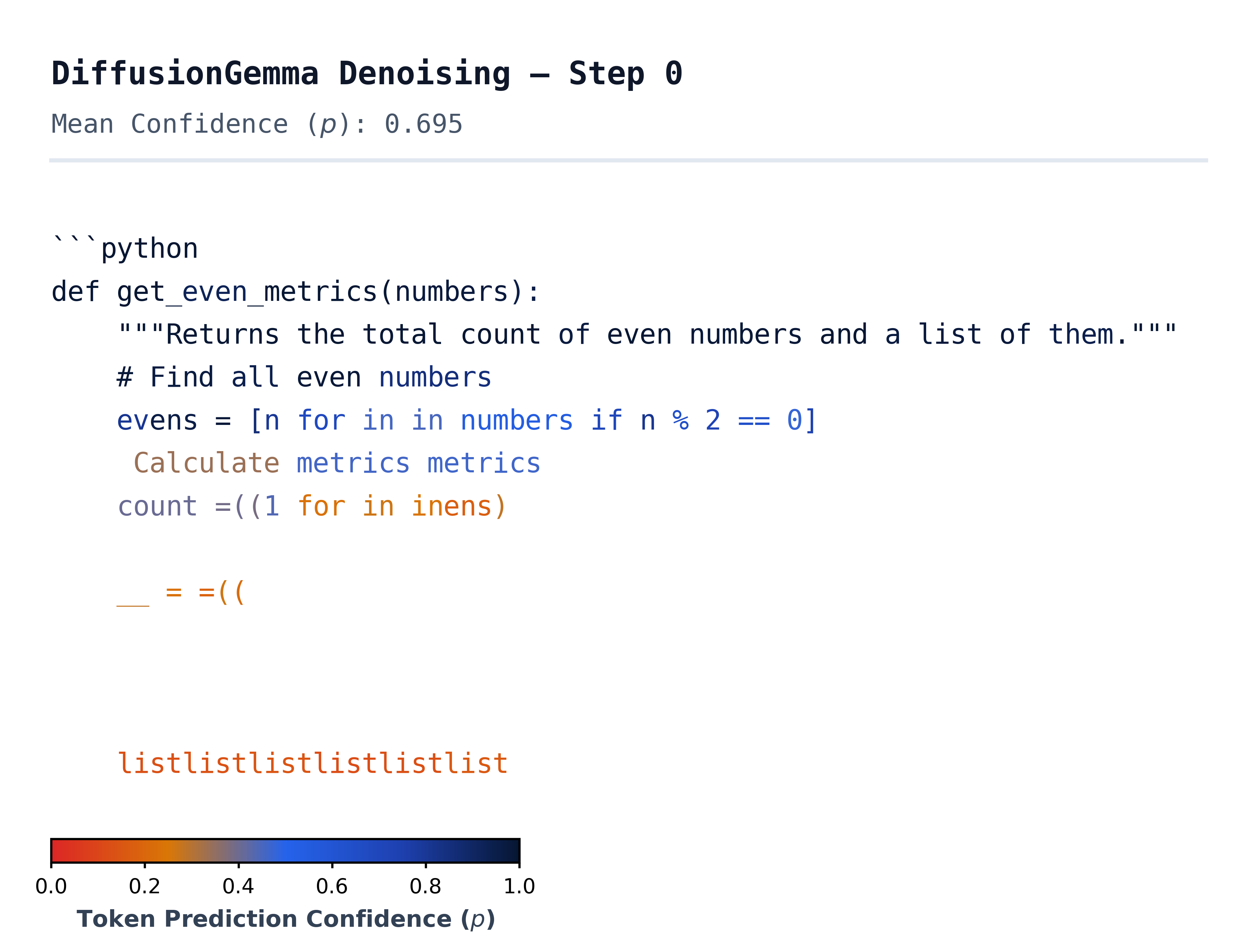}
\end{subfigure}
\hfill
\begin{subfigure}[t]{0.49\textwidth}
  \vspace{0pt}
  \centering
  \includegraphics[width=\textwidth]{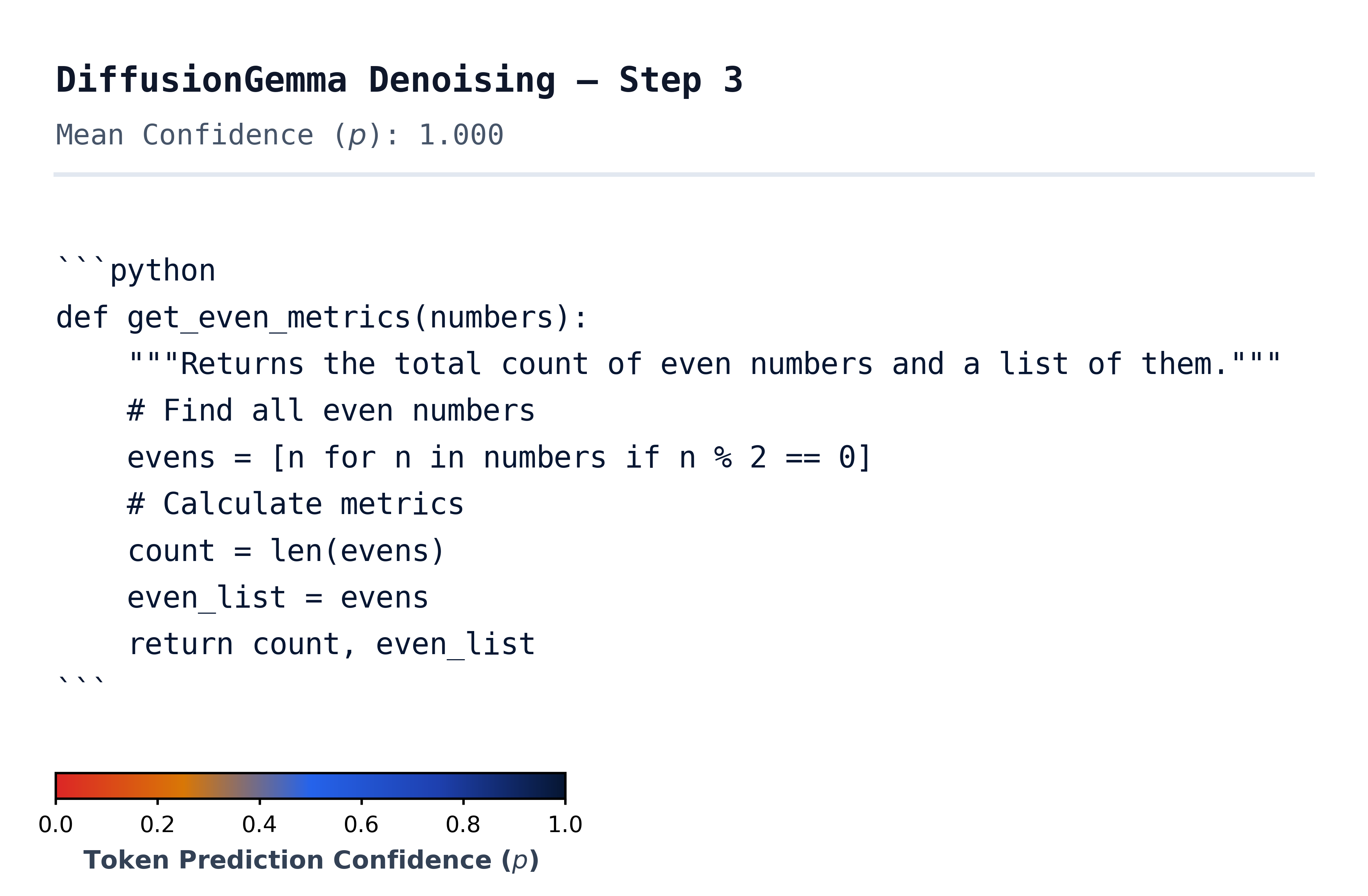}
\end{subfigure}
\caption{\textbf{DiffusionGemma code editing.} Per-token prediction confidence across denoising steps for a code debugging task. Color encodes model confidence $p$. Because the output space is strictly constrained by the input---the vast majority of tokens are copied verbatim, and only a localized semantic fix is required---the diffusion process converges rapidly. The corrected, bug-free implementation is generated in just three denoising steps.}
\label{fig:bug-fix}
\end{figure}

\end{document}